\pdfmapfile{+cm-super-t1.map}
\pdfmapfile{+cm-super-ts1.map}
\input{glyphtounicode}
\documentclass[10pt,twocolumn]{article}

\usepackage[utf8]{inputenc}
\usepackage[T1]{fontenc}
\usepackage[scaled=0.95]{helvet}

\usepackage[protrusion=true,expansion=true,stretch=15,shrink=15,step=5]{microtype}
\UseMicrotypeSet[expansion]{alltext-nott}
\usepackage{graphicx}
\usepackage{amsmath,amssymb}
\usepackage{booktabs}
\usepackage{array}
\usepackage{xcolor}
\usepackage{fancyhdr}
\usepackage{titlesec}
\usepackage{caption}
\usepackage{enumitem}
\usepackage[numbers,super,sort&compress]{natbib}
\usepackage{hyperref}

\usepackage{xurl}
\usepackage{placeins}
\usepackage[
  paperwidth=595.276pt,
  paperheight=790.866pt,
  left=40pt,
  right=40pt,
  top=46pt,
  bottom=48pt,
  columnsep=20pt
]{geometry}

\graphicspath{{figures/}}

\definecolor{NMIblue}{HTML}{0073A8}
\definecolor{NMIorange}{HTML}{F58220}
\definecolor{NMIdark}{HTML}{111111}
\newcommand{\publicationAuthorNames}{Haoyang Tian, Zhengwei Li, Houcheng Li, Xu Yang, Peigeng Gu, Xinyue Wang, Tairan Zhu, Shuo Jiang, Long Cheng}

\newcommand{\publicationAuthorListBroken}{\mbox{Haoyang Tian\textsuperscript{1,2,\textdagger}}, \mbox{Zhengwei Li\textsuperscript{1,2,\textdagger}}, \mbox{Houcheng Li\textsuperscript{1,2}}, \mbox{Xu Yang\textsuperscript{1,2}}, \mbox{Peigeng Gu\textsuperscript{1,2}},\\ \mbox{Xinyue Wang\textsuperscript{1,2}}, \mbox{Tairan Zhu\textsuperscript{1,2}}, \mbox{Shuo Jiang\textsuperscript{1,2}} \& \mbox{Long Cheng\textsuperscript{1,2,*}}}
\newcommand{\publicationAffiliations}{\textsuperscript{1}Institute of Automation, Chinese Academy of Sciences, Beijing 100190, China. \textsuperscript{2}School of Artificial Intelligence, University of Chinese Academy of Sciences, Beijing 100049, China.}
\newcommand{\publicationCorrespondingEmail}{long.cheng@ia.ac.cn}
\newcommand{\publicationCoFirstNote}{\textsuperscript{\textdagger}Co-first authors: Haoyang Tian, Zhengwei Li.}
\newcommand{\publicationContributions}{H.T. designed all methods and experiments and wrote and revised the manuscript. Z.L. proposed the sensing principle. H.L. advised on figure revisions. X.Y. advised on neural-network design. P.G. helped perform the calibration experiments. X.W. helped implement the interactive system. T.Z. and S.J. performed data cleaning. L.C. conceived the study, secured funding and supervised the research.}

\hypersetup{
  colorlinks=true,
  linkcolor=NMIblue,
  citecolor=NMIblue,
  urlcolor=NMIblue,
  pdftitle={High-resolution tactile observation makes hidden racket-control mechanisms learnable},
  pdfauthor={\publicationAuthorNames}
}

\setlist{nosep,leftmargin=*}
\DeclareCaptionLabelSeparator{bar}{\space|\space}
\titleformat{\section}
  {\large\bfseries}
  {}{0pt}{}
\titlespacing*{\section}{0pt}{1.25em}{0.45em}

\titleformat{\subsection}
  {\normalsize\bfseries}
  {}{0pt}{}
\titlespacing*{\subsection}{0pt}{0.9em}{0.25em}

\titleformat{\paragraph}[runin]
  {\normalsize\bfseries}
  {}{0pt}{}[. ]
\titlespacing*{\paragraph}{0pt}{0.65em}{0.25em}

\newcommand{\correspondingEmails}{\publicationCorrespondingEmail}

\fancypagestyle{firstpage}{%
  \fancyhf{}%
  \fancyfoot[L]{\small Haoyang Tian et al.}%
  \fancyfoot[R]{\small\thepage}%
}

\begin{document}
\pdfbookmark[0]{Main text}{document-start}
\thispagestyle{firstpage}
\twocolumn[{
\vspace*{-12pt}
{\color{NMIblue}\noindent\rule{\textwidth}{1.2pt}}\par
\vspace{8pt}
{\raggedright\fontsize{23}{26}\selectfont\bfseries High-resolution tactile observation makes hidden racket-control mechanisms learnable\par}
\vspace{14pt}
\noindent{\small\bfseries
\publicationAuthorListBroken}\par
\vspace{4pt}
\noindent{\footnotesize
\publicationAffiliations\ \textsuperscript{*}e-mail: \correspondingEmails.\ \publicationCoFirstNote}\par

\vspace{12pt}
\noindent\textbf{Abstract}\par
\vspace{3pt}
{\setlength{\parindent}{0pt}\normalsize Experts can master fine hand–object manipulation skills, yet limited conscious perception of rapid tactile interactions hinders their ability to articulate the underlying control mechanisms. Badminton racket control requires rapid, fine hand-force coordination, providing a demanding testbed for uncovering these mechanisms. An 864-node tactile–inertial glove with synchronized 100-Hz acquisition recorded over 14,000 strokes from more than 70 participants across five skills. Inspired by sensorimotor control, our mechanics-grounded two-layer framework inferred object-driving and grip-stabilizing control from tactile history, while resolving their spatiotemporal force realization. We found that higher proficiency was associated with selective rather than extensive control: greater skill specificity, anticipatory regulation, focal distal modulation and lower geometric torque cancellation, while spatiotemporal coordination showed a non-monotonic acquire–refine pattern. We translated these mechanisms into personalized guidance targeting force control and force–motion coordination. In a randomized cohort, mechanism-guided learners gained 0.72 expert-rated skill levels versus 0.05 with level-only feedback, making control mechanisms actionable.\par}
\vspace{12pt}
{\color{NMIdark}\noindent\rule{\textwidth}{0.4pt}}\par
\vspace{10pt}
}]

\begin{figure*}[p]
\centering
\includegraphics[width=\textwidth,height=0.85\textheight,keepaspectratio]{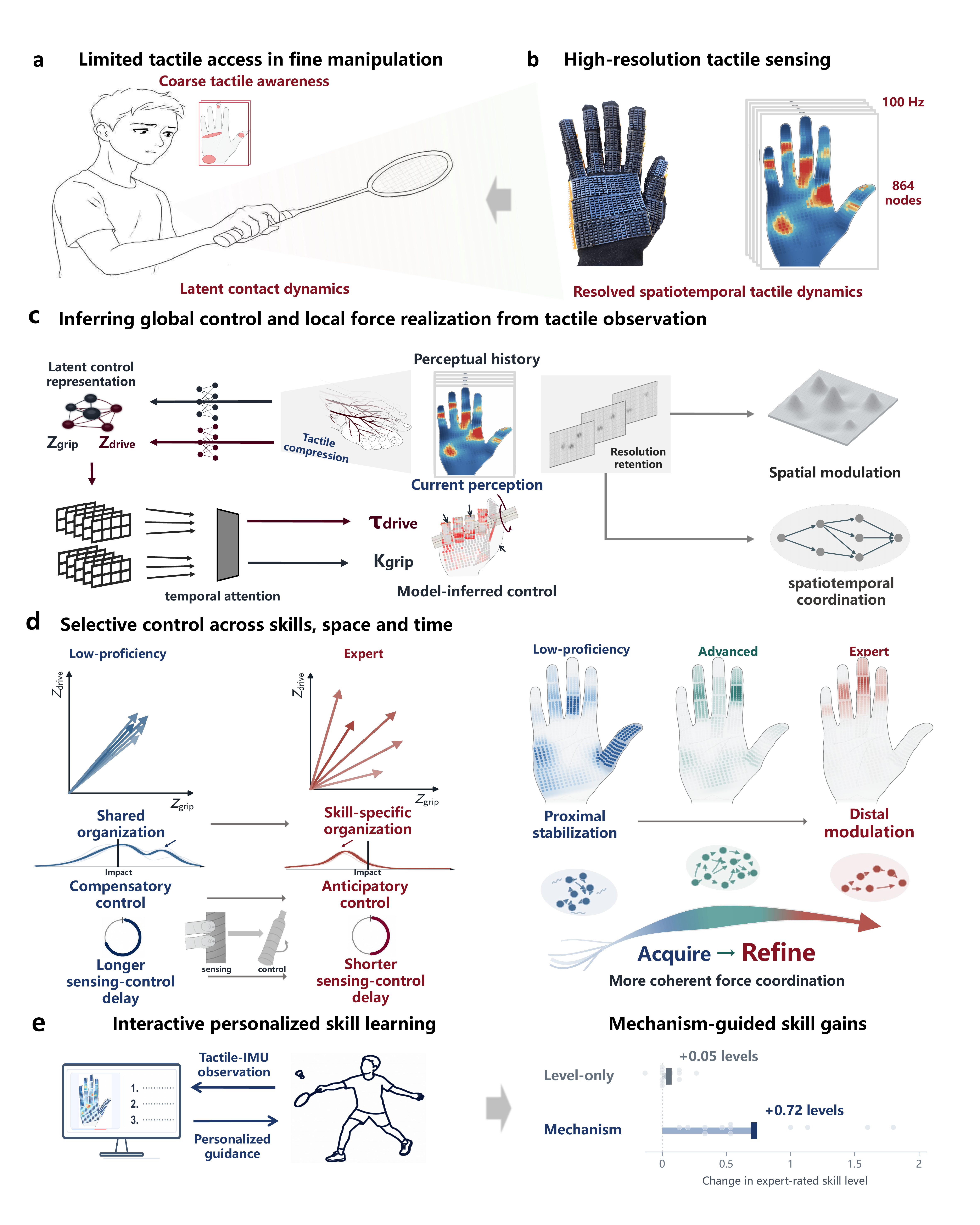}
\caption{\textbf{From high-resolution contact evidence to interpretable control knowledge and learning.} \textbf{a,} Distributed contact interactions during rapid action are difficult to consciously resolve. \textbf{b,} An 864-node glove records full-hand contact at a synchronized tactile--inertial rate of 100~Hz. \textbf{c,} Hand--racket reconstruction provides a common mechanical basis for the two layers. The control-organization layer (left) describes drive--grip regulation, while the spatiotemporal-realization layer (right) resolves local force modulation and regional coordination and assesses their mechanical effects. \textbf{d,} The control-organization layer reveals greater skill specificity, earlier preparation and shorter model-estimated perception-to-control intervals with expertise. The spatiotemporal-realization layer reveals focal distal modulation, lower geometric torque cancellation and acquire--refine coordination. \textbf{e,} Interactive learning with mechanism-guided feedback (left) and expert-rated skill gains after 10~min of practice under mechanism-guided or level-only feedback (right; $n=12$ per group). Dots show participant gains averaged across three raters and five skills. Full comparison and learning trajectories are shown in Fig.~\ref{fig:feedback-loop}.}
\label{fig:overview-sensing}
\end{figure*}

Experts can perform skilled actions fluently and almost automatically, yet often struggle to explain how they do so\cite{beilock2001fragility,flegal2008overthinking}. This is especially true for fine hand--object manipulation skills, in which highly practiced actions depend on rapidly changing, spatially distributed contact interactions that are difficult to observe, articulate and deliberately reproduce\cite{ericsson1993,masters1992,flanagan2006,johansson2009} (Fig.~\ref{fig:overview-sensing}a). Badminton racket control provides a demanding testbed for this general problem: rapid strokes require finely timed contact forces across individual finger segments and distinct palm regions\cite{song2020finehand,pan2024gripdistribution,liu2026badmintongrip}. Addressing this gap calls for machine intelligence that can first extend tactile observation beyond what humans can consciously resolve, and then translate the resulting high-dimensional contact dynamics back into control mechanisms that humans can understand and execute.

At the sensing level, human tactile perception provides only partial conscious access to rapidly changing, spatially distributed contact patterns: masking, movement-related suppression and limited tactile sequence memory can obscure localized and transient interactions\cite{auvray2008tactile,fuehrer2022suppression,klatzky2008nback}. Wearable tactile arrays provide objective measurements of the contact interface\cite{bianchi2016multimodal}, but trade-offs between spatial density and temporal sampling can still leave these fine patterns partly unresolved\cite{sundaram2019scalable,jiang2024,tian2025badmintonbackhand}. Capturing skilled interaction therefore requires dense, full-hand sensing with sufficient spatial and temporal resolution to preserve fine contact structure. At the interpretation level, however, richer measurements create a different challenge\cite{cranmer2020symbolic,rudin2019interpretable}. Fully data-driven models can exploit dense spatiotemporal signals\cite{luo2021textiles,tian2019} but often produce representations that are difficult to convert into executable control adjustments. Conversely, predefined spatial regions, temporal phases and handcrafted force descriptors\cite{pan2024gripdistribution} are easier to interpret, but may discard fine structure\cite{santello1998postural,santello2016synergies} and obscure control patterns that researchers do not already know or expect. A functional representation can draw on complementary neuromuscular mechanisms of hand–object control: differential agonist–antagonist recruitment contributes to movement-generating torque, whereas co-contraction modulates mechanical impedance to stabilize interaction\cite{hogan1984impedance,winges2007multidigit}. Interpreting these measurements therefore requires a representation that links tactile dynamics to functional control effects while retaining the local evidence needed to determine where and when those effects are realized.

We therefore developed a high-resolution tactile framework combining dense tactile observation with mechanics-grounded interpretation (Supplementary Video~1). Across the cross-sectional and learning cohorts, a full-hand tactile--inertial glove with 864 sensing nodes and synchronized acquisition at 100~Hz recorded more than 14,000 strokes across five badminton skills from over 70 participants, spanning beginner to national-team level (Fig.~\ref{fig:overview-sensing}b). Rather than reducing these measurements directly to predefined human features, we organize them at two complementary levels. The \emph{control-organization layer} uses a sensorimotor-inspired\cite{johansson1984precision,flanagan1997internal,shadmehr2010error} neural model to learn from tactile history how two coupled functions—torque-related motion generation (drive) and grasp stabilization (grip)—are coordinated across skills and stroke phases. The \emph{spatiotemporal-realization layer} asks where and when this control is physically implemented across the hand. An anatomy-informed physical reconstruction of the hand--racket interface provides the common physical basis for both layers, grounding distributed tactile measurements in their three-dimensional spatial locations and contact directions and thereby linking local contact patterns to functional control effects (Fig.~\ref{fig:overview-sensing}c). Physical constraints ground the interpretation, while the spatial units of coordination and their temporal relationships are learned from tactile evidence rather than prescribed in advance.

Comparisons across proficiency levels revealed that control became more versatile with expertise, yet was realized more selectively and coherently. Across five badminton skills, experts exhibited more skill-specific drive--grip regulation, earlier preparation for impact, more focal distal force modulation, and less geometric cancellation among torque contributions across the hand. More strikingly, spatiotemporal coordination followed a non-monotonic acquire--refine pattern: advanced players coordinated more regions over longer sequences, whereas experts combined fewer regions and shorter sequences with more coherent ordering (Fig.~\ref{fig:overview-sensing}d). We translated these mechanisms into personalized guidance targeting hand force control and force--motion coordination. With equal practice duration, mean expert-rated skill level (1--4) increased by 0.72 levels among 12 learners receiving mechanism-guided instruction, compared with 0.05 levels among 12 learners receiving level-only feedback (Fig.~\ref{fig:overview-sensing}e). Together, these results show how high-resolution tactile observation can reveal hidden racket-control mechanisms and make them learnable through personalized guidance.

\section*{Results}

\begin{figure*}[!t]
\centering
\includegraphics[width=\textwidth,height=0.68\textheight,keepaspectratio]{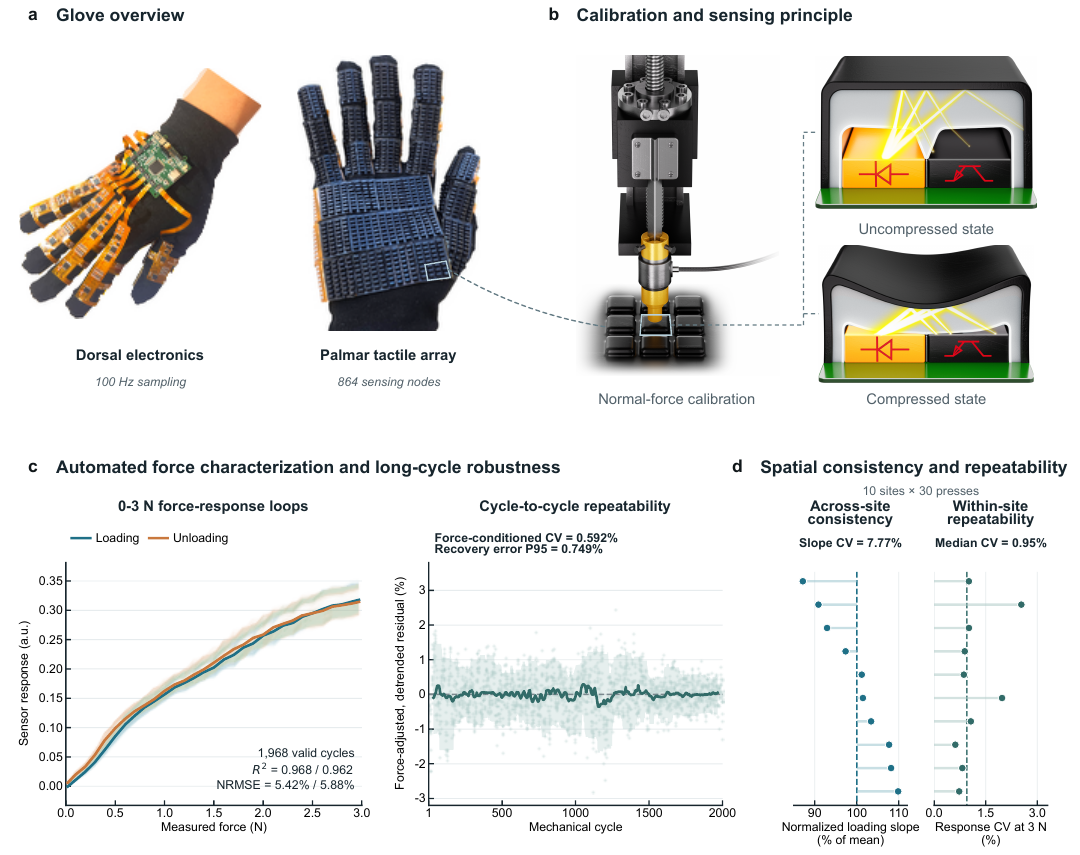}
\caption{\textbf{High-density tactile glove and sensor characterization.} \textbf{a,} Dorsal electronics (left) and palmar view of the 864-node tactile array (right), with synchronized tactile--inertial acquisition at 100~Hz. \textbf{b,} Normal-force calibration (left) and the optical sensing principle in uncompressed and compressed states (right). \textbf{c,} Automated 0--3~N loading--unloading characterization. Pale curves show 1,968 cycles with complete synchronized records; dark curves show across-cycle summaries. Loading and unloading fits use available measured points within 0--3~N. At right, force-adjusted, detrended residuals span 2,000 mechanical cycles; the dark line and shaded band show the rolling median and 5th--95th percentiles. Confidence intervals for the force-conditioned coefficient of variation (CV) use a segmented moving-block bootstrap. \textbf{d,} Loading slopes at ten sampled sites, normalized to the ten-site mean (left), and within-site CVs of the 3-N response (right). Rows are paired and sorted by slope; dashed lines mark 100\% and the median within-site CV. Each site contributed 29--30 quality-controlled presses, with 27--30 valid 3-N response bins. NRMSE, normalized root-mean-square error. Calibration, normalization and detrending procedures are detailed in Supplementary Methods~S1.}
\label{fig:sensor-characterization}
\end{figure*}

We developed an 864-node tactile--inertial glove with synchronized acquisition at 100~Hz (Fig.~\ref{fig:sensor-characterization}a and Supplementary Video~2). The study included two primary cohorts: 50 participants for mechanism discovery and 24 for skill-learning validation. The discovery database contains 7,736 strokes across five badminton skills. These high-resolution measurements enabled reconstruction of hand--racket mechanics, revealing how skilled players coordinate racket-driving acceleration with grip-related stabilization and realize this control through fine spatial and temporal coordination of hand forces. We then translated these mechanisms into personalized practice guidance and tested in the separate learning cohort whether brief mechanism-guided training could enable rapid acquisition of previously elusive hand-force and hand-motion control.

\subsection*{Sensing, reconstruction and functional validation}

Resolving fine-scale, rapid hand--object manipulation requires measurements that remain comparable across contact locations and repeated loading. Across 2,000 automated loading--unloading cycles over a 0--3~N range, the sensor showed near-linear force responses and close agreement between loading and unloading, with median cycle-wise $R^2$ values of 0.968 for loading and 0.962 for unloading. Consistent force responses across different sensing sites and repeatable responses at the same site under repeated loading supported comparison of distributed contact forces during vigorous racket-control actions (Fig.~\ref{fig:sensor-characterization}b--d and Supplementary Notes~1 and 3).

Physical reconstruction linked tactile measurements to where forces acted and how they could turn or stabilize the racket. The hand--racket pose prediction model (trained on paired tactile recordings and pose labels covering diverse grip postures) achieved validation mean absolute errors of 0.0453 in normalized grip position and $4.01^\circ$ in angle. Following physical refinement, the response-weighted median distance from tactile-active nodes to the handle surface was $0.652R$, where $R$ is the handle inradius (Supplementary Note~5). These results supported the use of reconstructed contact positions and force directions as a mechanical reference for subsequent control analysis.

To validate the functional roles of drive and grip, we compared model responses during posture holding and dynamic strokes. Without prescribing the drive--grip balance for either condition, the model showed stronger grip-related stabilization and weaker drive activity during posture holding, with the opposite pattern during dynamic strokes (Supplementary Note~6).

\begin{figure*}[!t]
\centering
\includegraphics[width=\textwidth,height=0.72\textheight,keepaspectratio,trim={0bp 16bp 0bp 14bp},clip]{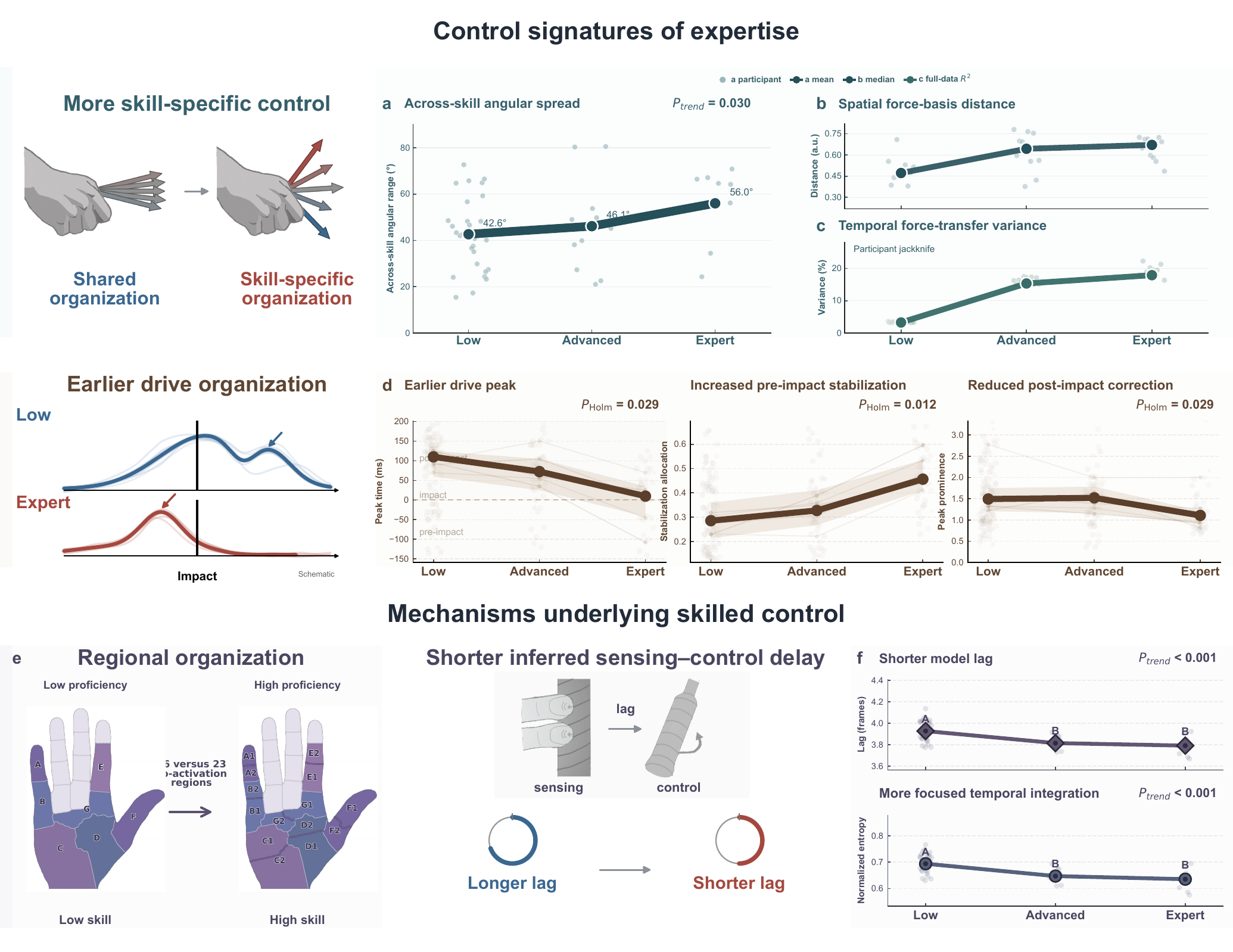}
\caption{\textbf{Task-specific functional organization and regional contact structure.} \textbf{a,} Within-participant angular range of drive--grip recruitment, defined as the maximum minus minimum angle across recorded skills. Pale points show individuals; connected large circles show group means (Low, $n=30$; Advanced, $n=11$; Expert, $n=8$). Mean ranges are 42.6, 46.1 and 56.0 degrees, respectively. The one-sided ordered permutation trend test uses these participant-level ranges ($P=0.030$). \textbf{b,} Pairwise distances between skill-specific spatial force bases, with level medians. \textbf{c,} Skill identity explains 3.25\%, 15.32\% and 17.89\% of regional recruitment-timing variance. Pale points show leave-one-participant jackknife estimates; solid points show full-data $R^2$ (participants with at least two recorded skills). \textbf{d,} Impact-aligned drive and grip-gate proxies. Thin lines summarize skills; pale points are participant--skill observations; thick lines and bands show pooled medians and 95\% confidence intervals. Trend tests use skill-adjusted ordered-level regression with participant-clustered errors; displayed $P_{\mathrm{Holm}}$ values are adjusted across the three d outcomes (raw values in Supplementary Methods~S4). \textbf{e,} Selected consensus partitions contain 16 Low and 23 pooled High co-activation regions. Matching shades identify corresponding regional families. \textbf{f,} Expected lag and normalized entropy of the learned temporal weighting profile. Points show participants; group summaries are means $\pm$ SEM (Low, $n=31$; Advanced, $n=11$; Expert, $n=8$). Different letters denote unadjusted two-sided Mann--Whitney $P<0.05$; $P_{\mathrm{trend}}$ denotes the Spearman ordered-group trend. Expected lag summarizes the model-estimated perception-to-control interval within the study cohort. Metric definitions are given in Methods and Supplementary Methods~S4.}
\label{fig:force-specificity-timing}
\end{figure*}

\subsection*{Greater skill specificity and earlier preparation\newline with expertise}

Having established the functional distinction between drive and grip, we next examined how their coordination varied across skills and proficiency levels. Higher-proficiency players tailored hand-force control more distinctly to each skill---in how they balanced drive and grip, where forces acted and when contact regions were recruited. We quantified the balance between driving object motion and stabilizing the grasp using a drive--grip angle: its mean range across skills widened from $42.6^\circ$ in Low players (beginners and intermediate players) to $56.0^\circ$ in Experts (trend $P=0.030$; Fig.~\ref{fig:force-specificity-timing}a). Spatial force patterns were also more distinct across skills in Advanced and Expert players than in Low players (Fig.~\ref{fig:force-specificity-timing}b), a difference preserved when the number of learned spatial patterns was held fixed (Supplementary Note~7). Differences between skills explained 17.89\% of the variation in inferred regional recruitment timing in Experts, compared with 3.25\% in Low players (Fig.~\ref{fig:force-specificity-timing}c).

Experts prepared the overall contact pattern earlier, while local force adjustments continued after impact. At higher proficiency, grip-related stabilization was allocated more strongly before impact, drive activity peaked earlier, and post-impact drive peaks were less prominent (Fig.~\ref{fig:force-specificity-timing}d). The average timing of force redistribution across the hand advanced from 19.0~ms after impact in Low players to 8.6~ms in Experts. In contrast, regional recruitment peaks did not advance consistently with proficiency and remained predominantly post-impact (Supplementary Note~8). Together, these findings indicate a selective shift toward anticipatory regulation, rather than a uniform advancement of all force adjustments.

Two complementary features may help explain this greater skill specificity and earlier preparation. First, differentiated regional co-activation may support more flexible, skill-specific control. In a pooled comparison, Low players had 16 co-activation regions, whereas Advanced and Expert players combined had 23 (Fig.~\ref{fig:force-specificity-timing}e). These findings suggest that skilled control treats the hand less as a uniformly coupled contact surface and more as a set of functionally differentiated regions that can be selectively coordinated to meet different task demands. This interpretation is consistent with physiological evidence that skilled motor control involves greater individuation of neuromuscular recruitment, enabling more flexible control of different effectors across task demands\cite{furuya2014individuated,kimoto2019neuromuscular}.

Second, a shorter perception-to-control interval may help explain the more anticipatory regulation observed at higher proficiency. In our human-inspired model, experts' current control depended on a more recent and concentrated portion of tactile history. The model-estimated perception-to-control interval decreased from 39.3~ms in Low players to 37.9~ms in Experts, with temporal weighting becoming more focused (ordered lag trend $P=1.67\times10^{-5}$; Fig.~\ref{fig:force-specificity-timing}f). The direction of this shift persisted across model and sampling controls (Supplementary Note~9). This pattern is consistent with motor-neuroscience evidence that practice can improve the efficiency and timing of sensorimotor and neuromuscular recruitment\cite{lay2002practice,wiegel2020training,akaiwa2020altered,yamashiro2013skillspecific}, potentially allowing skilled performers to adjust hand control more rapidly during high-speed actions.

\begin{figure*}[!t]
\centering
\includegraphics[width=\textwidth,height=0.7\textheight,keepaspectratio]{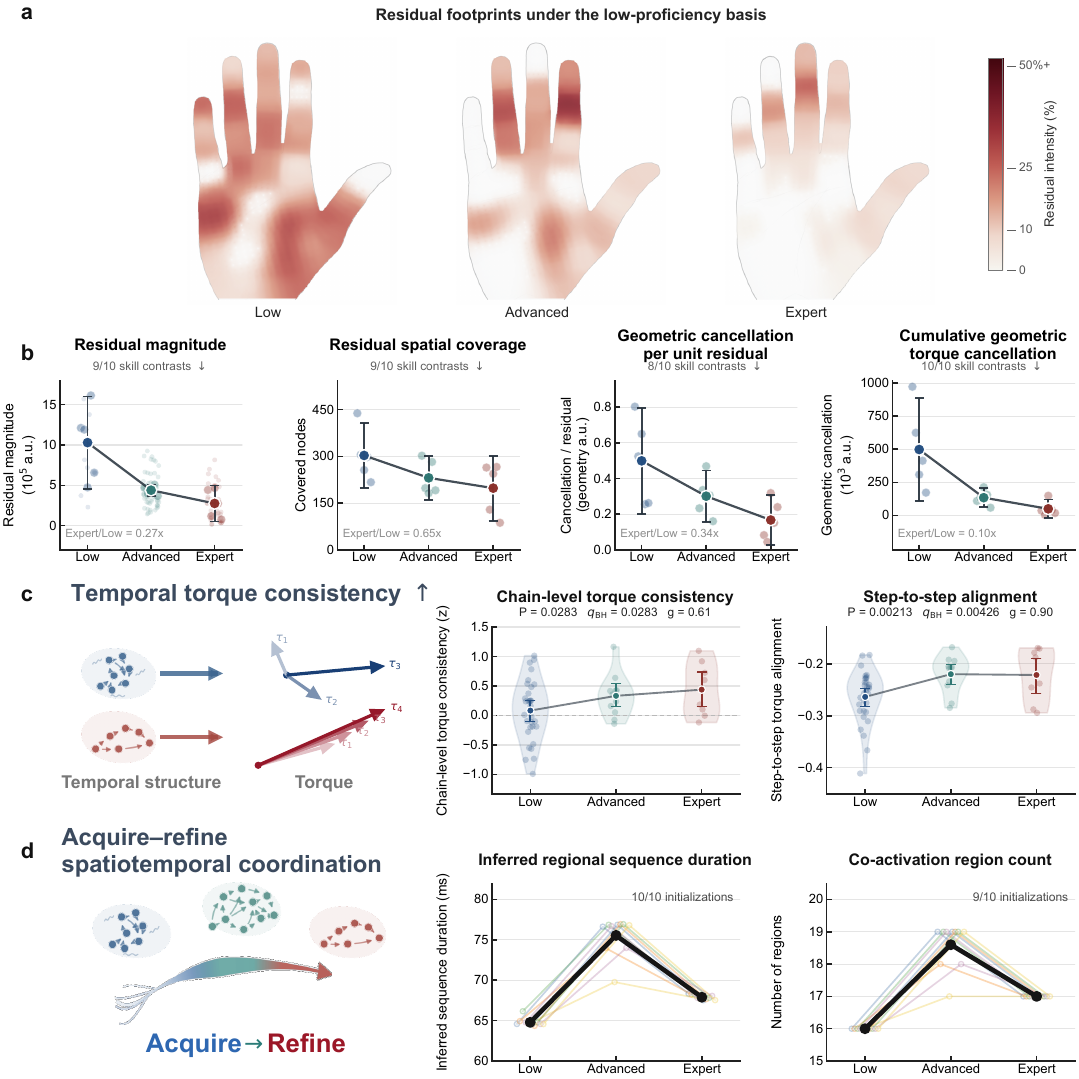}
\caption{\textbf{Focal residual contact and acquire--refine spatiotemporal coordination.} \textbf{a,} Positive residual maps relative to the low-proficiency spatial basis. \textbf{b,} Residual magnitude, spatial coverage, geometric cancellation per unit residual ($q_\tau=D_\tau/F_R$) and cumulative geometric torque cancellation ($D_\tau$). Here, $F_R$ is cumulative retained residual magnitude; $D_\tau$ accumulates the difference between summed local torque magnitudes and the resultant magnitude (Supplementary Methods~S5). Coverage counts nodes reaching the 10\% threshold in any frame of the 100-frame skill-by-group mean field. Large points and lines show means of five skill-specific estimates per group; error bars show 95\% $t$ intervals across skills. Smaller points show skill estimates. Faint participant--skill points in the magnitude panel are ratio-centred to the corresponding skill estimate. Annotations give descriptive counts of decreasing adjacent-level skill contrasts. \textbf{c,} TE-derived regional ordering evaluated with equal-magnitude unit-torque proxies from two-dimensional regional centres, using pooled-Low, Advanced and Expert maps with 16, 20 and 17 regions ($n=31$, 11 and 8 participants, respectively). Points and bars show means and bootstrap 95\% confidence intervals. Pooled High--Low comparisons use Welch tests; $q_{\mathrm{BH}}$ denotes Benjamini--Hochberg-adjusted $P$, and $g$ denotes Hedges' effect size. \textbf{d,} Inferred regional sequence duration and co-activation region count across ten algorithm initializations. Small points and thin lines show paired initialization results, with horizontal offsets for visibility; black points and thick lines show their means. Duration is the mean over 1,000 participant-bootstrap reconstructions at the original group sizes, then averaged equally across five skills within each initialization. Annotations count initializations with a strict Advanced peak. Initialization variability describes algorithmic repeatability. Skill-specific results and fixed-count spatial controls are reported in Supplementary Note~8.}
\label{fig:efficiency-compression}
\end{figure*}

\subsection*{Expertise refines local forces and regional coordination}

To examine how these broader control differences were realized locally, we characterized force modulation using patterns learned from Low players as a common reference. Experts showed more focal modulation in distal hand regions. Residual activity extended broadly across the palm and finger bases in Low players but was concentrated near the fingertips in Experts, with residual magnitude and spatial coverage at 27\% and 65\% of the Low-group values (Fig.~\ref{fig:efficiency-compression}a,b). These results indicate lower-magnitude, more localized force modulation in Experts.

Focal modulation alone does not necessarily imply mechanically coherent control: forces at different contact locations can generate opposing torques. We therefore quantified geometric torque cancellation. In the reconstructed geometry, cancellation per unit of remaining local activity was lower in Experts for every skill and reached 34\% of the Low-group value (Fig.~\ref{fig:efficiency-compression}b). Thus, expert force refinements were not only more focal but also less mechanically conflicting.

Beyond the selective spatial distribution of hand forces, skilled control also depends on how force modulation is organized over time across contact regions. We therefore identified how contact regions coordinate in a sequential manner. Higher-proficiency players exhibited more mechanically coherent temporal coordination. Chain-level torque consistency and step-to-step alignment both had higher means in Advanced and Expert players than in Low players (Fig.~\ref{fig:efficiency-compression}c), consistent with reduced geometric torque cancellation (Fig.~\ref{fig:efficiency-compression}b). Surprisingly, the broadest spatiotemporal coordination was not observed in Experts. Advanced players coordinated more co-active regions over longer inferred sequences, whereas Experts exhibited fewer regions and shorter sequences with more coherent ordering. Across ten algorithm initializations, this acquire--refine pattern occurred in 9/10 regional partitions and 10/10 five-skill-averaged sequence-duration estimates (Fig.~\ref{fig:efficiency-compression}d).

Resolving the temporal organization of hand forces depends on how the spatial units of force recruitment are defined. Predefined anatomical regions can merge distinct force patterns and obscure their temporal relationships. Learning both the number and boundaries of co-active regions for each proficiency level revealed that regional sequences were longest in Advanced players across all five skills. Imposing anatomical boundaries or a fixed region count can mix distinct force patterns, obscuring their temporal relationships and preventing consistent recovery of this pattern across skills (Supplementary Notes~8 and 10). A control mechanism can remain hidden not because it is absent, but because researchers search within the boundaries of what experts can already explain. Here, allowing the functional regions of the hand to emerge from tactile evidence revealed a spatiotemporal organization that predefined representations could obscure.

Beyond spatial representation, sensing resolution also determines how much of this organization can be resolved. Halving the sensor count retained 57.4\% of the boundaries between contact regions (Supplementary Notes~11 and 12). Reducing acquisition from 100 to 50~Hz retained 93.2--95.2\% of the repeatable timing relationships, whereas 30~Hz retained 54.6--64.3\% (Supplementary Note~13). These comparisons show that our high-resolution glove preserves the regional boundaries and timing relationships needed to reveal where local forces are refined and how hand regions coordinate.

\begin{figure*}[p]
\centering
\includegraphics[width=\textwidth,height=0.72\textheight,keepaspectratio]{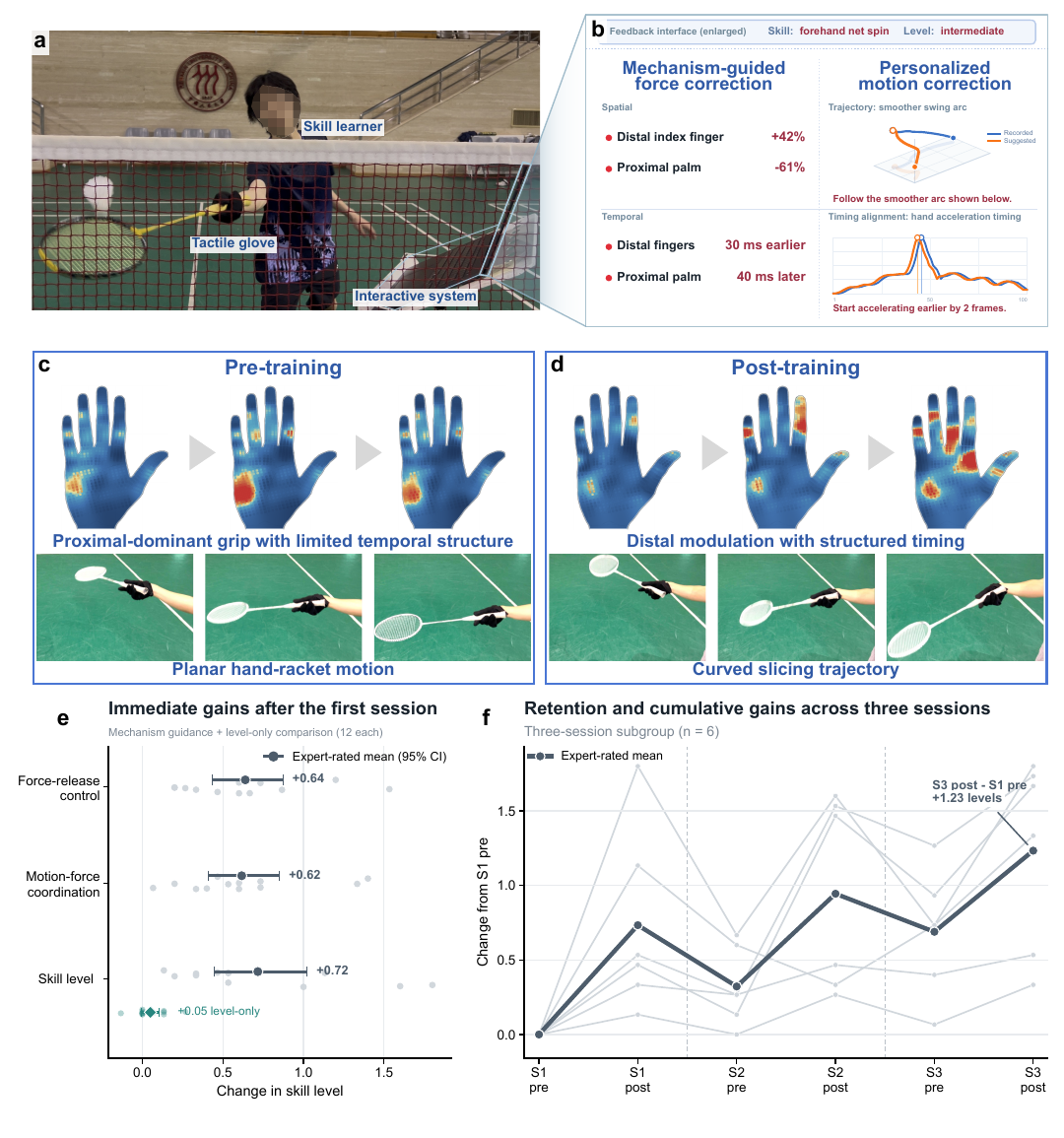}
\caption{\textbf{Mechanism-based guidance, level-only feedback and cumulative skill-level gains.} \textbf{a,} Real-world training setup combining tactile--inertial measurement with interactive feedback. \textbf{b,} Reconstructed interface linking a skill-level summary to contact guidance and personalized motion suggestions. Displayed percentages are representative learner-facing outputs. Spatial and timing differences from a task-matched Expert-group reference were ranked separately. The three largest discrepancies of each type identified locations for force adjustment and regions for earlier/later recruitment (Supplementary Methods~S6). \textbf{c,d,} Before/after exemplars contrast a proximal-dominant pattern with distal modulation across the stroke. Pale-grey arrows indicate earlier-to-later frame order; the lower row shows corresponding racket motion. All six tactile maps use the same intensity normalization. \textbf{e,} Participant-level changes in skill level after the first session. Skill level denotes the expert-rated 1--4 proficiency level assigned from observed task performance. Grey points and dark circles show mechanism-group expert-rated changes and means ($n=12$). Teal points and the diamond show expert-rated skill-level changes and the mean in the separate level-only group ($n=12$). Each group comprised four beginners, four intermediate players and four advanced players. Practice duration and raters were matched. Intervals are participant-bootstrap 95\% confidence intervals. Other outcome dimensions refer to the mechanism-feedback group. Force-release control is a stroke-specific composite: the four net-spinning skills use the formal hand-action/spin item and the rear-court transition uses release speed. Whole-body coordination is omitted from this panel. \textbf{f,} Trajectories of six mechanism-group participants (two per proficiency level) across three sessions separated by 1-day intervals, expressed as changes in expert-rated skill level from each participant's S1-pre value. Thin grey lines show individual trajectories; the dark line shows their mean. Dashed lines separate sessions. S2-pre and S3-pre describe maintenance between assessments, while S3-post is the cumulative endpoint. Confidence intervals are not drawn in f.}
\label{fig:feedback-loop}
\end{figure*}

\subsection*{Mechanism-guided practice improves skill and control coordination}

To test whether these discoveries could inform human learning, we translated the identified control patterns into personalized guidance targeting drive--grip organization, spatial force distribution, regional recruitment timing and hand force--motion coordination. We converted differences between each learner's spatiotemporal force pattern and the task-matched Expert reference into personalized force guidance, while a complementary motion model proposed hand-trajectory adjustments to improve force--motion coordination. Figures~\ref{fig:feedback-loop}c,d show the pre- to post-training changes in hand-force control and hand motion for a representative learner: contact shifted from proximal loading towards distal modulation, regional forces changed from near-synchronous exertion to sequential coordination, and hand motion changed from a predominantly planar trajectory to a more curved slicing path (Supplementary Video~3).

We tested whether this mechanism-guided feedback improved skill by randomly assigning 24 learners to mechanism-guided or level-only feedback. Both groups received model-estimated proficiency-level feedback, while only the mechanism-guided group received personalized force and motion corrections. Both groups practised for 10~min across the five skills. The same three experts independently rated all pre- and post-training clips presented in shuffled order, without labels indicating feedback condition, proficiency level or assessment time. Mean expert-rated skill level (1--4) increased by 0.72 levels with mechanism guidance versus 0.05 with level-only feedback, a between-group gain difference of 0.67 levels (95\% CI 0.39--0.98; $P=0.0014$; Fig.~\ref{fig:feedback-loop}e). Within the guided group, expert ratings also increased for force-release control (+0.64 levels) and hand motion--force coordination (+0.62 levels). All 12 guided learners improved in their five-skill mean rating (Supplementary Notes~15 and 16).

Beyond the observed performance gains, we examined whether guidance also shifted the targeted control patterns. Exploratory tactile analyses showed trends towards earlier drive regulation and greater skill specificity among guided learners. Their mean drive peak advanced from 33.57 to 18.78~ms after impact, while the cross-skill range of drive--grip recruitment widened by $3.13^\circ$. In contrast, the level-only group showed essentially no advance in drive timing (48.60 to 50.09~ms) and a $2.65^\circ$ reduction in cross-skill differentiation. Although these exploratory changes were not statistically conclusive, their directions were consistent with the earlier drive recruitment and greater task differentiation associated with higher proficiency in the discovery cohort (Supplementary Notes~15 and 16). These results suggest that learners tended to adjust their internal control organization in the direction targeted by the mechanism-based guidance.

To assess retention and cumulative learning, we followed six guided learners across three sessions separated by 1-day intervals. They accumulated a 1.23-level improvement (95\% CI 0.76--1.68; Fig.~\ref{fig:feedback-loop}f). Mean gains were 0.63 levels within sessions, with a 0.33-level decline between sessions, indicating cumulative improvement with partial retention.

\section*{Discussion}

This study uncovers hidden mechanisms of skilled hand--object control and translates them into guidance that improves skill learning. We find that control becomes more versatile with expertise, yet is realized more selectively. Experts tailored their control more distinctly to each skill and prepared earlier for impact. At the local level, fine force adjustments became more focal and shifted distally, with less mechanical opposition across the hand. More strikingly, spatiotemporal coordination followed an acquire--refine pattern: Advanced players coordinated more regions over longer sequences, whereas Experts achieved more compact and coherent coordination. Together, these findings suggest a simple principle: mastery is the ability to do more while needing to deploy less.

Discovering this organization requires overcoming bottlenecks in both tactile observation and representation. During rapid manipulation, humans cannot fully sense, retain and articulate transient, spatially distributed contact interactions. Our custom-built 864-node, 100-Hz glove records these interactions at high spatial and temporal resolution, preserving fine contact structure that lower-resolution sensing can obscure. A second bottleneck arises when interpretability is recovered through predefined anatomical regions, temporal phases or handcrafted descriptors, which can project newly measured interactions back to patterns that researchers already know or expect to find. Our framework addresses these bottlenecks by preserving fine-scale tactile structure during discovery while using mechanics-grounded representations to interpret the resulting control organization. In this sense, it first extends observation beyond human perceptual bandwidth and then returns the resolved structure to human cognitive bandwidth as interpretable control knowledge.

This distinction also reframes what interpretability means for tactile machine intelligence. Much previous tactile-learning work is evaluated by how well it supports machine-side recognition, prediction or control\cite{sundaram2019scalable,jiang2024,zhao2025embedding}. For human skill learning\cite{sigrist2013augmented,luo2024adaptive}, however, the central question is not whether a movement can be identified as expert, but why it is expert---what underlying control organization gives rise to skilled performance and whether that knowledge can be made actionable for a learner. Interpretability in this setting is therefore not simply agreement with prior intuition, but the ability to trace a functional explanation back to fine-scale contact evidence and translate it into an executable control adjustment. The learning experiment provided a practical test of this criterion: with equal practice duration, mechanism-guided learners improved by 0.72 expert-rated skill levels compared with 0.05 under level-only feedback, indicating that these discoveries could extend beyond descriptions of expertise to inform actionable guidance.

Extending this discovery-to-learning approach beyond the present testbed will require task-specific validation and more accurate contact mechanics. Badminton racket control served as a highly dynamic and precision-demanding testbed. Although the mechanism-discovery framework does not rely on badminton-specific control priors or densely annotated skill labels, its generalization beyond racket control remains to be tested. The glove primarily measures normal contact force, limiting friction and tangential-force modelling in shear-dependent manipulation. Finally, the \emph{acquire--refine} pattern of spatiotemporal coordination was inferred cross-sectionally across proficiency levels and requires longitudinal validation within individual learners. More broadly, this work establishes a discovery-to-learning approach that turns previously inaccessible contact dynamics into control knowledge for human skill acquisition.

\section*{Methods}

\subsection*{Study design and tactile acquisition}

We designed the study to discover control mechanisms from hand--object interactions and test their value for learning. After characterizing the tactile system, we recorded racket control across proficiency levels. Functional analysis separated motion generation from grip stabilization; spatial and temporal analysis identified where and in what order forces were coordinated. A separate learning cohort compared mechanism-guided practice with proficiency-level feedback alone. The discovery cohort contributed 7,736 strokes for mechanism analysis. The separate learning cohort contributed 6,989 strokes across the recorded sessions, giving 14,725 strokes overall. An additional 2,779 auxiliary posture sequences supported model development.

All experimental procedures were approved by the Ethics Committee of the Institute of Automation, Chinese Academy of Sciences (protocol no. IA21-2302-140202). All participants provided informed consent to participate and to the public sharing of de-identified tactile data. Consent did not permit public release of the underlying visual-recording dataset. Separate consent for publication was obtained from all individuals whose identifiable faces appear in the article or accompanying videos.

\paragraph{Tactile--inertial sensing glove}
To resolve contact changes alongside hand motion, the glove combined 864 tactile nodes across the fingers and palm with a six-axis inertial measurement unit (IMU). Hierarchical scanning and parallel module acquisition synchronized both signals at 100~Hz (Supplementary Figs.~1--3, Supplementary Table~1 and Supplementary Methods~S1).

\paragraph{Participants and racket-control tasks}
To compare control across proficiency and tasks, we recruited 50 discovery participants: 8 beginners, 23 intermediate players, 11 advanced players and 8 expert or national-team-level players. Beginners and intermediate players were pooled as Low because both groups generally lacked professional training; Advanced and Expert remained separate. Participants performed five fine-control badminton tasks\cite{song2020finehand,tian2025badmintonbackhand}: forehand inward spinning, forehand outward spinning, backhand inward spinning, backhand outward spinning and backhand rear-court transition. Skills were recorded in separate blocks of repeated trials; beginners could skip skills they could not execute reliably. Valid trials contained a single racket-string contact with the shuttle.

\paragraph{Data preprocessing}
To compare control at the same stage of a stroke, we aligned recordings to manually annotated racket--shuttle impact. Each valid stroke yielded a 100-frame segment containing 864 tactile channels, six IMU channels and timestamps, with impact at frame 50. After excluding sequences longer than 1.1~s, 7,736 strokes were retained for analysis. Tactile signals were baseline-corrected within each skill block.

\subsection*{Anatomy-informed hand--racket reconstruction}

To relate contact to its mechanical effect, we reconstructed sensor positions and force directions relative to the racket. To train the pose estimator used to initialize physical reconstruction, eight additional participants outside the two primary cohorts provided paired tactile recordings and pose labels spanning more than 100 grip postures. A GNN--LSTM model trained on this auxiliary dataset estimated hand--racket pose from tactile input. The estimated poses initialized a physical reconstruction over the current frame.

To maintain anatomical and geometric consistency, optimization brought active sensors towards the handle, aligned contact normals, penalized penetration, constrained pose and joints, and maintained sensor-position continuity. Reconstructed positions and normals then defined contact locations and torque contributions. For this evaluation, 10 discovery participants supplied configuration selection; after freezing the configuration, we evaluated seven-frame sequences from the remaining 40 members of the same cohort. Contact-distance and trajectory summaries used the final three frames (Supplementary Note~5 and Supplementary Methods~S7).

\subsection*{Control-organization layer}

To distinguish the functional roles of motion generation and grasp stabilization, we learned two mechanics-constrained representations from tactile history. A graph-informed encoder compressed each 864-node tactile frame into 192 features. It was pretrained to reconstruct tactile input and then frozen. A frame-wise projection produced two non-negative eight-dimensional representations, $Z_{\mathrm{drive}}$ and $Z_{\mathrm{grip}}$. The decoder combined their recent history with estimated pose, velocity and IMU state. The drive branch predicted a three-axis torque proxy $\boldsymbol\tau_{\mathrm{drive}}$ in model units and a velocity increment. The grip branch produced a dimensionless gate $\mathbf{k}_{\mathrm{grip}}\in[0,1]^3$ that attenuated the increment along each axis. The schematic label $K_{\mathrm{grip}}$ in Fig.~\ref{fig:overview-sensing}c denotes this stabilization output, not a measured physical stiffness. Mechanical and motion-prediction objectives established these roles (Supplementary Methods~S8).

To estimate which part of tactile history supported current control, the model weighted representations from $t-8$ to $t-3$, corresponding to 30--80~ms at 100~Hz. This window overlaps reported latencies of rapid tactile feedback during precision grip\cite{johansson1987signals}. The six weights combined a shared component with participant-specific offsets and were normalized to sum to one. Each participant used the same weights across drive and grip branches and across skills. Proficiency labels were not inputs. Main control training used fixed uniform weights. We then held the encoder, projections and decoder fixed and fitted only the temporal weights. Expected lag and normalized temporal entropy were
\begin{equation}
L=\sum_{k=3}^{8}w_k k,
\end{equation}
\begin{equation}
H_{\mathrm{temp}}=-\frac{\sum_{k=3}^{8}w_k\log w_k}{\log 6}.
\end{equation}
Here, $w_k$ is the weight assigned to a delay of $k$ frames. Multiplying $L$ by 10~ms gives the model-estimated perception-to-control interval. Lower $H_{\mathrm{temp}}$ means that the model concentrates on a narrower portion of tactile history.

Control-model development used the 7,736 discovery-cohort strokes together with 2,779 auxiliary posture sequences. A random 90/10 sequence-level split allocated these stroke and posture sequences to training and validation. Temporal-weight comparisons used all 50 discovery participants (31 Low, 11 Advanced and 8 Expert). Comparisons of drive--grip recruitment across skills required at least two recorded skills. Architecture and training details are provided in Supplementary Methods~S8 and Supplementary Notes~6 and 9.

\paragraph{Drive--grip recruitment and impact-related metrics}
For each participant and skill, drive--grip recruitment used the scalar trajectories $d(t)=\|Z_{\mathrm{drive}}(t)\|_2$ and $g(t)=\|Z_{\mathrm{grip}}(t)\|_2$ over frames 10--90. For the pre- and post-impact phases separately, we subtracted a five-frame baseline, applied fixed axis scales derived from the discovery data, and fitted a principal direction through the origin. Writing the fitted unit direction as $\mathbf v=(v_g,v_d)$, with grip and drive components respectively, each phase angle was $(180/\pi)\operatorname{atan2}(|v_d|,|v_g|)$ in degrees from the grip axis. The recruitment angle for each skill was the arithmetic mean of the two phase angles. Angular spread in Fig.~\ref{fig:force-specificity-timing}a was the maximum minus minimum skill angle within a participant, in degrees.

Drive-peak time was the timing of maximum drive activity relative to impact (Fig.~\ref{fig:force-specificity-timing}d). Stabilization allocation was $A_{\mathrm{pre}}/(A_{\mathrm{pre}}+A_{\mathrm{post}}+10^{-6})$, where $A_{\mathrm{pre}}$ and $A_{\mathrm{post}}$ were the within-participant--skill medians of stroke-wise peak prominences of the axis-averaged grip gate, before and after impact. Post-impact peak prominence was the corresponding median for $\|\boldsymbol\tau_{\mathrm{drive}}(t)\|_2$ after impact, in model units. Prominence measures a peak's height above its bounding troughs; Supplementary Methods~S4 specifies the boundary rule, time windows, aggregation and all main Fig.~3 metric definitions.

\subsection*{Spatiotemporal-realization layer}

To explain how functional control was implemented across the hand, we analysed local force distribution and regional coordination. Coordination followed one continuous workflow: identify contact regions, estimate temporal relationships, and construct an ordering tree. Region structure and recruitment order jointly described the spatial and temporal organization of contact.

\paragraph{Spatial force realization}
To identify local force patterns beyond those captured by a low-proficiency reference, we first decomposed tactile data using non-negative matrix factorization\cite{lee1999nmf},
\begin{equation}
X \approx WH.
\end{equation}
Rows of $X$ are frames and columns are sensors; rows of $H$ describe spatial force patterns, and $W$ gives their activation over time. For Fig.~\ref{fig:force-specificity-timing}b, each skill's spatial patterns were separately mean-centred across sensors and normalized to unit Euclidean norm; one-to-one matching maximized their summed cosine similarity between skills. Spatial force-basis distance was one minus the mean matched similarity, using the smaller basis count when ranks differed; each proficiency group was summarized by the median over ten skill pairs. For residual analysis, we held the Low reference basis $H$ fixed and fitted its activation to target-group data,
\begin{equation}
W^{*}=\arg\min_{W\geq0}\|X-WH\|_{F}^{2}.
\end{equation}
The residual
\begin{equation}
R=X-W^{*}H
\end{equation}
describes local activity left unexplained by this reference. For Fig.~\ref{fig:efficiency-compression}a,b, we matched reconstruction energy per segment and expressed residuals relative to the reconstructed response as percentages. Signed curves were averaged within skill-by-group combinations and smoothed. Positive responses were retained as $S_{ti}$ for frame $t$ and sensor $i$. Coverage counted sensors with retained responses in any frame (Supplementary Methods~S5).

To assess the residual's mechanical contribution, we combined it with reconstructed contact positions and normals. Cumulative geometric cancellation, $D_\tau$, measured how much local torque contributions opposed one another. Dividing by cumulative retained percentage-residual response, $F_R=\sum_{t,i}S_{ti}$, gave cancellation per unit residual, $q_\tau=D_\tau/F_R$, defined when $F_R>0$. These quantities use residual-response and reconstructed-geometry units. Evaluation used 3,820 segments from 25 participants (6 Low, 11 Advanced and 8 Expert), comprising 102 participant--skill combinations (Supplementary Methods~S5).

\paragraph{From contact regions to temporal coordination}
To identify the spatial units whose timing would be compared, we grouped neighbouring sensors with coordinated activity into regions of interest (ROIs). The 864-node graph combined physical adjacency, tactile co-activation and contact-normal similarity. Leiden\cite{traag2019leiden} resolution was a searched parameter, selected for each proficiency group by the same held-out-person criterion. Consensus across participant resamples identified recurring boundaries. These contact-based regions supplied both the spatial map and the regional signals for the subsequent timing analysis. The Low-versus-pooled-High comparison in Fig.~\ref{fig:force-specificity-timing}e summarizes regional granularity (16 versus 23 regions), whereas the acquire--refine analysis uses separate Low, Advanced and Expert partitions. The selected reference has 16, 20 and 17 regions, respectively. Fig.~\ref{fig:efficiency-compression}d summarizes ten additional consensus initializations with spatial settings held fixed.

To determine how activity progressed between these regions, we aggregated tactile signals within each ROI and analysed their frame-to-frame changes. Conditional transfer entropy (TE)\cite{schreiber2000transfer,barnett2009gaussian} measured how much one region's past activity improved prediction of another region's current activity beyond its own past and changes in total hand force. For each directed pair, we compared delays of 1--5 frames (10--50~ms) and retained the delay with the largest TE. Links were tested against 200 within-stroke circular shifts and corrected for multiple comparisons. A cohort link was considered stable when it recurred in at least 70\% of contributing participants.

To summarize regional recruitment order, we constructed a directed tree maximizing retained TE weight, excluding cycles and allowing each region at most one preceding region. Disconnected branches remained separate. Summing delays along each branch assigned relative recruitment times; their range defined completion time. Region count described the spatial extent of co-activation, and sequence duration described its temporal span. For each initialization, we averaged duration over 1,000 participant-bootstrap reconstructions at the original group sizes, then equally across five skills.

To assess the mechanical coherence of this spatial and temporal sequence, we assigned each ROI a unit-length torque direction from its centre on the two-dimensional hand map. Equal-magnitude, two-dimensional torque proxies isolated the geometric effects of regional location and recruitment order from force magnitude. At each step, torque consistency was the length of the cumulative torque vector divided by the number of included unit vectors. Chain-level torque consistency was the area under this curve over normalized sequence progress, expressed as a z-score relative to 1,000 randomized orders. Integration began with the second-step value, extended to the start of the sequence. Step-to-step alignment was the mean cosine between each new torque vector and the cumulative vector from preceding steps, omitting steps with a zero cumulative vector. The full workflow used 50 participants (31 Low, 11 Advanced and 8 Expert), contributing 217 participant--skill observations. Comparisons of regional timing across skills required at least two recorded skills. Details are provided in Supplementary Methods~S10.

Temporal force-transfer variance in Fig.~\ref{fig:force-specificity-timing}c quantified the percentage of regional recruitment-timing variation explained by skill identity. After subtracting each participant's cross-skill mean timing profile and standardizing regional coordinates within each proficiency group, we calculated the ratio of between-skill to total sums of squares, expressed as a percentage. Missing regional times were handled as described in Supplementary Methods~S4.

\paragraph{Resolution and representation controls}
To assess regional representations, we compared Anatomical15 (14 finger regions and one palm region)\cite{sundaram2019scalable}, Learned15 (adaptive regions merged to 15 by adjacent-region co-assignment) and Adaptive (learned boundaries and region counts). Maps were fitted on 24 participants and evaluated on the other 26, with the split reversed for sensitivity. Outcomes were whole-hand normalized-waveform reconstruction error, within-region dispersion of palm activity time centroids, and within-region palm force-direction coherence. Scores were averaged within participant before paired sign-flip tests, participant-bootstrap intervals and joint Holm correction (Supplementary Methods~S11). Separately, coordination controls used fixed counts of 15--25 with boundaries learned separately by proficiency or shared across groups, plus the anatomical map (Supplementary Note~8). Spatial resolution analyses reduced tactile nodes from 864 to 27; temporal analyses reduced sampling from 100 to 20~Hz (Supplementary Methods~S13--S14).

\subsection*{Mechanism-guided skill learning}

To make the mechanisms actionable, we compared each learner's impact-aligned contact pattern and regional timing with an Expert-group spatiotemporal force reference for the same skill. Spatial and timing differences were ranked separately by absolute magnitude. The three largest spatial discrepancies were used to identify locations where contact should be increased or reduced, while the three largest timing discrepancies identified regions whose recruitment should be advanced or delayed. Drive--grip profiles supplied functional context, and the interface displayed these prioritized instructions alongside tactile maps, timing visualizations and synchronized video (Supplementary Methods~S6).

To suggest motion changes compatible with the learner's contact pattern, a skill-specific reward model $\mathcal{R}(X,U)$ scored coordination between tactile sequence $X$ and six-axis inertial trajectory $U$. Training favoured higher-proficiency sequences and recorded tactile--motion pairings over cross-skill mismatches, temporally shifted pairs and perturbed motion. These ranking objectives defined the learned coordination score. A learned motion model generated candidate trajectories while tactile input was held fixed. We selected the candidate using
\begin{equation}
U^{*}
=
\arg\max_{U_s\in\mathcal{C}}
\left[
\mathcal{R}(X_{\mathrm{fixed}},U_s)
-
\frac{\lambda_d}{6T}
\|U_s-U_{\mathrm{orig}}\|_{F}^{2}
\right].
\end{equation}
Here, $U\in\mathbb{R}^{T\times6}$ contains three acceleration and three angular-velocity channels, standardized using training-set means and standard deviations. $\mathcal{C}$ is the candidate set, $s$ indexes candidate trajectories, $U_{\mathrm{orig}}$ is the recorded trajectory, $T$ is the number of frames and $\lambda_d$ weights deviation. Dividing the squared Frobenius norm by $6T$ gives mean-squared deviation over all normalized IMU samples. The illustrated search used eight candidates and $\lambda_d=0$. The selected trajectory was converted into a hand-motion suggestion (Supplementary Methods~S6).

To test learning benefits, a computer program randomly assigned 24 learners, distinct from the discovery cohort, after selection was complete. Twelve received machine-estimated proficiency-level feedback plus mechanism-derived contact and motion guidance; twelve received machine-estimated proficiency-level feedback alone. Each group contained four beginners, four intermediate players and four advanced players. Practice duration was matched at 10~min, with learners freely allocating time among the five skills. The system delivered feedback; experts assessed performance separately.

To evaluate performance gains, the same three experts rated pre- and post-training clips from all five skills. Clips were shuffled and proficiency and feedback-group identifiers removed. Experts scored each independently, with instructions not to infer assessment time. The primary outcome was the change in expert-rated skill level (1--4), which assessed execution of the five recorded tasks. The ratings used the same four-level proficiency framework as the discovery cohort, but were assigned directly from observed task performance. Discovery-cohort proficiency additionally incorporated training background and competition history. Observable skill-level descriptors and their literature basis are provided in Supplementary Methods~S15. At each assessment, ratings were averaged across experts within each skill and then equally across skills, yielding one pre/post change per participant.

To examine which aspects of control improved, experts also rated force-release control and hand motion--force coordination in the guidance group. To assess retention and accumulation, six guided learners, two from each proficiency level, completed three sessions (S1--S3) separated by 1-day intervals. Cumulative gain compared S1-pre with S3-post; S2-pre and S3-pre assessed performance maintained between sessions (Supplementary Methods~S6 and S15).

To examine internal control changes, exploratory tactile analyses compared early and late trials for each participant, skill and session. Endpoints comprised the first and last $k$ trials, where $k=\min\{\max[5,\lceil0.2N\rceil],\lfloor N/2\rfloor\}$ and $N$ is the trial count. This rule targeted 20\% per endpoint, included at least five when available and prevented overlap. Trials with non-zero preprocessed tactile input were retained. The discovery encoder and channel normalization remained fixed; participant identity and proficiency were not model inputs.

To quantify the targeted changes in timing and task specificity, we measured drive-peak time and the range of drive--grip recruitment angles across skills. Drive-peak time was the time of the maximum of $\|Z_{\mathrm{drive}}\|$ over frames 10--90, expressed relative to impact at frame 50. Times were averaged within each skill and then equally across the five skills. Recruitment angles described the balance of drive and grip activity using the same pre-/post-impact baseline correction, discovery-derived axis scales and direction fitting as Fig.~\ref{fig:force-specificity-timing}a. Each participant's cross-skill range was the maximum minus the minimum of the five angles. These tactile measures complemented the expert-rated primary outcome.

\subsection*{Statistical analysis}

To account for repeated strokes and skills from the same person, individual-level comparisons used participant summaries or participant-clustered inference. Ordered proficiency trends used permutation or rank-based tests. The three Fig.~\ref{fig:force-specificity-timing}d trend tests formed one Holm-corrected family. Sensor-wise comparisons controlled the false discovery rate\cite{benjamini1995fdr}; multiple regional and temporal comparisons used Holm or Benjamini--Hochberg correction where specified. Figure~\ref{fig:efficiency-compression}b summarized five skill-specific aggregate estimates per group. Figure~\ref{fig:efficiency-compression}d reports the frequency of a strict Advanced peak in region count and five-skill-averaged sequence duration across ten algorithm initializations. Full definitions are provided in Supplementary Methods~S4.

To estimate the benefit of guidance, we compared participant-average rating changes between groups with a two-sided Welch test. Confidence intervals used 20,000 participant-bootstrap draws, resampling independently within each feedback group for between-group gain intervals. A baseline-adjusted analysis used HC3 standard errors, which allow unequal residual variance.

To assess exploratory changes in tactile control, we analysed drive-peak time and recruitment-angle range at the participant level. A regression of late drive-peak time on feedback group and early drive-peak time estimated the additional advance with mechanism guidance, using HC3 standard errors and a 95\% $t$ interval. Mean recruitment-range changes used 20,000 participant-bootstrap draws. The six-person cumulative comparison paired S1-pre with S3-post.

\section*{Competing interests}
The authors declare no competing interests.

\section*{Data availability}
The de-identified tactile data supporting this study will be made available in the \texttt{data/} directory of the project repository: \url{https://github.com/haoyangtian/High-resolution-tactile-observation-makes-hidden-racket-control-mechanisms-learnable/tree/main/data}. The release will include tactile recordings from the 50-participant racket-control cohort, the 12-participant mechanism-feedback group and the separate 12-participant level-only-feedback group, together with tactile recordings and numerical hand--racket posture labels used for pose estimation. The underlying visual-recording dataset will not be publicly released, in accordance with participant consent. Identifiable facial images included in this article and the accompanying videos are published with the consent of the individuals shown.

\section*{Code availability}
The code necessary to reproduce the main results will be made available in the \texttt{code/} directory of the project repository: \url{https://github.com/haoyangtian/High-resolution-tactile-observation-makes-hidden-racket-control-mechanisms-learnable/tree/main/code}.

\bibliographystyle{sn-nature}
\bibliography{refs}

\begin{thebibliography}{10}
\expandafter\ifx\csname url\endcsname\relax
  \def\url#1{\burl{#1}}\fi
\expandafter\ifx\csname urlprefix\endcsname\relax\def\urlprefix{URL }\fi
\providecommand{\bibinfo}[2]{#2}
\providecommand{\eprint}[2][]{\url{#2}}
\providecommand{\doi}[1]{\url{https://doi.org/#1}}
\bibcommenthead

\bibitem{beilock2001fragility}
\bibinfo{author}{Beilock, S.~L.} \& \bibinfo{author}{Carr, T.~H.}
\newblock \bibinfo{title}{On the fragility of skilled performance: What governs
  choking under pressure?}
\newblock \emph{\bibinfo{journal}{J. Exp. Psychol. Gen.}}
  \textbf{\bibinfo{volume}{130}}, \bibinfo{pages}{701--725}
  (\bibinfo{year}{2001}).

\bibitem{flegal2008overthinking}
\bibinfo{author}{Flegal, K.~E.} \& \bibinfo{author}{Anderson, M.~C.}
\newblock \bibinfo{title}{Overthinking skilled motor performance: or why those
  who teach can't do}.
\newblock \emph{\bibinfo{journal}{Psychon. Bull. Rev.}}
  \textbf{\bibinfo{volume}{15}}, \bibinfo{pages}{927--932}
  (\bibinfo{year}{2008}).

\bibitem{ericsson1993}
\bibinfo{author}{Ericsson, K.~A.}, \bibinfo{author}{Krampe, R.~T.} \&
  \bibinfo{author}{Tesch-R{\"o}mer, C.}
\newblock \bibinfo{title}{The role of deliberate practice in the acquisition of
  expert performance}.
\newblock \emph{\bibinfo{journal}{Psychol. Rev.}}
  \textbf{\bibinfo{volume}{100}}, \bibinfo{pages}{363--406}
  (\bibinfo{year}{1993}).

\bibitem{masters1992}
\bibinfo{author}{Masters, R. S.~W.}
\newblock \bibinfo{title}{Knowledge, knerves and know-how: The role of explicit
  versus implicit knowledge in the breakdown of a complex motor skill under
  pressure}.
\newblock \emph{\bibinfo{journal}{Br. J. Psychol.}}
  \textbf{\bibinfo{volume}{83}}, \bibinfo{pages}{343--358}
  (\bibinfo{year}{1992}).

\bibitem{flanagan2006}
\bibinfo{author}{Flanagan, J.~R.}, \bibinfo{author}{Bowman, M.~C.} \&
  \bibinfo{author}{Johansson, R.~S.}
\newblock \bibinfo{title}{Control strategies in object manipulation tasks}.
\newblock \emph{\bibinfo{journal}{Curr. Opin. Neurobiol.}}
  \textbf{\bibinfo{volume}{16}}, \bibinfo{pages}{650--659}
  (\bibinfo{year}{2006}).

\bibitem{johansson2009}
\bibinfo{author}{Johansson, R.~S.} \& \bibinfo{author}{Flanagan, J.~R.}
\newblock \bibinfo{title}{Coding and use of tactile signals from the fingertips
  in object manipulation tasks}.
\newblock \emph{\bibinfo{journal}{Nat. Rev. Neurosci.}}
  \textbf{\bibinfo{volume}{10}}, \bibinfo{pages}{345--359}
  (\bibinfo{year}{2009}).

\bibitem{song2020finehand}
\bibinfo{author}{Song, X.}, \bibinfo{author}{Peng, Y.}, \bibinfo{author}{Hu,
  B.} \& \bibinfo{author}{Liu, W.}
\newblock \bibinfo{title}{Characterization of the fine hand movement in
  badminton by a smart glove}.
\newblock \emph{\bibinfo{journal}{Instrum. Sci. Technol.}}
  \textbf{\bibinfo{volume}{48}}, \bibinfo{pages}{443--458}
  (\bibinfo{year}{2020}).

\bibitem{pan2024gripdistribution}
\bibinfo{author}{Pan, K.-M.}, \bibinfo{author}{Kao, F.} \&
  \bibinfo{author}{Tsai, C.-L.}
\newblock \bibinfo{title}{Grip force and pressure distribution analysis of
  three badminton forehand overhead strokes of male and female players: A case
  study}.
\newblock \emph{\bibinfo{journal}{J. Phys. Educ. Sport}}
  \textbf{\bibinfo{volume}{24}}, \bibinfo{pages}{1231--1244}
  (\bibinfo{year}{2024}).

\bibitem{liu2026badmintongrip}
\bibinfo{author}{Liu, D.}, \bibinfo{author}{Che, L.}, \bibinfo{author}{Qi, F.},
  \bibinfo{author}{Zhang, K.} \& \bibinfo{author}{Cao, D.}
\newblock \bibinfo{title}{Evaluating the sports performance of badminton
  players based on grip strength of the real hitting scenario}.
\newblock \emph{\bibinfo{journal}{Sci. Rep.}} \textbf{\bibinfo{volume}{16}},
  \bibinfo{pages}{5055} (\bibinfo{year}{2026}).

\bibitem{auvray2008tactile}
\bibinfo{author}{Auvray, M.}, \bibinfo{author}{Gallace, A.},
  \bibinfo{author}{Hartcher-O'Brien, J.}, \bibinfo{author}{Tan, H.~Z.} \&
  \bibinfo{author}{Spence, C.}
\newblock \bibinfo{title}{Tactile and visual distractors induce change
  blindness for tactile stimuli presented on the fingertips}.
\newblock \emph{\bibinfo{journal}{Brain Res.}} \textbf{\bibinfo{volume}{1213}},
  \bibinfo{pages}{111--119} (\bibinfo{year}{2008}).

\bibitem{fuehrer2022suppression}
\bibinfo{author}{Fuehrer, E.}, \bibinfo{author}{Voudouris, D.},
  \bibinfo{author}{Lezkan, A.}, \bibinfo{author}{Drewing, K.} \&
  \bibinfo{author}{Fiehler, K.}
\newblock \bibinfo{title}{Tactile suppression stems from specific sensorimotor
  predictions}.
\newblock \emph{\bibinfo{journal}{Proc. Natl Acad. Sci. USA}}
  \textbf{\bibinfo{volume}{119}}, \bibinfo{pages}{e2118445119}
  (\bibinfo{year}{2022}).

\bibitem{klatzky2008nback}
\bibinfo{author}{Klatzky, R.~L.} \emph{et~al.}
\newblock \bibinfo{title}{An n-back task using vibrotactile stimulation with
  comparison to an auditory analogue}.
\newblock \emph{\bibinfo{journal}{Behav. Res. Methods}}
  \textbf{\bibinfo{volume}{40}}, \bibinfo{pages}{367--372}
  (\bibinfo{year}{2008}).

\bibitem{bianchi2016multimodal}
\bibinfo{author}{Bianchi, M.} \emph{et~al.}
\newblock \bibinfo{title}{A multi-modal sensing glove for human
  manual-interaction studies}.
\newblock \emph{\bibinfo{journal}{Electronics}} \textbf{\bibinfo{volume}{5}},
  \bibinfo{pages}{42} (\bibinfo{year}{2016}).

\bibitem{sundaram2019scalable}
\bibinfo{author}{Sundaram, S.} \emph{et~al.}
\newblock \bibinfo{title}{Learning the signatures of the human grasp using a
  scalable tactile glove}.
\newblock \emph{\bibinfo{journal}{Nature}} \textbf{\bibinfo{volume}{569}},
  \bibinfo{pages}{698--702} (\bibinfo{year}{2019}).

\bibitem{jiang2024}
\bibinfo{author}{Jiang, C.} \emph{et~al.}
\newblock \bibinfo{title}{Capturing forceful interaction with deformable
  objects using a deep learning-powered stretchable tactile array}.
\newblock \emph{\bibinfo{journal}{Nat. Commun.}} \textbf{\bibinfo{volume}{15}},
  \bibinfo{pages}{9513} (\bibinfo{year}{2024}).

\bibitem{tian2025badmintonbackhand}
\bibinfo{author}{Tian, H.}, \bibinfo{author}{Gao, Y.}, \bibinfo{author}{Yang,
  F.}, \bibinfo{author}{Li, Z.} \& \bibinfo{author}{Cheng, L.}
\newblock \bibinfo{title}{Preliminary study on the badminton backhand skill
  analysis based on the tactile glove}.
\newblock \bibinfo{howpublished}{In \emph{Proc. 47th Annual International
  Conference of the IEEE Engineering in Medicine and Biology Society (EMBC)}
  1--6 (IEEE, 2025)}.

\bibitem{cranmer2020symbolic}
\bibinfo{author}{Cranmer, M.} \emph{et~al.}
\newblock \bibinfo{title}{Discovering symbolic models from deep learning with
  inductive biases}.
\newblock \bibinfo{howpublished}{In \emph{Advances in Neural Information
  Processing Systems} \textbf{33}, 17429--17442 (Curran Associates, 2020)}.

\bibitem{rudin2019interpretable}
\bibinfo{author}{Rudin, C.}
\newblock \bibinfo{title}{Stop explaining black box machine learning models for
  high stakes decisions and use interpretable models instead}.
\newblock \emph{\bibinfo{journal}{Nat. Mach. Intell.}}
  \textbf{\bibinfo{volume}{1}}, \bibinfo{pages}{206--215}
  (\bibinfo{year}{2019}).

\bibitem{luo2021textiles}
\bibinfo{author}{Luo, Y.} \emph{et~al.}
\newblock \bibinfo{title}{Learning human--environment interactions using
  conformal tactile textiles}.
\newblock \emph{\bibinfo{journal}{Nat. Electron.}}
  \textbf{\bibinfo{volume}{4}}, \bibinfo{pages}{193--201}
  (\bibinfo{year}{2021}).

\bibitem{tian2019}
\bibinfo{author}{Tian, S.} \emph{et~al.}
\newblock \bibinfo{title}{Manipulation by feel: Touch-based control with deep
  predictive models}.
\newblock \bibinfo{howpublished}{In \emph{Proc. 2019 International Conference
  on Robotics and Automation (ICRA)} 818--824 (IEEE, 2019)}.

\bibitem{santello1998postural}
\bibinfo{author}{Santello, M.}, \bibinfo{author}{Flanders, M.} \&
  \bibinfo{author}{Soechting, J.~F.}
\newblock \bibinfo{title}{Postural hand synergies for tool use}.
\newblock \emph{\bibinfo{journal}{J. Neurosci.}} \textbf{\bibinfo{volume}{18}},
  \bibinfo{pages}{10105--10115} (\bibinfo{year}{1998}).

\bibitem{santello2016synergies}
\bibinfo{author}{Santello, M.} \emph{et~al.}
\newblock \bibinfo{title}{Hand synergies: integration of robotics and
  neuroscience for understanding the control of biological and artificial
  hands}.
\newblock \emph{\bibinfo{journal}{Phys. Life Rev.}}
  \textbf{\bibinfo{volume}{17}}, \bibinfo{pages}{1--23} (\bibinfo{year}{2016}).

\bibitem{hogan1984impedance}
\bibinfo{author}{Hogan, N.}
\newblock \bibinfo{title}{Adaptive control of mechanical impedance by
  coactivation of antagonist muscles}.
\newblock \emph{\bibinfo{journal}{IEEE Trans. Autom. Control}}
  \textbf{\bibinfo{volume}{29}}, \bibinfo{pages}{681--690}
  (\bibinfo{year}{1984}).

\bibitem{winges2007multidigit}
\bibinfo{author}{Winges, S.~A.}, \bibinfo{author}{Soechting, J.~F.} \&
  \bibinfo{author}{Flanders, M.}
\newblock \bibinfo{title}{Multidigit control of contact forces during transport
  of handheld objects}.
\newblock \emph{\bibinfo{journal}{J. Neurophysiol.}}
  \textbf{\bibinfo{volume}{98}}, \bibinfo{pages}{851--860}
  (\bibinfo{year}{2007}).

\bibitem{johansson1984precision}
\bibinfo{author}{Johansson, R.~S.} \& \bibinfo{author}{Westling, G.}
\newblock \bibinfo{title}{Roles of glabrous skin receptors and sensorimotor
  memory in automatic control of precision grip when lifting rougher or more
  slippery objects}.
\newblock \emph{\bibinfo{journal}{Exp. Brain Res.}}
  \textbf{\bibinfo{volume}{56}}, \bibinfo{pages}{550--564}
  (\bibinfo{year}{1984}).

\bibitem{flanagan1997internal}
\bibinfo{author}{Flanagan, J.~R.} \& \bibinfo{author}{Wing, A.~M.}
\newblock \bibinfo{title}{The role of internal models in motion planning and
  control: evidence from grip force adjustments during movements of hand-held
  loads}.
\newblock \emph{\bibinfo{journal}{J. Neurosci.}} \textbf{\bibinfo{volume}{17}},
  \bibinfo{pages}{1519--1528} (\bibinfo{year}{1997}).

\bibitem{shadmehr2010error}
\bibinfo{author}{Shadmehr, R.}, \bibinfo{author}{Smith, M.~A.} \&
  \bibinfo{author}{Krakauer, J.~W.}
\newblock \bibinfo{title}{Error correction, sensory prediction, and adaptation
  in motor control}.
\newblock \emph{\bibinfo{journal}{Annu. Rev. Neurosci.}}
  \textbf{\bibinfo{volume}{33}}, \bibinfo{pages}{89--108}
  (\bibinfo{year}{2010}).

\bibitem{furuya2014individuated}
\bibinfo{author}{Furuya, S.}, \bibinfo{author}{Nakamura, A.} \&
  \bibinfo{author}{Nagata, N.}
\newblock \bibinfo{title}{Acquisition of individuated finger movements through
  musical practice}.
\newblock \emph{\bibinfo{journal}{Neuroscience}}
  \textbf{\bibinfo{volume}{275}}, \bibinfo{pages}{444--454}
  (\bibinfo{year}{2014}).

\bibitem{kimoto2019neuromuscular}
\bibinfo{author}{Kimoto, Y.}, \bibinfo{author}{Oku, T.} \&
  \bibinfo{author}{Furuya, S.}
\newblock \bibinfo{title}{Neuromuscular and biomechanical functions subserving
  finger dexterity in musicians}.
\newblock \emph{\bibinfo{journal}{Sci. Rep.}} \textbf{\bibinfo{volume}{9}},
  \bibinfo{pages}{12224} (\bibinfo{year}{2019}).

\bibitem{lay2002practice}
\bibinfo{author}{Lay, B.~S.}, \bibinfo{author}{Sparrow, W.~A.},
  \bibinfo{author}{Hughes, K.~M.} \& \bibinfo{author}{O'Dwyer, N.~J.}
\newblock \bibinfo{title}{Practice effects on coordination and control,
  metabolic energy expenditure, and muscle activation}.
\newblock \emph{\bibinfo{journal}{Hum. Mov. Sci.}}
  \textbf{\bibinfo{volume}{21}}, \bibinfo{pages}{807--830}
  (\bibinfo{year}{2002}).

\bibitem{wiegel2020training}
\bibinfo{author}{Wiegel, P.} \& \bibinfo{author}{Leukel, C.}
\newblock \bibinfo{title}{Training of a discrete motor skill in humans is
  accompanied by increased excitability of the fastest corticospinal
  connections at movement onset}.
\newblock \emph{\bibinfo{journal}{J. Physiol.}} \textbf{\bibinfo{volume}{598}},
  \bibinfo{pages}{3485--3500} (\bibinfo{year}{2020}).

\bibitem{akaiwa2020altered}
\bibinfo{author}{Akaiwa, M.}, \bibinfo{author}{Iwata, K.},
  \bibinfo{author}{Saito, H.}, \bibinfo{author}{Sasaki, T.} \&
  \bibinfo{author}{Sugawara, K.}
\newblock \bibinfo{title}{Altered somatosensory evoked potentials associated
  with improved reaction time in a simple sensorimotor response task following
  repetitive practice}.
\newblock \emph{\bibinfo{journal}{Brain Behav.}} \textbf{\bibinfo{volume}{10}},
  \bibinfo{pages}{e01624} (\bibinfo{year}{2020}).

\bibitem{yamashiro2013skillspecific}
\bibinfo{author}{Yamashiro, K.} \emph{et~al.}
\newblock \bibinfo{title}{Skill-specific changes in somatosensory-evoked
  potentials and reaction times in baseball players}.
\newblock \emph{\bibinfo{journal}{Exp. Brain Res.}}
  \textbf{\bibinfo{volume}{225}}, \bibinfo{pages}{197--203}
  (\bibinfo{year}{2013}).

\bibitem{zhao2025embedding}
\bibinfo{author}{Zhao, Z.} \emph{et~al.}
\newblock \bibinfo{title}{Embedding high-resolution touch across robotic hands
  enables adaptive human-like grasping}.
\newblock \emph{\bibinfo{journal}{Nat. Mach. Intell.}}
  \textbf{\bibinfo{volume}{7}}, \bibinfo{pages}{889--900}
  (\bibinfo{year}{2025}).

\bibitem{sigrist2013augmented}
\bibinfo{author}{Sigrist, R.}, \bibinfo{author}{Rauter, G.},
  \bibinfo{author}{Riener, R.} \& \bibinfo{author}{Wolf, P.}
\newblock \bibinfo{title}{Augmented visual, auditory, haptic, and multimodal
  feedback in motor learning: a review}.
\newblock \emph{\bibinfo{journal}{Psychon. Bull. Rev.}}
  \textbf{\bibinfo{volume}{20}}, \bibinfo{pages}{21--53}
  (\bibinfo{year}{2013}).

\bibitem{luo2024adaptive}
\bibinfo{author}{Luo, Y.} \emph{et~al.}
\newblock \bibinfo{title}{Adaptive tactile interaction transfer via digitally
  embroidered smart gloves}.
\newblock \emph{\bibinfo{journal}{Nat. Commun.}} \textbf{\bibinfo{volume}{15}},
  \bibinfo{pages}{868} (\bibinfo{year}{2024}).

\bibitem{johansson1987signals}
\bibinfo{author}{Johansson, R.~S.} \& \bibinfo{author}{Westling, G.}
\newblock \bibinfo{title}{Signals in tactile afferents from the fingers
  eliciting adaptive motor responses during precision grip}.
\newblock \emph{\bibinfo{journal}{Exp. Brain Res.}}
  \textbf{\bibinfo{volume}{66}}, \bibinfo{pages}{141--154}
  (\bibinfo{year}{1987}).

\bibitem{lee1999nmf}
\bibinfo{author}{Lee, D.~D.} \& \bibinfo{author}{Seung, H.~S.}
\newblock \bibinfo{title}{Learning the parts of objects by non-negative matrix
  factorization}.
\newblock \emph{\bibinfo{journal}{Nature}} \textbf{\bibinfo{volume}{401}},
  \bibinfo{pages}{788--791} (\bibinfo{year}{1999}).

\bibitem{traag2019leiden}
\bibinfo{author}{Traag, V.~A.}, \bibinfo{author}{Waltman, L.} \&
  \bibinfo{author}{van Eck, N.~J.}
\newblock \bibinfo{title}{From {Louvain} to {Leiden}: guaranteeing
  well-connected communities}.
\newblock \emph{\bibinfo{journal}{Sci. Rep.}} \textbf{\bibinfo{volume}{9}},
  \bibinfo{pages}{5233} (\bibinfo{year}{2019}).

\bibitem{schreiber2000transfer}
\bibinfo{author}{Schreiber, T.}
\newblock \bibinfo{title}{Measuring information transfer}.
\newblock \emph{\bibinfo{journal}{Phys. Rev. Lett.}}
  \textbf{\bibinfo{volume}{85}}, \bibinfo{pages}{461--464}
  (\bibinfo{year}{2000}).

\bibitem{barnett2009gaussian}
\bibinfo{author}{Barnett, L.}, \bibinfo{author}{Barrett, A.~B.} \&
  \bibinfo{author}{Seth, A.~K.}
\newblock \bibinfo{title}{{Granger} causality and transfer entropy are
  equivalent for {Gaussian} variables}.
\newblock \emph{\bibinfo{journal}{Phys. Rev. Lett.}}
  \textbf{\bibinfo{volume}{103}}, \bibinfo{pages}{238701}
  (\bibinfo{year}{2009}).

\bibitem{benjamini1995fdr}
\bibinfo{author}{Benjamini, Y.} \& \bibinfo{author}{Hochberg, Y.}
\newblock \bibinfo{title}{Controlling the false discovery rate: a practical and
  powerful approach to multiple testing}.
\newblock \emph{\bibinfo{journal}{J. R. Stat. Soc. B}}
  \textbf{\bibinfo{volume}{57}}, \bibinfo{pages}{289--300}
  (\bibinfo{year}{1995}).

\end{thebibliography}


\begin{thebibliography}{1}
\expandafter\ifx\csname url\endcsname\relax
  \def\url#1{\burl{#1}}\fi
\expandafter\ifx\csname urlprefix\endcsname\relax\def\urlprefix{URL }\fi
\providecommand{\bibinfo}[2]{#2}
\providecommand{\eprint}[2][]{\url{#2}}
\providecommand{\doi}[1]{\url{https://doi.org/#1}}
\bibcommenthead

\bibitem{tian2025badmintonbackhand}
\bibinfo{author}{Tian, H.}, \bibinfo{author}{Gao, Y.}, \bibinfo{author}{Yang,
  F.}, \bibinfo{author}{Li, Z.} \& \bibinfo{author}{Cheng, L.}
\newblock \bibinfo{title}{Preliminary study on the badminton backhand skill
  analysis based on the tactile glove}.
\newblock \bibinfo{howpublished}{In \emph{Proc. 47th Annual International
  Conference of the IEEE Engineering in Medicine and Biology Society (EMBC)}
  1--6 (IEEE, 2025)}.

\bibitem{tekscanGrip2026}
\bibinfo{author}{{Tekscan}}.
\newblock \bibinfo{title}{{Grip System}}.
\newblock
  \bibinfo{howpublished}{\url{https://www.tekscan.com/products-solutions/systems/grip-system}}.
\newblock \bibinfo{note}{Accessed 25 October 2026}.

\bibitem{sundaram2019}
\bibinfo{author}{Sundaram, S.} \emph{et~al.}
\newblock \bibinfo{title}{Learning the signatures of the human grasp using a
  scalable tactile glove}.
\newblock \emph{\bibinfo{journal}{Nature}} \textbf{\bibinfo{volume}{569}},
  \bibinfo{pages}{698--702} (\bibinfo{year}{2019}).

\bibitem{jiang2024}
\bibinfo{author}{Jiang, C.} \emph{et~al.}
\newblock \bibinfo{title}{Capturing forceful interaction with deformable
  objects using a deep learning-powered stretchable tactile array}.
\newblock \emph{\bibinfo{journal}{Nat. Commun.}} \textbf{\bibinfo{volume}{15}},
  \bibinfo{pages}{9513} (\bibinfo{year}{2024}).

\bibitem{buscher2012tactile}
\bibinfo{author}{B{\"u}scher, G.~H.}, \bibinfo{author}{K{\~o}iva, R.},
  \bibinfo{author}{Sch{\"u}rmann, C.}, \bibinfo{author}{Haschke, R.} \&
  \bibinfo{author}{Ritter, H.~J.}
\newblock \bibinfo{title}{Tactile dataglove with fabric-based sensors}.
\newblock \bibinfo{howpublished}{In Proc. 2012 12th IEEE-RAS International
  Conference on Humanoid Robots 204--209 (IEEE, 2012)}.

\bibitem{li2024tactile}
\bibinfo{author}{Li, Y.} \emph{et~al.}
\newblock \bibinfo{title}{A tactile glove based on capacitive pressure sensor
  array for object shape recognition}.
\newblock \emph{\bibinfo{journal}{IEEE Sens. J.}}
  \textbf{\bibinfo{volume}{24}}, \bibinfo{pages}{39690--39698}
  (\bibinfo{year}{2024}).

\bibitem{ppsFingerTPS2026}
\bibinfo{author}{{Pressure Profile Systems}}.
\newblock \bibinfo{title}{{FingerTPS II}}.
\newblock \bibinfo{howpublished}{\url{https://pressureprofile.com/finger-tps}}.
\newblock \bibinfo{note}{Accessed 25 October 2026}.

\bibitem{ppsTactileGlove2026}
\bibinfo{author}{{Pressure Profile Systems}}.
\newblock \bibinfo{title}{{TactileGlove II}}.
\newblock
  \bibinfo{howpublished}{\url{https://pressureprofile.com/tactile-glove}}.
\newblock \bibinfo{note}{Accessed 25 October 2026}.

\bibitem{casebolt2020badminton}
\bibinfo{author}{Casebolt, K.} \& \bibinfo{author}{Zhang, P.}
\newblock \bibinfo{title}{An authentic badminton game performance assessment
  rubric}.
\newblock \emph{\bibinfo{journal}{Strategies}} \textbf{\bibinfo{volume}{33}},
  \bibinfo{pages}{8--13} (\bibinfo{year}{2020}).

\end{thebibliography}

\titlespacing*{\section}{0pt}{0.65em}{0.25em}
\section*{Acknowledgements}
This work was supported in part by Brain Science and Brain-like Intelligence Technology-National Science and Technology Major Project (Grant No.2025ZD0215600), in part by the College Student Research and Career-Creation Program of Beijing City under Grant S202414430023, in part by the Beijing Municipal Natural Science Foundation under Grant QY24385 and Grant L243014, in part by the National Natural Science Foundation of China under Grant 62333023 and Grant 62025307, and in part by the Beijing Nova Program under Grant 20230484400.

\section*{Author contributions}
\publicationContributions

\enlargethispage{\baselineskip}
\section*{Additional information}
Supplementary information is available for this paper. Correspondence and requests for materials should be addressed to L.C.
\end{document}


\pdfbookmark[0]{Supplementary information}{document-start}
\begin{center}
{\Large\supplementTitle\par}
\vspace{8pt}
{\normalsize\publicationAuthorListBroken\par}
\end{center}
\noindent{\footnotesize\publicationAffiliations\ \textsuperscript{*}Corresponding author: Long Cheng; e-mail: \publicationCorrespondingEmail.\ \publicationCoFirstNote\par}
\vspace{6pt}

\pdfbookmark[0]{Contents}{supplement-contents}
\begingroup
\small
\setlength{\parskip}{2pt}
\tableofcontents
\endgroup
\clearpage

\let\originalSupplementSection\section
\RenewDocumentCommand{\section}{s o m}{%
  \IfBooleanTF{#1}{%
    \originalSupplementSection*{\phantomsection#3}%
    \addcontentsline{toc}{section}{#3}%
  }{%
    \IfNoValueTF{#2}{\originalSupplementSection{#3}}%
      {\originalSupplementSection[#2]{#3}}%
  }%
}

\section*{Supplementary Note 1: Tactile--inertial glove platform}

\begin{figure}[!htbp]
\centering
\includegraphics[width=0.80\textwidth]{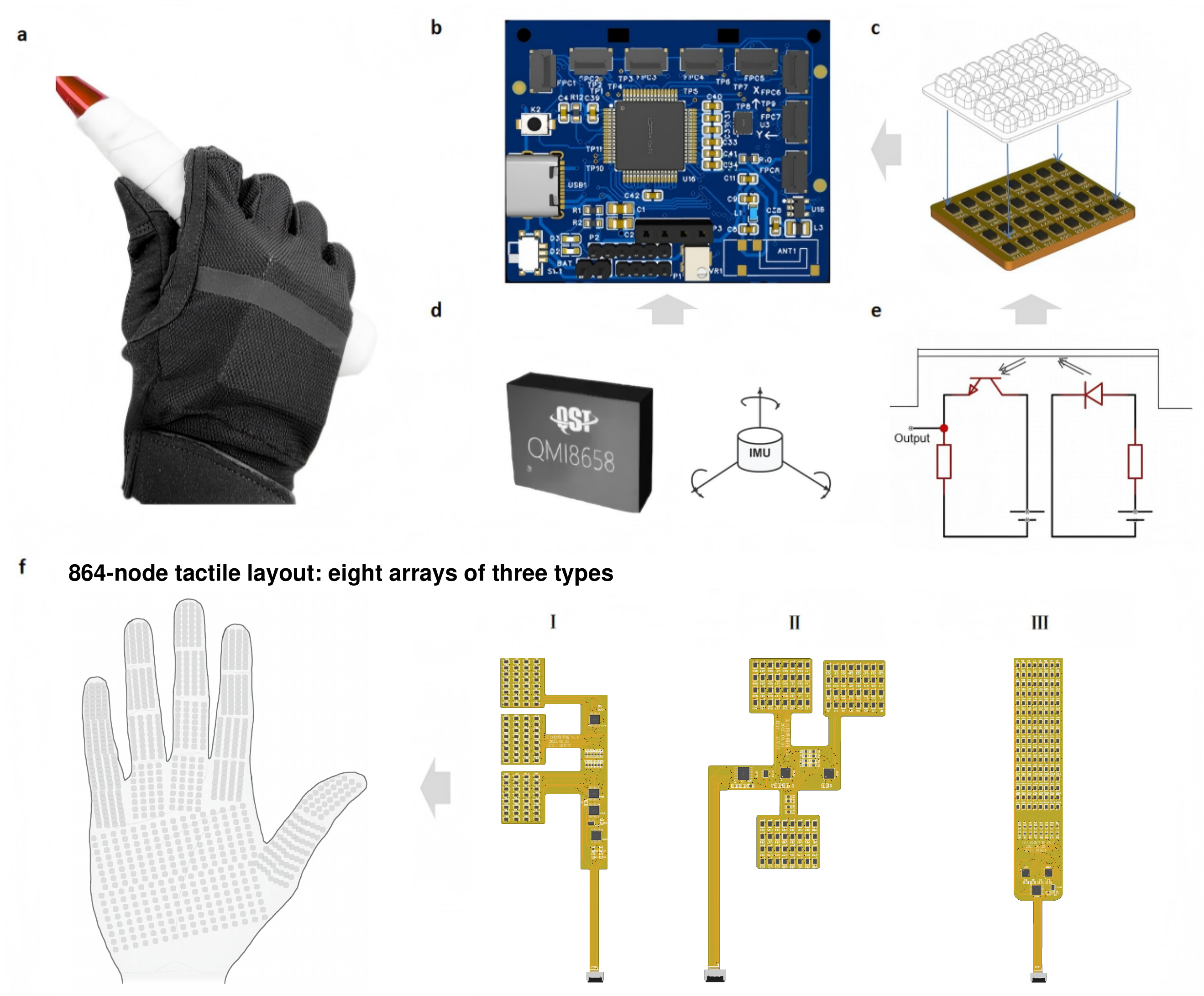}
\caption{\textbf{Tactile--inertial glove hardware.} \textbf{a,} Worn prototype. \textbf{b,} Dorsal acquisition board. \textbf{c,} Layered sensing unit with structured elastic contact layer. \textbf{d,} Six-axis inertial measurement unit. \textbf{e,} Sensing-circuit principle. \textbf{f,} Eight arrays of three types: four Type I arrays for the non-thumb fingers (96 nodes each), one Type II thumb array (96 nodes), and three Type III palm arrays (128 nodes each).}
\label{suppfig:glove-platform}\end{figure}

To capture where the hand applies force and how it moves during a stroke, the glove records normal contact signals and hand motion synchronously at 100~Hz. Four Type-I arrays cover the non-thumb fingers with 96 nodes each, one Type-II thumb array contains 96, and three Type-III palm arrays contain 128 each. These eight arrays provide $4\times96+1\times96+3\times128=864$ nodes. Flexible array tails connect to a dorsal acquisition board and six-axis inertial measurement unit (IMU). The layout covers fingertips, phalanges, thumb and palm while keeping finger gaps and bending regions flexible. Supplementary Fig.~\ref{suppfig:glove-platform} shows the sensing layers, board and layout; Supplementary Methods~S1 describes acquisition and calibration.

\FloatBarrier

\section*{Supplementary Note 2: Wearable spatiotemporal sensing capacity}

Resolving rapid hand contact requires both dense spatial coverage and frequent sampling. We therefore compared glove- and finger-worn systems by node count and system-level acquisition rate (Supplementary Table~\ref{supptable:glove-comparison}). Node count denotes physical sampling positions or manufacturer-defined sensing regions; acquisition rate denotes the reported rate for the whole system. Individual-element response times were kept separate. These measures capture the sensing capacity available within wearable constraints on flexibility, wiring and power.

The present glove combines 864 full-hand nodes with synchronized inertial acquisition at 100~Hz, yielding 86,400 sensing-node frames s\(^{-1}\). Our earlier badminton glove used 48 pressure units at approximately 50~Hz for skill recognition and finger-contribution analysis\cite{tian2025badmintonbackhand}. The present node--frame throughput also exceeds that of the 18-node, 500-Hz Tekscan system\cite{tekscanGrip2026}, the 548-node, approximately 7.3-Hz STAG glove\cite{sundaram2019} and the up-to-1,152-node, 13-Hz stretchable glove\cite{jiang2024}. This combination supports measurement of local contact changes within a stroke.

\begin{table}[!htbp]
\centering
\small
\setlength{\tabcolsep}{3pt}
\caption{\textbf{Representative tactile-glove and finger-worn sensing capacity.} Node counts and system-level acquisition rates are listed as reported by the corresponding studies or product pages. ``Not explicitly reported'' means that the source did not provide a system-level tactile acquisition rate. Element response time was not converted into acquisition frequency.}
\label{supptable:glove-comparison}
\begin{tabular}{@{}L{0.50\textwidth}L{0.23\textwidth}L{0.23\textwidth}@{}}
\toprule
System & Number of sensing nodes & Acquisition rate (Hz) \\
\midrule
Fabric-based tactile dataglove \cite{buscher2012tactile} & 54 & 1000 \\
Prior badminton glove \cite{tian2025badmintonbackhand} & 48 & 50 \\
Tekscan TactileGlove \cite{tekscanGrip2026} & 18 & 500 \\
Scalable tactile glove (STAG) \cite{sundaram2019} & 548 & 7.3 \\
Capacitive pressure-array tactile glove \cite{li2024tactile} & 137 & Not explicitly reported \\
Stretchable tactile glove for forceful interaction \cite{jiang2024} & 1,152 & 13 \\
PPS FingerTPS II \cite{ppsFingerTPS2026} & 6 & 40 \\
PPS TactileGlove II \cite{ppsTactileGlove2026} & 65 & 100 \\
This work & 864 & 100 \\
\bottomrule
\end{tabular}
\end{table}

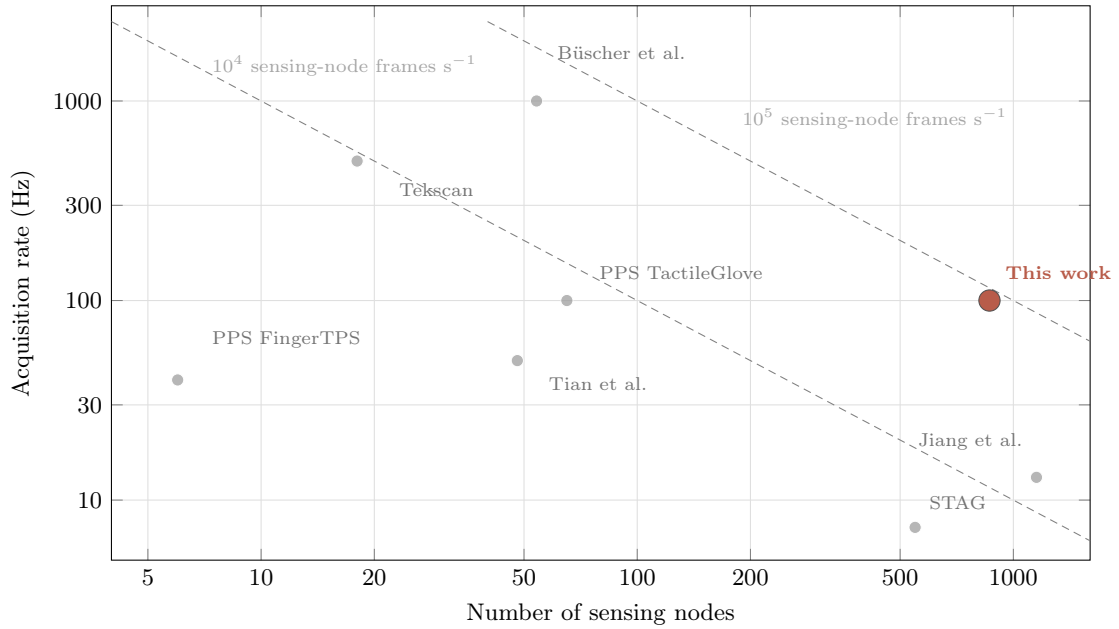
\begin{figure}[!htbp]
\centering
\definecolor{gloveMapPrior}{HTML}{9A9A9A}
\definecolor{gloveMapThisWork}{HTML}{B85C4A}
\begin{tikzpicture}
\begin{axis}[
  width=0.88\textwidth,
  height=0.54\textwidth,
  xmin=4,
  xmax=1600,
  ymin=5,
  ymax=3000,
  xmode=log,
  ymode=log,
  xlabel={Number of sensing nodes},
  ylabel={Acquisition rate (Hz)},
  xtick={5,10,20,50,100,200,500,1000},
  xticklabels={5,10,20,50,100,200,500,1000},
  ytick={10,30,100,300,1000},
  yticklabels={10,30,100,300,1000},
  grid=both,
  minor grid style={gray!12},
  major grid style={gray!25},
  tick label style={font=\small},
  label style={font=\small},
  clip=false
]
\addplot[only marks,mark=*,mark size=2.3pt,mark options={draw=white,line width=0.25pt,fill=gloveMapPrior},opacity=0.72] coordinates {(54,1000)};
\node[anchor=south west,font=\scriptsize,text=black!55] at (axis cs:58,1450) {B{\"u}scher et al.};
\addplot[only marks,mark=*,mark size=2.3pt,mark options={draw=white,line width=0.25pt,fill=gloveMapPrior},opacity=0.72] coordinates {(48,50)};
\node[anchor=north west,font=\scriptsize,text=black!55] at (axis cs:55,46) {Tian et al.};
\addplot[only marks,mark=*,mark size=2.3pt,mark options={draw=white,line width=0.25pt,fill=gloveMapPrior},opacity=0.72] coordinates {(18,500)};
\node[anchor=north west,font=\scriptsize,text=black!55] at (axis cs:22,430) {Tekscan};
\addplot[only marks,mark=*,mark size=2.3pt,mark options={draw=white,line width=0.25pt,fill=gloveMapPrior},opacity=0.72] coordinates {(548,7.3)};
\node[anchor=south west,font=\scriptsize,text=black!55] at (axis cs:565,8.1) {STAG};
\addplot[only marks,mark=*,mark size=2.3pt,mark options={draw=white,line width=0.25pt,fill=gloveMapPrior},opacity=0.72] coordinates {(1152,13)};
\node[anchor=south east,font=\scriptsize,text=black!55] at (axis cs:1120,16) {Jiang et al.};
\addplot[only marks,mark=*,mark size=2.3pt,mark options={draw=white,line width=0.25pt,fill=gloveMapPrior},opacity=0.72] coordinates {(6,40)};
\node[anchor=south west,font=\scriptsize,text=black!55] at (axis cs:7,52) {PPS FingerTPS};
\addplot[only marks,mark=*,mark size=2.3pt,mark options={draw=white,line width=0.25pt,fill=gloveMapPrior},opacity=0.72] coordinates {(65,100)};
\node[anchor=south west,font=\scriptsize,text=black!55] at (axis cs:75,115) {PPS TactileGlove};
\addplot[only marks,mark=*,mark size=4.0pt,mark options={draw=black!70,line width=0.35pt,fill=gloveMapThisWork}] coordinates {(864,100)};
\node[anchor=south west,font=\scriptsize\bfseries,text=gloveMapThisWork] at (axis cs:900,115) {This work};
\addplot[densely dashed,gray] coordinates {(4,2500) (1600,6.25)};
\node[anchor=south west,font=\scriptsize,gray!70] at (axis cs:7,1200) {\(10^4\) sensing-node frames s\(^{-1}\)};
\addplot[densely dashed,gray] coordinates {(40,2500) (1600,62.5)};
\node[anchor=south west,font=\scriptsize,gray!70] at (axis cs:180,650) {\(10^5\) sensing-node frames s\(^{-1}\)};
\end{axis}
\end{tikzpicture}
\caption{\textbf{Spatiotemporal comparison of representative tactile-glove systems.} Node count and nominal acquisition rate are shown on logarithmic axes. Prior systems are grey and this work is dark coral; dashed curves indicate constant sensing-node frames per second. The 137-node capacitive array is omitted because its reported 90-ms element response time does not establish a system-level acquisition rate. Supplementary Table~\ref{supptable:glove-comparison} includes systems without a verified acquisition rate.}
\label{suppfig:glove-resolution-map}
\end{figure}
\FloatBarrier

\section*{Supplementary Note 3: Calibration platform and sensor characterization}

The sensor responded nearly linearly during both loading and unloading and showed similar sensitivity across sampled sites (main Fig.~2). Automated cycling over 0--3~N produced 1,968 complete synchronized records from 2,000 mechanical cycles. Median cycle-wise $R^2$ values over measured points within 0--3~N were 0.968 for loading and 0.962 for unloading. Across ten sites, the loading-slope coefficient of variation was 7.77\%.

\begin{figure}[!htbp]\centering
\includegraphics[height=0.38\textheight,width=0.85\textwidth,keepaspectratio]{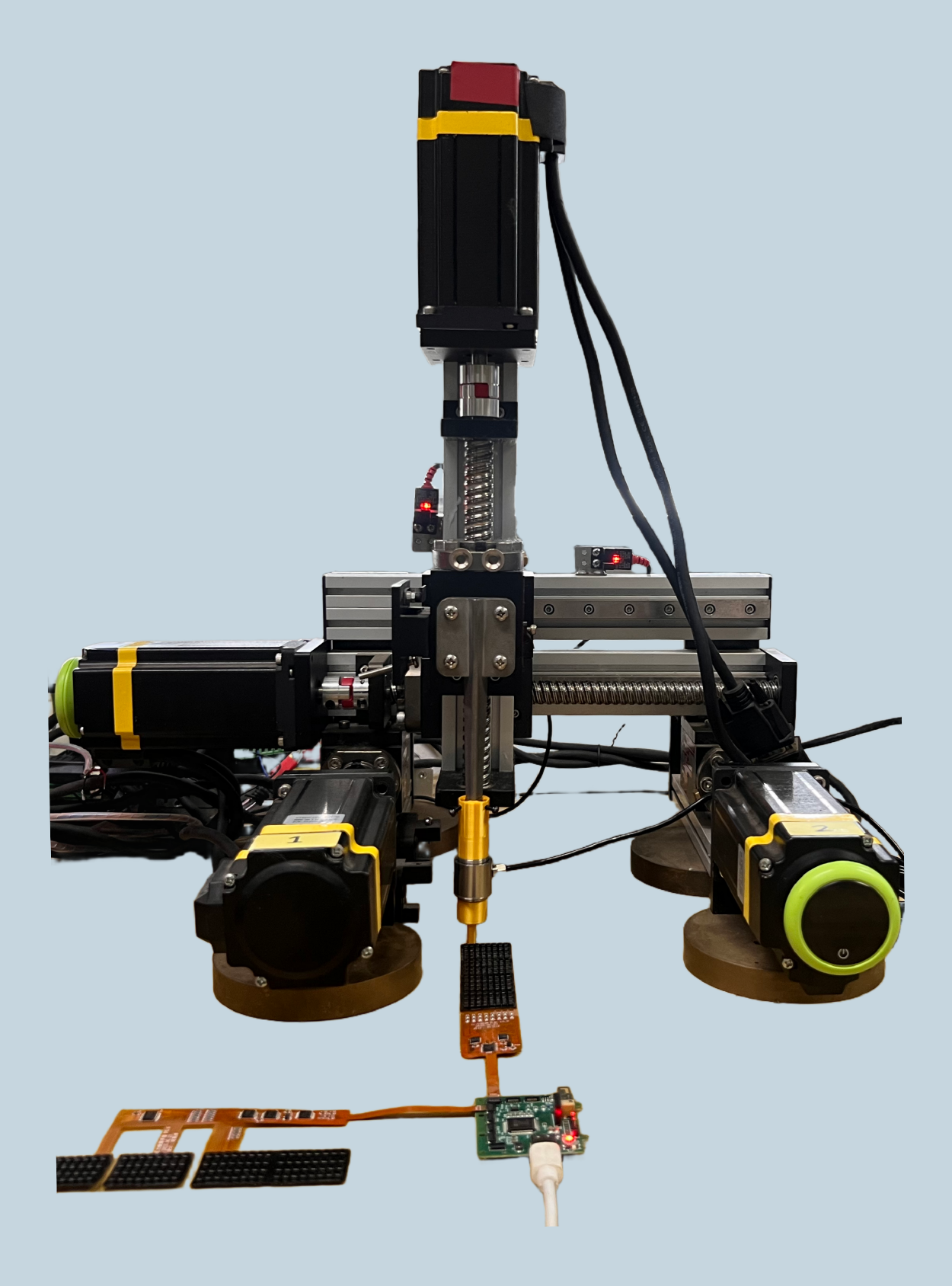}
\caption{\textbf{Normal-force calibration apparatus.} A motor-driven stage loads the tactile specimen while a reference force sensor measures applied force. Synchronized loading and unloading records provide the calibration pairs.}
\label{suppfig:sensor-performance}\end{figure}\FloatBarrier

\section*{Supplementary Note 4: Racket-control tasks}

The five tasks sampled different demands on grasp, force direction and racket-face control. Four strokes---forehand inward spinning, forehand outward spinning, backhand inward spinning and backhand outward spinning---impart spin near the net through brief racket-face adjustments around shuttle contact. The fifth, backhand rear-court transition, uses a different grip and force direction. Participant eligibility and trial criteria are given in Supplementary Methods~S2.

\section*{Supplementary Note 5: Reconstructing hand--racket contact mechanics}

To explain how hand contact affects the racket, reconstruction placed tactile nodes in a common racket-centred frame. Tactile signals and estimated pose guided the fit, while anatomical and mechanical constraints determined sensor positions and contact-normal directions. These estimates supplied contact locations, moment arms and force directions for the functional model and residual-torque analysis.

Pose estimation supplied the starting geometry for contact refinement. Separate graph--LSTM regressors predicted normalized grip position $b_{\mathrm{norm}}$ and hand--racket angle, with selected-checkpoint validation mean absolute errors of 0.0453 and $4.01^\circ$. Both used the same posture-label-stratified split: 96 training, nine validation and nine test files (seed 42). For dynamic strokes, the saved pose predictions supplied one estimate per tactile frame. Physical refinement then adjusted contact positions and normals before force and torque descriptors were calculated.

To assess reconstruction across people, we used the 50-participant dynamic cohort. For this evaluation, ten participants supplied configuration selection; the selected configuration was then frozen and evaluated in the remaining 40. This fixed-configuration assessment used the existing reconstruction-development cohort. Each participant contributed one seven-frame sequence, with the final three frames used for contact summaries. We compared complete tactile input with a condition that omitted fixed sensor regions from physical refinement; both could use complete tactile input for pose initialization (Supplementary Methods~S7).

Reconstructed active contact lay closer to the handle when local tactile evidence was available. With complete input, the response-weighted median node-to-surface distance across all active nodes was $0.652R$, where $R$ is the fixed handle inradius. The mean was $0.837R$ (95\% participant-bootstrap confidence interval, $0.750$--$0.931R$), and 32 of 40 participants had a mean below $R$. For active nodes in omitted regions, the median distance was $1.100R$ and the mean was $1.274R$ (95\% confidence interval, $1.059$--$1.519R$). The median describes typical contact agreement, while the mean and 90th percentile retain the influence of larger distances (Supplementary Table~\ref{supptable:contact-reconstruction}).

\begin{table}[!htbp]
\centering
\small
\caption{\textbf{Contact geometry in the temporal-refinement benchmark.} All distances are normalized by the fixed handle inradius $R$. Complete-input contact distances cover all active nodes; regional-omission distances cover active nodes in the omitted regions. Tactile-response weights were normalized within each valid frame; frames were weighted equally within participants, and participants equally. Medians and 90th percentiles use this same weighted distribution. Parentheses in the mean-distance row give 95\% participant-bootstrap confidence intervals for the mean. Two regional-omission frames had no above-threshold omitted-region response and supplied no contact-distance value. Penetration summaries include all 864 nodes in all 120 evaluated frames in each condition.}
\label{supptable:contact-reconstruction}
\begin{tabular}{@{}L{0.44\textwidth}>{\centering\arraybackslash}p{0.24\textwidth}>{\centering\arraybackslash}p{0.24\textwidth}@{}}
\toprule
Measure & Complete input & Regional omission \\
\midrule
Participants & 40 & 40 \\
Valid contact frames & 120 & 118 \\
Weighted median distance / $R$ & 0.652 & 1.100 \\
Weighted mean distance / $R$ & 0.837 (0.750--0.931) & 1.274 (1.059--1.519) \\
Weighted 90th-percentile distance / $R$ & 1.695 & 2.772 \\
Participants with mean distance below $R$ & 32/40 & 16/40 \\
Node observations with penetration $>0.05R$ & 0.1225\% & 0.1100\% \\
\bottomrule
\end{tabular}
\end{table}

Adding continuity terms for three-dimensional nodes and increasing the solver budget reduced mean node acceleration from 0.409 to 0.158 with complete input and from 0.992 to 0.178 with regional omission, reductions of 61.4\% and 82.0\%. Acceleration used actual frame intervals and units of $R/(10\,\mathrm{ms})^2$; it decreased in 31 and 32 of 40 participants, respectively. Temporal refinement reduced abrupt changes in reconstructed node motion while allowing overall displacement to be retained. Displacement and solver diagnostics are reported in Supplementary Methods~S7.

\begin{figure}[!htbp]
\centering
\includegraphics[width=0.98\textwidth]{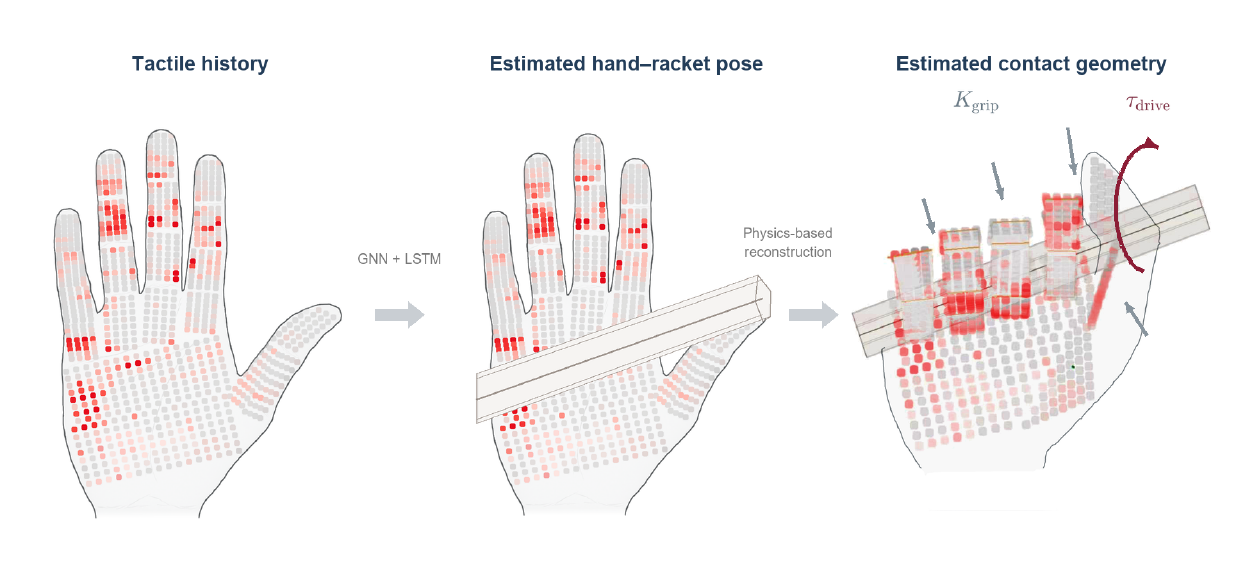}
\caption{\textbf{From tactile history to hand--racket contact mechanics.} Left, full-hand tactile input. Middle, pose estimation with a graph neural network and long short-term memory model (GNN + LSTM). Right, contact geometry after physical refinement. Grey straight arrows represent grip stabilization ($K_{\mathrm{grip}}$); red curved arrows represent racket-driving torque ($\tau_{\mathrm{drive}}$). These model variables describe the mechanical roles of contact. The example matches Supplementary Video~1.}
\label{suppfig:spatial-anchor}
\end{figure}
\FloatBarrier

\section*{Supplementary Note 6: Functional validation of drive--grip representations}

Holding a racket and moving it place different demands on drive and grip control. We compared dynamic strokes with posture-only validation sequences to assess whether the learned responses were consistent with the functional roles of motion generation and grip stabilization. Model objectives and implementation are specified in Supplementary Methods~S8.

Posture holding shifted control towards stabilization, while dynamic strokes recruited more drive activity. Across dynamic and posture validation sequences, drive-latent root-mean-square magnitude was 11.20 and 7.64, respectively, a 31.8\% reduction during posture holding. The mean grip gate was 0.726 during strokes and 0.876 during posture holding, using the same global-window definition. The sequence-averaged drive-to-total latent-norm ratio was 0.529 versus 0.232. Together, these responses distinguish movement generation from stabilization while allowing both to contribute.

\FloatBarrier
To describe the full range of responses, summaries included all dynamic validation sequences from the 50 discovery participants and all posture validation sequences from the eight additional participants in the auxiliary posture dataset. For each sequence, the drive-to-total ratio divided mean drive norm by the sum of mean drive and grip norms. Signed correlations compared their time series. Maximum absolute lagged correlations used demeaned cross-correlation across all overlapping lags, normalized by full-sequence Euclidean norms. Sequence allocation is specified in Supplementary Methods~S8.

\begin{figure}[!htbp]
\centering
\includegraphics[width=0.98\textwidth]{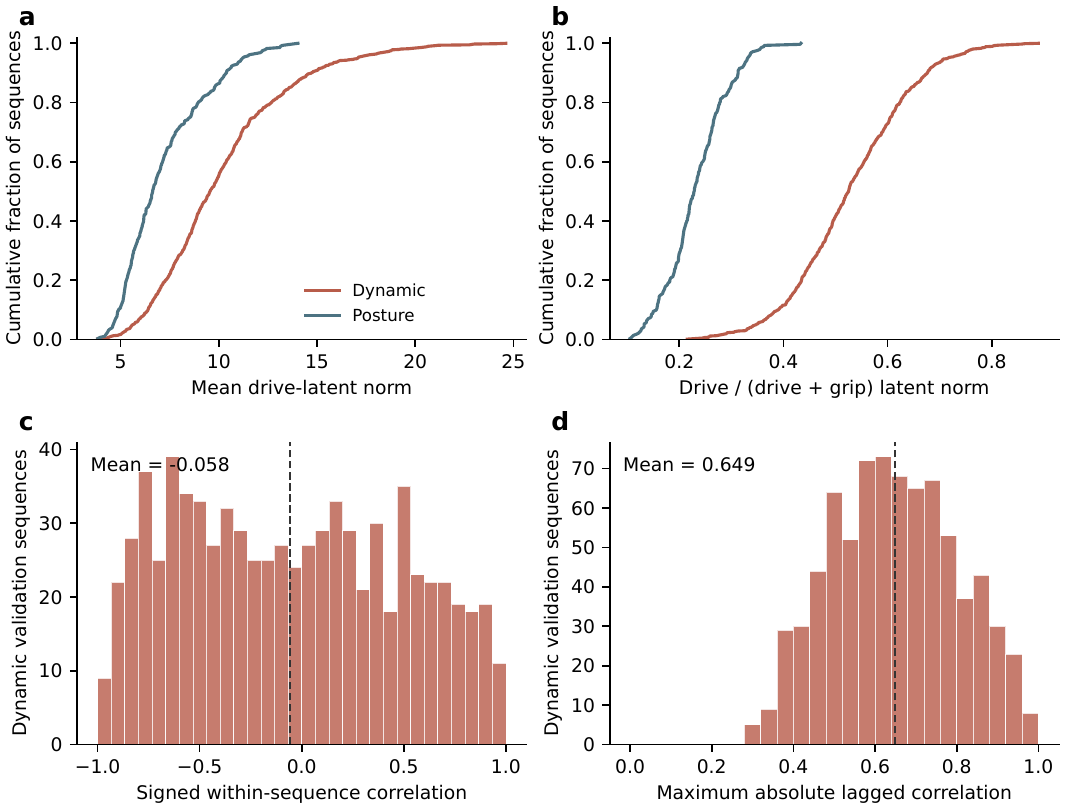}
\caption{\textbf{Drive--grip responses during movement and posture holding.} \textbf{a,b,} Empirical cumulative distributions of sequence-mean drive norm and drive-to-total norm ratio. \textbf{c,d,} Signed Pearson correlation and maximum absolute lagged correlation between drive and grip norms within dynamic validation sequences. Dashed lines mark distribution means. Each observation is a sequence; sequences from the same participant are repeated observations.}
\label{suppfig:macro-distributions}
\end{figure}
\FloatBarrier

\section*{Supplementary Note 7: Fixed-rank sensitivity of skill-specific spatial force patterns}
\label{suppnote:fixed-rank-spatial}

To test whether skill-specific spatial force organization depended on NMF rank selection, we repeated the analysis using fixed ranks of 4--10, with three independent initializations per rank.

Across all 21 rank--initialization settings, both Advanced and Expert players consistently showed greater between-skill spatial force-basis distances than Low players. A further increase from Advanced to Expert was consistently observed at ranks 5, 6 and 8, but varied at other ranks. At rank 10, held-out explained energy for Experts (12.01\%) exceeded that of the originally selected basis (11.60\%), indicating that the rank dependence was not simply due to insufficient reconstruction capacity.

These results support stronger skill-specific spatial force organization at higher proficiency, while indicating that the finer distinction between Advanced and Expert depends on the spatial basis representation.

\begin{table}[!htbp]
\centering\small
\caption{\textbf{Fixed-rank sensitivity of spatial force-basis distances.} Distances are medians over the ten skill pairs for the primary initialization. The final column counts initializations with the strict ordering Low $<$ Advanced $<$ Expert, out of three.}
\label{supptable:fixed-rank-spatial}
\setlength{\tabcolsep}{9pt}
\begin{tabular}{crrrc}
\toprule
Fixed rank $K$ & Low & Advanced & Expert & Strictly increasing initializations \\
\midrule
4 & 0.4885 & 0.7318 & 0.7121 & 0/3 \\
5 & 0.5454 & 0.6397 & 0.6922 & 3/3 \\
6 & 0.5576 & 0.6686 & 0.6706 & 3/3 \\
7 & 0.5352 & 0.6569 & 0.6497 & 1/3 \\
8 & 0.5551 & 0.6604 & 0.6835 & 3/3 \\
9 & 0.5574 & 0.6713 & 0.6669 & 0/3 \\
10 & 0.5596 & 0.6924 & 0.6746 & 0/3 \\
\bottomrule
\end{tabular}
\end{table}
\FloatBarrier

\section*{Supplementary Note 8: Spatial and temporal organization of regional coordination}
\label{suppnote:realization-roi}

Temporal coordination in skilled hand control is expressed through the recruitment of co-active contact regions. These regions provide the spatial units whose force changes can be compared in time. If a partition merges contact sites with distinct recruitment profiles, their temporal differences may be obscured by aggregation; conversely, splitting sites that act together can fragment a coherent force-recruitment pattern. Identifying regional organization from tactile data is therefore an integral part of discovering temporal coordination, rather than merely a preprocessing step.

\paragraph{Adaptive resolution under a common selection rule.}
To let contact structure determine regional detail, all proficiency groups used the same 864-node graph and parameter-selection procedure. A common normal-to-tactile weight was chosen from 0.1--0.9 in steps of 0.1. At the selected weight of 0.6, pooled Low and High groups used the same Leiden-resolution grid, 0.5--1.5 in steps of 0.1, and the same held-out-person score. Resolution was a searched parameter: the shared selection rule allowed each group to select the value best supported by its contact data, yielding 0.5 and 1.4. Each group used 50 participant-bootstrap graphs to form consensus regions. Reference maps used 29 Low and 19 High participants; later temporal and impact-aligned analyses used the expanded cohort under the eligibility rules below.

Higher-proficiency contact consistently separated into more regions under this common selection rule. Across all five consensus seeds (43--47), Low maps contained 15--16 regions and High maps 22--24 regions. The seed-45 reference partitions were used in quantitative ordering analyses; the Low/High maps displayed in main Fig.~3e contained 16 and 23 regions, respectively. Main Fig.~4d summarizes ten additional consensus initializations. The consensus-seed score was highest for seed 45 (Supplementary Methods~S10). Boundaries varied, but the difference in granularity persisted. Selection scores evaluated initial partitions, whereas final region counts followed bootstrap co-assignment and spatial consensus. These counts describe groups of co-active sensors, rather than independent physiological control units.

\begin{figure}[!htbp]\centering
\includegraphics[width=\textwidth]{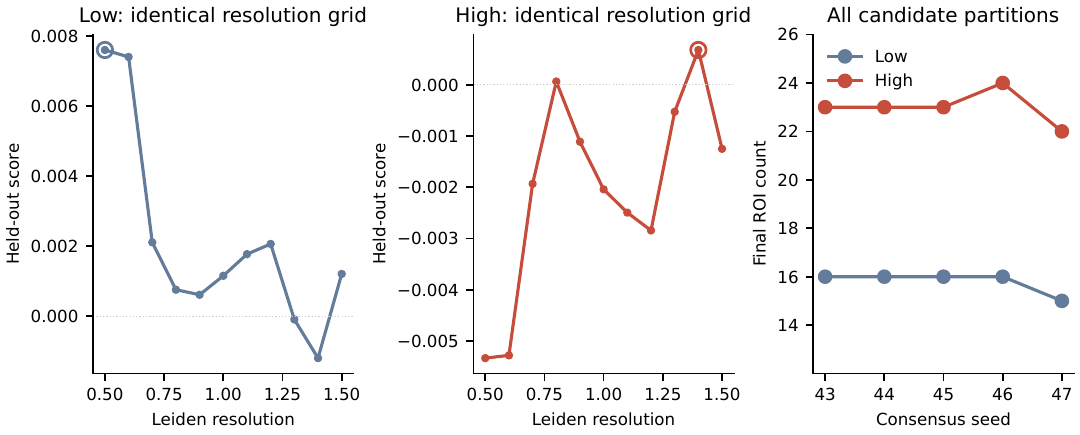}
\caption{\textbf{Regional granularity under common search rules.} Left and centre, held-out scores over the same resolution grid at a common normal-to-tactile weight of 0.6. Open circles mark the selected resolution. Right, final Low and High ROI counts for every archived consensus seed. Search scores and final consensus counts describe different stages of the partitioning procedure.}
\label{suppfig:roi-selection}\end{figure}\FloatBarrier

\begingroup\emergencystretch=1.5em
\paragraph{Regional extent and temporal organization.}
The main acquire--refine analysis used adaptive, proficiency-specific maps, allowing both region count and boundaries to follow the contact structure within each level. The selected reference partitions contained 16, 20 and 17 regions in Low, Advanced and Expert. Along the TE-derived order, mean geometric advantages over shuffled sequences were 0.083, 0.333 and 0.438 (main Fig.~4c). Region count and sequence duration describe the spatial extent and temporal span of coordination within these learned partitions.\par
\endgroup

\paragraph{Repeatability across consensus initializations.}
Main Fig.~4d summarizes ten additional consensus initializations (seeds 63--72), using the same spatial settings and participant membership. Regional sequence duration used the pooled Low, Advanced and Expert groups, retaining the original participant count in each skill--group cell and averaging 1,000 participant-bootstrap reconstructions. The acquire--refine pattern required Advanced $>$ Low and Advanced $>$ Expert, with ties excluded. It appeared in 9/10 regional partitions and 49/50 initialization--skill duration comparisons; all five skills satisfied it in 9/10 initializations. Averaging duration across five skills retained the pattern in all ten initializations (Supplementary Table~\ref{supptable:seed-regional-duration}). The sole exception was seed 67: region counts were 16/17/17, and forehand outward-spinning duration was 67.43/68.92/73.16~ms in Low/Advanced/Expert. Acquisition was retained in both cases, but refinement was not. Initialization frequencies quantify computational repeatability on the same recordings.

\begin{table}[!htbp]
\centering\scriptsize
\caption{\textbf{Regional counts and skill-specific sequence durations across ten initializations.} Entries list Low/Advanced/Expert. Region counts refer to each learned partition; durations (ms) are means of 1,000 participant-bootstrap reconstructions at the original group sizes. BH and FH denote backhand and forehand; in and out denote inward and outward spinning. The BH transition is the rear-court transition. Main Fig.~4d averages the five duration estimates within each initialization.}
\label{supptable:seed-regional-duration}
\setlength{\tabcolsep}{3pt}
\renewcommand{\arraystretch}{1.2}
\begin{tabular}{rcccccc}
\toprule
Seed & Regions & BH transition & BH in & BH out & FH in & FH out \\
\midrule
63 & 16/19/17 & 60.77/76.46/72.09 & 66.38/76.67/67.51 & 64.83/78.39/64.93 & 63.78/75.72/64.13 & 67.17/75.91/73.01 \\
64 & 16/18/17 & 61.41/72.65/72.18 & 66.57/73.50/67.09 & 64.91/76.10/64.61 & 64.54/73.35/64.48 & 67.25/74.28/72.86 \\
65 & 16/19/17 & 64.41/76.40/72.18 & 66.97/77.23/67.09 & 67.45/78.66/64.61 & 65.56/75.63/64.48 & 66.43/76.40/72.86 \\
66 & 16/19/17 & 61.33/76.13/72.03 & 64.98/77.40/64.79 & 66.28/77.91/63.47 & 64.27/75.85/63.37 & 67.08/76.18/75.46 \\
67 & 16/17/17 & 61.14/71.45/69.71 & 66.46/70.17/67.28 & 64.33/70.38/64.97 & 63.67/67.96/63.41 & 67.43/68.92/73.16 \\
68 & 16/19/17 & 60.57/76.40/68.91 & 66.41/77.23/66.30 & 64.89/78.66/64.77 & 63.35/75.63/63.68 & 67.24/76.40/73.97 \\
69 & 16/19/17 & 62.19/76.35/69.31 & 66.01/77.73/66.42 & 64.47/78.21/64.52 & 63.13/75.60/63.56 & 65.97/76.80/74.50 \\
70 & 16/19/17 & 61.33/76.35/72.03 & 64.98/77.73/64.79 & 66.28/78.21/63.47 & 64.27/75.60/63.37 & 67.08/76.80/75.46 \\
71 & 16/18/17 & 60.77/72.62/71.79 & 66.38/74.04/66.27 & 64.83/75.37/64.82 & 63.78/73.70/63.84 & 67.17/74.28/73.59 \\
72 & 16/19/17 & 60.80/76.06/66.40 & 66.73/77.47/63.80 & 65.02/78.45/66.07 & 63.37/75.38/66.30 & 66.95/76.70/75.12 \\
\bottomrule
\end{tabular}
\end{table}\FloatBarrier

\paragraph{Fixed-count and anatomical controls.}
To separate the effects of region count and boundary placement, we compared two fixed-count representations. At each integer count $K=15,\ldots,25$, fixed-count, level-specific maps learned boundaries separately within Low, Advanced and Expert; fixed-count, pooled maps learned one set of boundaries jointly from all levels. These conditions constrain region count, unlike the adaptive, proficiency-specific maps in the main analysis. Spatial fitting included all 50 participants (31 Low, 11 Advanced and 8 Expert), with equal participant weights. Both procedures used the same 864 nodes, normal-to-tactile weight 0.6, fixed Leiden resolution 0.6 and 50 bootstrap partitions. Spatially adjacent regions were joined by node-weighted mean co-assignment, retaining the existing finger-segment grouping constraints and a minimum region size of five nodes; cuts of the resulting hierarchy gave each exact $K$. Proficiency labels defined membership only for the separate-map fits and did not enter the pooled fitting objective. No TE endpoint was used to select boundaries or $K$.

The fixed anatomical map used 15 regions: three segments per non-thumb finger, two thumb segments and one palm region, as defined in Supplementary Note~10. Only the anatomical definition was reused; temporal outcomes were recomputed on the same recordings as the two fixed-count controls, rather than taken from the held-out-trial comparison in that note. All three controls retained the same 207 participant--skill combinations after requiring at least seven strokes per combination. Spatial fitting and temporal eligibility therefore involved different participant counts. TE used 100-Hz data, delays of 10--50~ms, 200 within-stroke circular source shifts, link-wise FDR below 0.05 and participant recurrence of at least 70\%. Regional signal normalization, temporal ordering and torque geometry were held constant across conditions.

\paragraph{Dependence of the acquire--refine pattern on spatial representation.}
To compare spatial representations under matched sampling, we re-evaluated the adaptive reference and two shared-reference maps using the same 1,000 whole-participant bootstrap samples as the fixed-count and anatomical controls. All comparisons retained the same 207 eligible participant--skill observations and sequence-duration estimator.

Under this matched protocol, the adaptive, proficiency-specific maps (16/20/17 regions) recovered an Advanced sequence-duration peak across all five skills, with a mean peak margin of +11.06~ms. Fixed-count learned maps recovered the peak in two to four skills across the tested counts, while the anatomical map recovered it in four. Notably, a shared map derived from Advanced players also recovered the peak in all five skills, although with a smaller mean margin (+3.51~ms). The remaining comparisons are summarized in Supplementary Table~\ref{supptable:partition-scan-all-k}.

\begin{table}[!htbp]
\centering\footnotesize
\caption{\textbf{Matched spatial-representation comparisons and fixed-count sensitivity.} The upper section compares all representations under the same 1,000 whole-participant bootstrap samples. Bootstrap-mean peaks count skills with Advanced strictly above both Low and Expert; ties do not count. Mean peak margin is Advanced minus the larger of Low and Expert, calculated from bootstrap-mean durations within each skill and averaged across all five skills. For fixed-count summaries, peak counts are ranges across $K=15,\ldots,25$ and margins average equally over these 11 counts. In the lower section, each paired cell reports level-specific $\mid$ pooled maps at the same fixed region count, with boundaries learned separately by level or jointly across levels. Boot denotes the mean over bootstrap graphs. The last two columns report participant-level High minus Low differences, where High pools Advanced and Expert. Chain advantage is standardized against shuffled orders; alignment is a mean cosine. Counts summarize descriptive peak recovery.}
\label{supptable:partition-scan-all-k}
\setlength{\tabcolsep}{4pt}
\renewcommand{\arraystretch}{1.15}
\begin{tabular}{@{}L{0.44\textwidth}cc@{}}
\toprule
Spatial representation & \shortstack{Bootstrap-mean\\peaks / 5} & \shortstack{Mean peak\\margin (ms)} \\
\midrule
Adaptive, proficiency-specific (16/20/17) & 5 & $+11.06$ \\
Fixed-count, level-specific ($K=15$--$25$) & 2--4 & $+1.95$ \\
Fixed-count, pooled ($K=15$--$25$) & 2--4 & $+0.80$ \\
Anatomical (15 regions) & 4 & $+2.74$ \\
Shared, Low-derived (16 regions) & 3 & $+2.38$ \\
Shared, Advanced-derived (20 regions) & 5 & $+3.51$ \\
\bottomrule
\end{tabular}
\par\vspace{8pt}
\begin{tabular}{rccc}
\toprule
\multicolumn{4}{c}{Fixed-count level-specific $\mid$ pooled; peak counts are out of five skills} \\
\addlinespace
$K$ & Duration peaks & Chain advantage & Alignment \\
 & (boot) & (High--Low) & (High--Low) \\
\midrule
15 & 4 $\mid$ 4 & $-0.0052$ $\mid$ $+0.0618$ & $-0.0341$ $\mid$ $+0.0101$ \\
16 & 3 $\mid$ 3 & $+0.0966$ $\mid$ $+0.0767$ & $+0.0051$ $\mid$ $+0.0147$ \\
17 & 4 $\mid$ 3 & $+0.1941$ $\mid$ $+0.1845$ & $+0.0719$ $\mid$ $+0.0187$ \\
18 & 4 $\mid$ 4 & $+0.1619$ $\mid$ $+0.1132$ & $+0.0486$ $\mid$ $+0.0198$ \\
19 & 4 $\mid$ 3 & $+0.1994$ $\mid$ $+0.0444$ & $+0.0326$ $\mid$ $+0.0094$ \\
20 & 4 $\mid$ 4 & $+0.1319$ $\mid$ $+0.0056$ & $+0.0003$ $\mid$ $+0.0192$ \\
21 & 4 $\mid$ 4 & $+0.1070$ $\mid$ $+0.0313$ & $+0.0263$ $\mid$ $+0.0146$ \\
22 & 4 $\mid$ 4 & $+0.0564$ $\mid$ $-0.0103$ & $+0.0237$ $\mid$ $+0.0148$ \\
23 & 3 $\mid$ 3 & $+0.0733$ $\mid$ $+0.0049$ & $+0.0341$ $\mid$ $+0.0088$ \\
24 & 4 $\mid$ 2 & $+0.1201$ $\mid$ $-0.1109$ & $+0.0180$ $\mid$ $+0.0123$ \\
25 & 2 $\mid$ 3 & $+0.1034$ $\mid$ $-0.1248$ & $+0.0212$ $\mid$ $+0.0009$ \\
\bottomrule
\end{tabular}
\end{table}\FloatBarrier

\paragraph{Retention of mechanical-coherence trends.}
For fixed-count, pooled maps, High-minus-Low chain-level torque advantage was positive at eight of 11 counts and step-to-step alignment at all 11. Expert-minus-Low differences were positive at eight and ten counts, respectively. For fixed-count, level-specific maps, the corresponding High-minus-Low directions were positive at ten counts for each metric. On the anatomical map, mean chain advantages relative to shuffled orders were 0.332/0.320/0.652 in Low/Advanced/Expert, whereas mean step alignment was $-0.179/-0.162/-0.141$. Experts thus retained higher mean values on both measures; chain advantage did not increase monotonically across the three groups. The anatomical Expert-minus-Low contrasts were 0.321 (95\% CI 0.059 to 0.615) for chain advantage and 0.037 (0.017 to 0.059) for alignment. These are secondary, unadjusted intervals for the Expert-minus-Low comparison.

\paragraph{Participant-level inference and scope of the controls.}
We used 1,000 whole-participant bootstrap resamples within levels, retaining each person's available skills and sharing draws across counts and map conditions. Learned maps were held fixed. Graphs and sequences were reconstructed in each resample. For duration, the tested effect was the skill average of Advanced minus the larger of Low and Expert. For geometric outcomes, skills were averaged within participant before contrasting pooled High with Low. The pooled-map scan averaged each effect over all 11 counts, which were treated as dependent sensitivity conditions rather than independent replicates.

For pooled and anatomical maps, 999 whole-participant proficiency-label permutations preserved group sizes and supplied conditional tests; Holm-adjusted values retain the original four-endpoint family. None of the three reported endpoints passed this correction (Supplementary Table~\ref{supptable:partition-control-inference}). Across pooled-map tests at individual counts, no positive effect survived Benjamini--Hochberg correction either. Fixed-count level-specific-minus-pooled contrasts and their paired intervals are reported alongside the endpoint tests in Supplementary Table~\ref{supptable:partition-control-inference}. At matched region counts, the scan-averaged comparisons did not establish a significant advantage of fixed-count, level-specific over pooled maps (Supplementary Table~\ref{supptable:partition-control-inference}). Label permutations were restricted to maps with common node identities across levels; comparisons involving level-specific maps used paired participant-bootstrap intervals. Permutation tests used complete-sample graph estimates. Bootstrap-mean peak counts were summarized separately because graph thresholding and tree reconstruction are nonlinear.

\begin{table}[!htbp]
\centering\footnotesize
\caption{\textbf{Effect sizes and uncertainty under alternative regional representations.} For duration, the effect is Advanced minus the larger of Low and Expert, calculated within skill before averaging across the five skills. For the geometric endpoints, the effect is High minus Low after averaging available skills within participant. Fixed-count pooled-map effects and level-specific-minus-pooled differences additionally average over $K=15,\ldots,25$; the anatomical map has $K=15$. Intervals are 95\% participant-bootstrap percentile intervals conditional on fixed maps. Permutation tests are one-sided for duration and two-sided for geometric effects; Holm-adjusted values retain the original four-endpoint family within each map analysis. Paired level-specific-minus-pooled contrasts use bootstrap intervals. For duration, these inferential comparisons use complete-sample graph estimates; bootstrap-mean duration sensitivity is summarized in Supplementary Table~\ref{supptable:partition-scan-all-k}.}
\label{supptable:partition-control-inference}
\setlength{\tabcolsep}{5pt}
\renewcommand{\arraystretch}{1.1}
\begin{tabular}{lrc rr}
\toprule
Endpoint & Effect & 95\% CI & $P$ & $P_{\rm Holm}$ \\
\midrule
\multicolumn{5}{l}{\textit{Fixed-count, pooled maps: all 11 counts}} \\
Sequence duration (ms) & $-10.73$ & $[-19.45,\ +6.00]$ & 0.652 & 1.000 \\
Chain advantage ($z$) & $+0.0251$ & $[-0.2333,\ +0.2962]$ & 0.857 & 1.000 \\
Step alignment & $+0.0130$ & $[-0.0046,\ +0.0315]$ & 0.223 & 0.892 \\
\addlinespace
\multicolumn{5}{l}{\textit{Fixed anatomical map, 15 regions}} \\
Sequence duration (ms) & $-16.00$ & $[-24.00,\ +14.00]$ & 0.822 & 0.880 \\
Chain advantage ($z$) & $+0.1283$ & $[-0.1565,\ +0.4447]$ & 0.440 & 0.880 \\
Step alignment & $+0.0253$ & $[+0.0016,\ +0.0502]$ & 0.065 & 0.260 \\
\addlinespace
\multicolumn{5}{l}{\textit{Fixed-count, level-specific minus pooled maps: all 11 counts}} \\
Sequence duration (ms) & $+1.82$ & $[-10.73,\ +12.91]$ & -- & -- \\
Chain advantage ($z$) & $+0.0875$ & $[-0.0175,\ +0.1896]$ & -- & -- \\
Step alignment & $+0.0095$ & $[-0.0030,\ +0.0211]$ & -- & -- \\
\addlinespace
\bottomrule
\end{tabular}
\end{table}\FloatBarrier

Together, these comparisons show that spatial representation influences how clearly the acquire--refine pattern can be resolved. Adaptive, proficiency-specific regions yielded the strongest and most consistent pattern, whereas anatomical, fixed-count and shared representations generally weakened its magnitude or consistency across skills. This suggests that imposing spatial partitions can merge distinct force-recruitment patterns and obscure their temporal relationships, highlighting the value of learning co-active regions separately across proficiency levels.

\paragraph{Impact-aligned realization.}
To determine when the whole hand redistributed contact, we aligned complete sensor records to impact at frame 50 (one-based) and used a common $-200$ to $+200$~ms interval. Each frame's 864-node contact vector was divided by its sum, then averaged across strokes within participant and skill. Spatial reconfiguration was half the $L_1$ distance between consecutive normalized vectors, assigned to the midpoint between frames. Its temporal centroid gives the average time of redistribution; its pre-impact share gives the fraction occurring before contact with the shuttle. Normalization prevents large overall forces from dominating these timing measures.

To locate the timing of local adjustments within the same contact pattern, we applied the 16-region Low map to every participant. Regional mean forces were normalized across regions in each frame and averaged across strokes. Recruitment-peak time was the midpoint of the largest positive frame-to-frame change in a region's share within the same interval. Each participant--skill summary contained the median regional peak time and the fraction of regions peaking before impact. We averaged skills within participants and used 10,000 stratified participant-bootstrap samples for percentile 95\% intervals. All five skills and all 50 dynamic participants were retained.

Experts redistributed whole-hand contact earlier, while local adjustments continued after impact. Mean spatial centroids were 19.0, 14.1 and 8.6~ms after impact in Low, Advanced and Expert, with pre-impact shares of 41.6\%, 44.5\% and 47.8\%. Centroids advanced across both adjacent proficiency comparisons in every skill. The fractions of regions with pre-impact recruitment peaks were 35.8\%, 39.1\% and 39.2\%, but median peak times were 45.6, 38.0 and 55.6~ms after impact. Thus, earlier overall preparation coexisted with later local changes. These impact-aligned measures locate force redistribution within the stroke; TE completion time measures the span of the inferred regional sequence.

\begin{figure}[!htbp]\centering
\includegraphics[width=\textwidth]{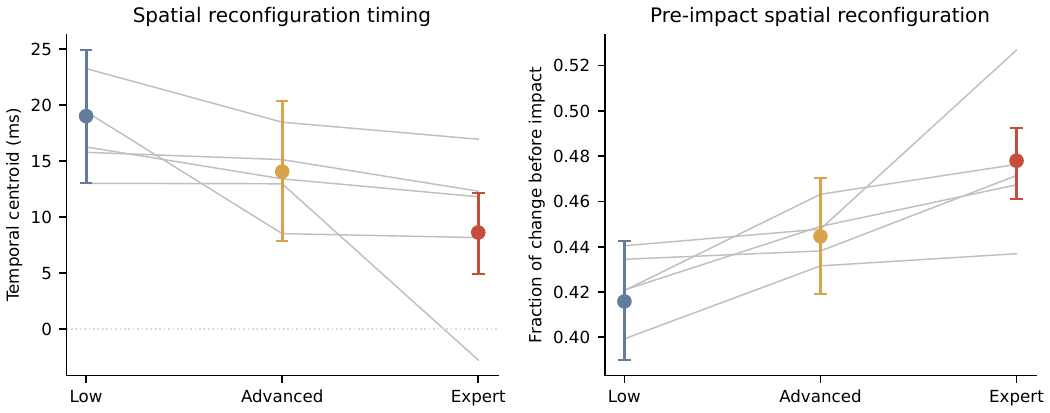}
\caption{\textbf{Whole-hand contact redistribution relative to impact.} Left, average timing of contact redistribution. Right, fraction of redistribution occurring before impact. Grey lines show skills; coloured points and bars show participant-weighted means and bootstrap 95\% intervals (31 Low, 11 Advanced and 8 Expert).}
\label{suppfig:impact-realization}\end{figure}\FloatBarrier

\section*{Supplementary Note 9: Model temporal weighting and sensitivity}
\label{suppnote:lag-robustness}

Higher-proficiency control placed more weight on recent tactile history (main Fig.~3f; Spearman ordered-group trend $P=1.67\times10^{-5}$). Mean expected lags were 3.927, 3.815 and 3.791 frames in Low (beginner and intermediate), Advanced and Expert participants. Normalized temporal entropies were 0.694, 0.647 and 0.635, indicating more concentrated weighting at higher proficiency. Expected lag averages the delays weighted by the model within its 3--8-frame history window; entropy measures the spread of those weights. To assess sensitivity, we varied initialization, optimization, model capacity and sampling while retaining the 7,736-stroke discovery dataset, the auxiliary posture sequences and their sequence-level split (Supplementary Methods~S9).

\paragraph{Initialization and optimization.}
The shorter Expert interval persisted across attention-fitting choices. Eight refits held the rest of the model fixed while varying initialization, optimization and weighting settings. Expert-minus-Low contrasts were negative in all eight validation-selected fits, ranging from -3.58 to -1.86~ms. Complete settings and confidence intervals are reported with the fitting protocol in Supplementary Methods~S9 (Supplementary Table~\ref{supptable:lag-attention}). These refits preserved the direction of the proficiency contrast while changing its estimated size and uncertainty.

\paragraph{Representation and sampling.}
The same direction persisted when we changed representation size and sampling rate. Refitting the frame-wise projection, drive/grip heads and decoder at latent dimensions 4, 8 and 16, with three seeds per dimension and a fixed encoder, gave negative Expert-minus-Low contrasts in 9/9 fits. A separate test used the same history times (40, 60 and 80~ms) at 100~Hz and both phases of 50-Hz decimation, with three initializations per condition; all 9/9 contrasts were negative. Both rates used the same causal prefilter, physical interval and torque target, with motion-context differences expressed in common time units. This matched-rate test is reported separately from the native-data capacity test (Supplementary Table~\ref{supptable:lag-capacity}).

Across all 26 conditions, Expert model profiles placed more weight on recent tactile history than Low model profiles; 21 contrasts had 95\% confidence intervals below zero. These conditions reuse the same participants and provide unadjusted sensitivity intervals. Individual temporal weights also reduced validation prediction loss relative to uniform weights by 0.0395 [0.0320, 0.0475], shared global weights by 0.0039 [0.0028, 0.0050], and shuffled profiles by 0.0008 [0.0004, 0.0013]. Together, these tests link the temporal-weight pattern to fitted control prediction. The interval describes the decoder's use of tactile history; it is not a physiological delay measurement.

\FloatBarrier

\begin{table}[!htbp]
\centering\small
\caption{\textbf{Capacity and sampling sensitivity.} Each row is a separately trained projection/control model followed by attention fitting. Native conditions use lags 3--8 at 100~Hz; matched sampling conditions use 40, 60 and 80~ms. Epoch denotes the selected attention epoch. All rows use the same 50 dynamic participants for lag summaries.}
\label{supptable:lag-capacity}
\begin{tabular}{L{0.42\textwidth}rL{0.38\textwidth}}
\toprule Condition / dimension / seed & Epoch & Expert--Low lag (ms), 95\% CI \\
\midrule
matched100\_d8\_s41 & 31 & -0.95 [-1.69, -0.37] \\
matched100\_d8\_s42 & 32 & -0.76 [-1.33, -0.28] \\
matched100\_d8\_s43 & 32 & -1.00 [-1.81, -0.35] \\
native\_d16\_s41 & 32 & -0.75 [-1.38, -0.27] \\
native\_d16\_s42 & 32 & -0.96 [-1.74, -0.38] \\
native\_d16\_s43 & 32 & -1.13 [-1.89, -0.51] \\
native\_d4\_s41 & 32 & -1.09 [-1.90, -0.47] \\
native\_d4\_s42 & 32 & -0.89 [-1.49, -0.41] \\
native\_d4\_s43 & 32 & -0.95 [-1.70, -0.34] \\
native\_d8\_s41 & 32 & -1.22 [-2.33, -0.40] \\
native\_d8\_s42 & 32 & -1.10 [-1.95, -0.43] \\
native\_d8\_s43 & 32 & -1.24 [-2.08, -0.57] \\
rate50even\_d8\_s41 & 31 & -0.91 [-1.68, -0.32] \\
rate50even\_d8\_s42 & 32 & -0.68 [-1.18, -0.23] \\
rate50even\_d8\_s43 & 32 & -0.91 [-1.65, -0.32] \\
rate50odd\_d8\_s41 & 31 & -0.98 [-1.79, -0.33] \\
rate50odd\_d8\_s42 & 32 & -0.83 [-1.45, -0.31] \\
rate50odd\_d8\_s43 & 32 & -0.91 [-1.61, -0.33] \\
\bottomrule
\end{tabular}
\end{table}
\FloatBarrier
Exploratory tests at 25, 20, 12.5 and 10~Hz did not resolve a further change: all five paired-change 95\% intervals relative to matched 100-Hz controls included zero. Full comparisons are retained in the analysis records. These tests concern model history weighting; Supplementary Note~13 examines sampling of regional coordination.

\section*{Supplementary Note 10: Data-derived and anatomical regions}
\label{suppnote:anatomical-roi}

We compared three representations of the same 864 tactile nodes: a fixed anatomical map with 15 regions, a learned map merged to 15 regions, and the native adaptive learned map. The anatomical definition followed Sundaram et al.\cite{sundaram2019}, with three segments per non-thumb finger, two thumb segments and one palm region. The comparison used 7,736 strokes from 50 participants. Maps were fitted using 24 participants and evaluated on 26 held-out participants; reversing these roles provided a sensitivity analysis on the same cohort (Supplementary Methods~S11).

\begin{figure}[!htbp]
\centering
\includegraphics[width=\textwidth]{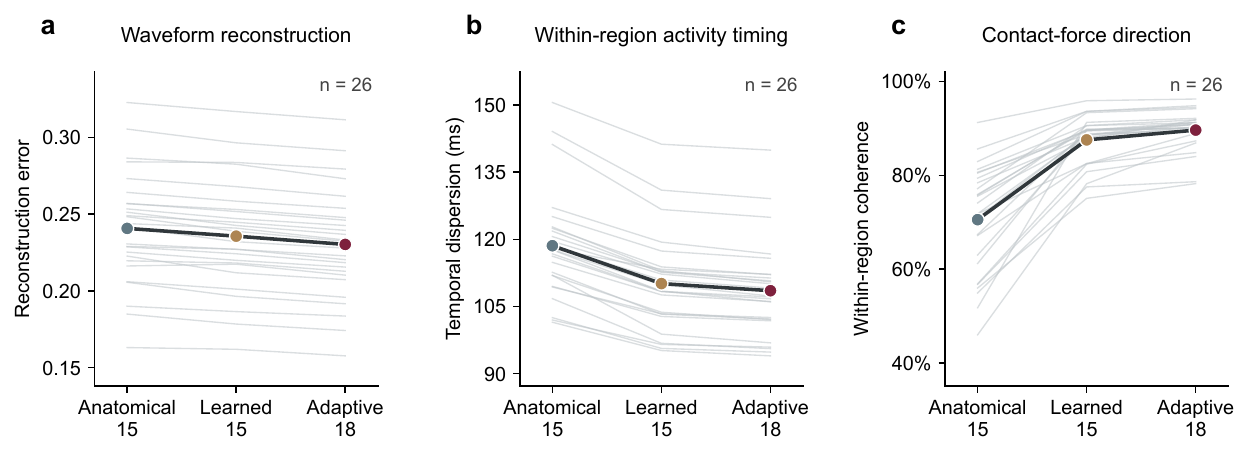}
\caption{\textbf{Regional force representation with anatomical and learned partitions.} \textbf{a,} Whole-hand normalized waveform reconstruction error. \textbf{b,} Within-region dispersion of palm-node force-activity time centroids. \textbf{c,} Within-region coherence of reconstructed palm contact-force directions. Anatomical and count-matched learned maps contain 15 regions; native adaptive maps contain 18 in this split. Pale lines connect the same held-out participants across maps; dark lines and coloured points show means ($n=26$). Maps were fitted on the other 24 participants. All pairwise comparisons passed the joint Holm-adjusted tests ($P_{\rm Holm}=3.4\times10^{-4}$). Reversed-split results and paired intervals are reported in Supplementary Table~\ref{supptable:anatomical-primary}.}
\label{suppfig:anatomical-primary}
\end{figure}

\paragraph{Regional allocation and representation.}
The adaptive maps contained 18 regions in the original split and 17, 19 and 18 regions for Low, Advanced and Expert training groups in the reversed split. Whereas the anatomical map represented the palm as one region, adaptive maps retained eight palm subregions and learned 15-region maps retained seven or eight. We assessed the resulting representations through whole-hand waveform reconstruction, within-region palm activity timing and within-region palm contact-force direction (Supplementary Fig.~\ref{suppfig:anatomical-primary}; Supplementary Table~\ref{supptable:anatomical-primary}).

\paragraph{Three-way quantitative comparison.}
In the original held-out split, mean normalized waveform error was 0.24069 for anatomical regions, 0.23563 for learned 15-region maps and 0.23025 for adaptive maps. Relative to anatomical regions, error decreased by 2.10\% and 4.34\%, respectively. Palm activity-time dispersion decreased from 118.56~ms to 110.11 and 108.52~ms, corresponding to reductions of 7.13\% and 8.47\%. Palm force-direction coherence increased from 70.53\% to 87.54\% and 89.63\%, gains of 17.01 and 19.10 percentage points. All three pairwise contrasts for each metric passed the two-sided participant-level test after Holm adjustment ($P_{\rm Holm}=3.4\times10^{-4}$).

In the reversed split, adaptive maps reduced waveform error by 5.03\% and activity-time dispersion by 8.44\%, and increased force-direction coherence by 17.50 percentage points relative to anatomical regions (all $P_{\rm Holm}=3.4\times10^{-4}$). Adaptive maps also improved these metrics relative to learned 15-region maps ($P_{\rm Holm}=3.4\times10^{-4}$, 0.0478 and 0.0479, respectively). These comparisons describe the combined effects of regional allocation, boundary placement and adaptive count; subdivision itself favours these representation metrics.

The aim of contact-based partitioning is to explore force-coactivation patterns across proficiency levels with reduced reliance on predefined anatomical boundaries, and to establish data-derived coactive regions as a spatial basis for analysing temporal coordination. The earlier within-participant held-out-trial illustration in Supplementary Fig.~\ref{suppfig:anatomical-palm} shows palm waveform differences obscured by whole-palm averaging. Its selected regions and peak-normalized waveforms describe within-group temporal profiles. Regional sequence-duration comparisons are reported separately in Supplementary Note~8.

\begin{table}[!htbp]
\centering\footnotesize
\caption{\textbf{Anatomical, count-matched learned and adaptive learned representations.} Entries are participant means after averaging available skills within each participant. The change and its 95\% participant-bootstrap interval compare adaptive with anatomical regions. Coherence and its change are expressed in percent and percentage points, respectively. Lower error and dispersion, and higher coherence, indicate greater within-region fidelity or consistency. Two-sided sign-flip tests use the joint Holm correction specified in Supplementary Methods~S11.}
\label{supptable:anatomical-primary}
\setlength{\tabcolsep}{2.5pt}
\renewcommand{\arraystretch}{1.12}
\begin{tabular}{L{0.19\textwidth}rrrL{0.27\textwidth}r}
\toprule
Metric & Anatomical 15 & Learned 15 & Adaptive & Change [95\% CI] & $P_{\rm Holm}$ \\
\midrule
\multicolumn{6}{l}{\textit{Original split: 24 fitting participants, 26 evaluation participants; adaptive $K=18$}} \\
Waveform error & 0.24069 & 0.23563 & 0.23025 & $-0.01045$ [-0.01156, -0.00930] & $3.4\times10^{-4}$ \\
Timing dispersion (ms) & 118.56 & 110.11 & 108.52 & $-10.04$ [-11.01, -9.14] & $3.4\times10^{-4}$ \\
Force coherence (\%) & 70.53 & 87.54 & 89.63 & $+19.10$ [+15.78, +22.62] & $3.4\times10^{-4}$ \\
\midrule
\multicolumn{6}{l}{\textit{Reversed split: 26 fitting participants, 24 evaluation participants; adaptive $K=17/19/18$}} \\
Waveform error & 0.23904 & 0.23090 & 0.22702 & $-0.01202$ [-0.01438, -0.00982] & $3.4\times10^{-4}$ \\
Timing dispersion (ms) & 118.14 & 108.74 & 108.18 & $-9.97$ [-11.25, -8.82] & $3.4\times10^{-4}$ \\
Force coherence (\%) & 75.96 & 93.09 & 93.46 & $+17.50$ [+13.95, +22.00] & $3.4\times10^{-4}$ \\
\bottomrule
\end{tabular}
\end{table}

\FloatBarrier

\begin{figure}[!htbp]
\centering
\includegraphics[width=\textwidth]{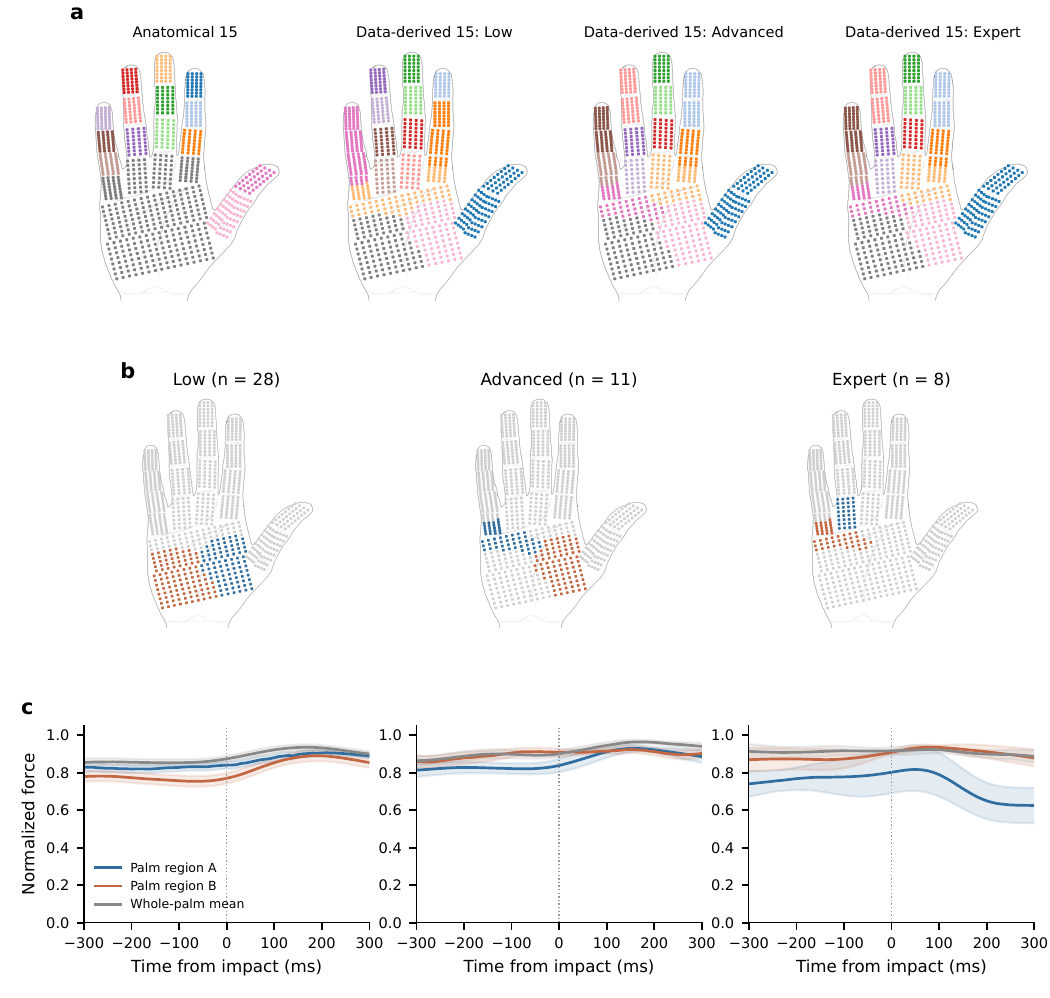}
\caption{\textbf{Local palm dynamics obscured by whole-palm averaging.} \textbf{a,} Anatomical and data-derived 15-region maps. Colours identify regions within each map. The anatomical map treats the palm as one region. \textbf{b,} Two adjacent palm subregions selected separately for each proficiency group using training trials of backhand inward spinning. Selection maximized the training waveform contrast among adjacent pairs containing at least 16 palm nodes each. \textbf{c,} Corresponding held-out waveforms and the whole-palm mean (Low, $n=28$; Advanced, $n=11$; Expert, $n=8$). Each curve was normalized to its own peak within participant before averaging. Blue and orange curves correspond to the subregions in b; grey shows the separately normalized whole-palm mean. Lines and shading show participant means and SEM; zero marks impact. Comparisons concern waveform shape within each group: selected locations differ across groups, and amplitudes do not represent absolute force.}
\label{suppfig:anatomical-palm}
\end{figure}
\FloatBarrier

\section*{Supplementary Note 11: Spatial sampling of local contact structure}

Dense sampling preserved local contact boundaries that became blurred when fewer sensors were available. Relative to the 864-node reference, boundary recall and preservation of small contact regions were 0.574 and 0.945 at 432 nodes, falling to 0.123 and 0.348 at 108 nodes. Thus, boundaries could shift even while many small regions remained distinguishable.

To measure this loss, we reduced node count and evaluated held-out co-activation modularity, boundary recall and preservation of small regions of interest (ROIs). The analysis used 7,736 strokes from 50 participants (Supplementary Methods~S3). Reference partitions and their training subsets were fixed, and the remaining recordings supplied evaluation data. Each node count used five subsampling repeats, each spanning 21 normal-force weights (0--1 in steps of 0.05) and nine resolution settings (1--5 in steps of 0.5), for 945 parameter settings. We averaged the 189 settings within each repeat, then reported the mean and standard deviation across repeats (Supplementary Table~\ref{supptable:spatial-retention}).

Regional temporal-sampling results are reported in Supplementary Note~13.

\begin{figure}[!htbp]\centering
\includegraphics[width=\textwidth]{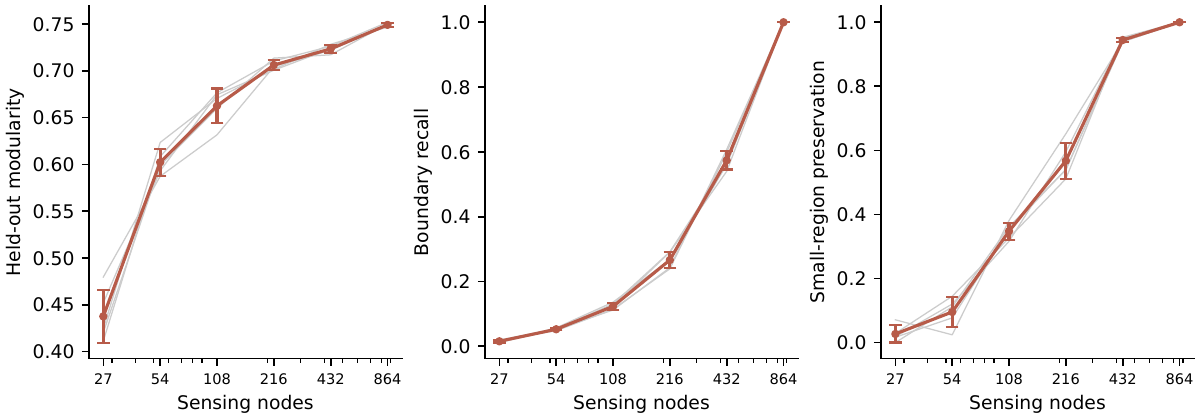}
\caption{\textbf{Spatial retention of contact structure.} Held-out co-activation modularity, boundary recall and fine-scale ROI preservation across node counts. Thin lines show the five subsampling repeats; points and bars show their means and standard deviations after averaging the 189 parameter settings within each repeat (Supplementary Table~\ref{supptable:spatial-retention}).}
\label{suppfig:resolution-necessity}\end{figure}
\FloatBarrier

\begin{table}[!htbp]
\centering\small
\caption{\textbf{Spatial retention summarized across five subsampling repeats.} Entries are mean $\pm$ standard deviation of repeat-level means, each averaged over 189 parameter cells. These standard deviations describe subsampling-repeat variation.}
\label{supptable:spatial-retention}
\begin{tabular}{rccc}
\toprule
Nodes & Modularity & Boundary recall & Fine-scale preservation \\
\midrule
27 & 0.438 $\pm$ 0.028 & 0.016 $\pm$ 0.004 & 0.027 $\pm$ 0.027 \\
54 & 0.602 $\pm$ 0.014 & 0.053 $\pm$ 0.003 & 0.096 $\pm$ 0.047 \\
108 & 0.663 $\pm$ 0.018 & 0.123 $\pm$ 0.010 & 0.348 $\pm$ 0.026 \\
216 & 0.706 $\pm$ 0.006 & 0.266 $\pm$ 0.025 & 0.568 $\pm$ 0.056 \\
432 & 0.723 $\pm$ 0.004 & 0.574 $\pm$ 0.028 & 0.945 $\pm$ 0.007 \\
864 & 0.749 $\pm$ 0.002 & 1.000 $\pm$ 0.000 & 1.000 $\pm$ 0.000 \\
\bottomrule
\end{tabular}
\end{table}

\FloatBarrier

\section*{Supplementary Note 12: Spatial-resolution sensitivity of residual metrics}
\label{suppnote:residual-resolution}

Reconstructing residual patterns from sparse tactile inputs differed from sparsely reading out patterns already inferred at full resolution. To test how spatial downsampling affected proficiency-related force organization, we recomputed the four residual metrics in main Fig.~4b at 864, 432, 216, 108, 54 and 27 nodes. All resolutions used the same training participants, held-out comparisons and skill-specific ranks, with three initialization seeds per skill (Supplementary Methods~S13). We distinguished sparse-input refits, which reduce the raw tactile observations before refitting, from subsampling a residual field already inferred at full resolution. Sparse-input refits in Supplementary Figs.~\ref{suppfig:residual-resolution-refit} and \ref{suppfig:residual-resolution-smoothing} retained the original spatial and temporal smoothing. Sensitivity results omitting spatial smoothing are retained in Supplementary Table~\ref{supptable:residual-resolution-skills} and the analysis records. Main Fig.~4b corresponds to the full-resolution residual-field baseline, whereas each refit condition uses its own re-estimated 864-node baseline averaged across three seeds. We counted decreases from Low to Advanced and Advanced to Expert within each skill, for ten planned contrasts. Ties and unavailable comparisons did not count as decreases; these counts describe directional consistency among contrasts that share observations.

\paragraph{Recovered decreasing contrasts.}
For sparse-input refits with spatial smoothing, coarser sampling reduced the number of recovered proficiency-related decreases. Decreasing contrasts at 864, 54 and 27 nodes were (10/10, 8/10, 8/10, 9/10), (5/10, 5/10, 1/10, 5/10) and (4/10, 4/10, 0/10, 4/10) for magnitude, coverage, cancellation per unit residual and cumulative torque cancellation, respectively. Supplementary Fig.~\ref{suppfig:residual-resolution-refit} shows all six resolutions; Supplementary Fig.~\ref{suppfig:residual-resolution-smoothing} shows the sizes of reductions relative to each skill's native Low reference.

\begin{figure}[!htbp]\centering
\includegraphics[width=\textwidth]{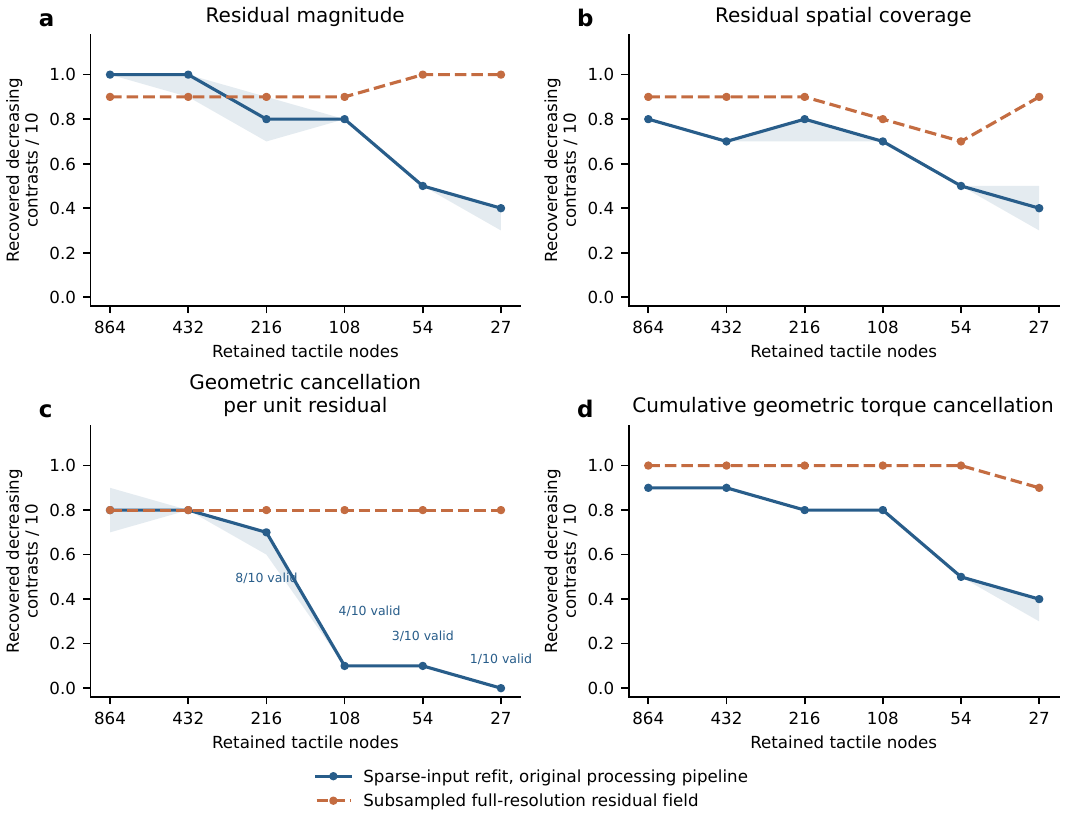}
\caption{\textbf{Recovered decreasing contrasts across six spatial resolutions.} Blue curves refit sparse tactile inputs under the original spatial and temporal smoothing pipeline; bands span three initialization seeds. Orange dashed curves subsample residual fields already inferred using all 864 nodes, without refitting. All curves use the ten planned contrasts as the denominator; annotations identify reductions in measurable contrasts. Straight segments connect evaluated resolutions.}
\label{suppfig:residual-resolution-refit}\end{figure}\FloatBarrier

\begin{table}[!htbp]\centering\small
\caption{\textbf{Recovered decreasing contrasts in residual metrics.} Entries count decreasing adjacent-level contrasts among five skills (ten planned contrasts). Sparse-input refits average three seeds before taking differences; the third condition subsamples a field inferred at full resolution. Main Fig.~4b uses that full-resolution baseline, whereas each refit condition uses its own re-estimated 864-node baseline. Parentheses give the number of measurable contrasts when below ten. At 54 nodes under the original pipeline, all five Expert fields have zero residual magnitude, spatial coverage and cumulative cancellation; cancellation per unit residual is undefined (N/A). For this ratio, three of the ten planned adjacent contrasts remain measurable and one decreases, reported as 1/10 (3).}
\label{supptable:residual-resolution-skills}
\begin{tabular}{L{.34\textwidth}rrrrrr}\toprule Metric & 864 & 432 & 216 & 108 & 54 & 27 \\ \midrule
\multicolumn{7}{l}{\textit{Sparse-input refit, with spatial smoothing}}\\
Residual magnitude & 10/10 & 10/10 & 8/10 & 8/10 & 5/10 & 4/10\\
Residual spatial coverage & 8/10 & 7/10 & 8/10 & 7/10 & 5/10 & 4/10\\
Geometric cancellation per unit residual & 8/10 & 8/10 & 7/10 (8) & 1/10 (4) & 1/10 (3) & 0/10 (1)\\
Cumulative geometric torque cancellation & 9/10 & 9/10 & 8/10 & 8/10 & 5/10 & 4/10\\
\addlinespace
\multicolumn{7}{l}{\textit{Sparse-input refit, without spatial smoothing}}\\
Residual magnitude & 10/10 & 10/10 & 9/10 & 9/10 & 8/10 & 10/10\\
Residual spatial coverage & 6/10 & 6/10 & 4/10 & 4/10 & 5/10 & 6/10\\
Geometric cancellation per unit residual & 6/10 & 6/10 & 6/10 & 8/10 & 5/10 & 4/10\\
Cumulative geometric torque cancellation & 8/10 & 8/10 & 7/10 & 10/10 & 7/10 & 7/10\\
\addlinespace
\multicolumn{7}{l}{\textit{Subsampled full-resolution residual field}}\\
Residual magnitude & 9/10 & 9/10 & 9/10 & 9/10 & 10/10 & 10/10\\
Residual spatial coverage & 9/10 & 9/10 & 9/10 & 8/10 & 7/10 & 9/10\\
Geometric cancellation per unit residual & 8/10 & 8/10 & 8/10 & 8/10 & 8/10 & 8/10\\
Cumulative geometric torque cancellation & 10/10 & 10/10 & 10/10 & 10/10 & 10/10 & 9/10\\
\addlinespace
\bottomrule\end{tabular}\end{table}\FloatBarrier

\paragraph{Residual detectability at sparse sampling.}
Under the original processing pipeline, reducing tactile density weakened the recovery of proficiency-related residual patterns. At 54 nodes, all five Expert residual fields fell below the positive-residual threshold, leaving cancellation per unit residual undefined. Sensitivity to spatial smoothing indicates that this loss of detectability reflects both reduced sensing density and the processing applied to sparse inputs.

\begin{figure}[!htbp]\centering
\includegraphics[width=\textwidth]{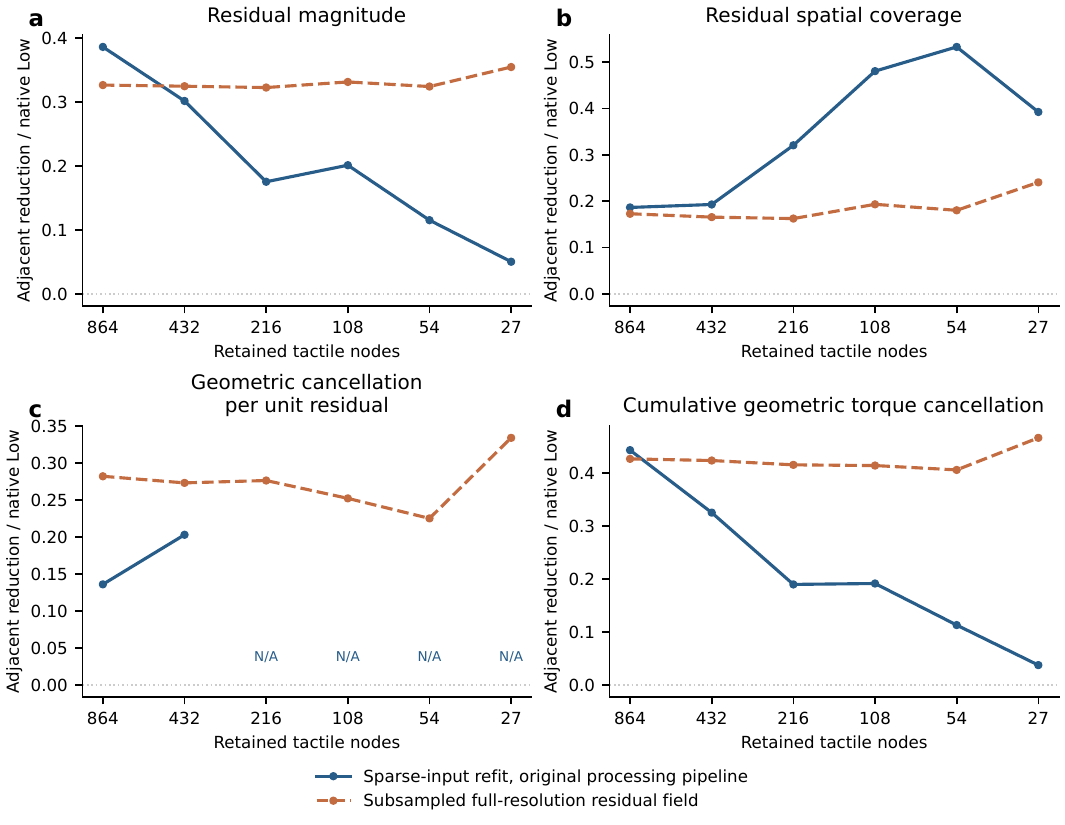}
\caption{\textbf{Signed residual reductions across spatial resolutions.} Each skill's two adjacent reductions are normalized by its Low estimate at 864 nodes, then averaged across contrasts and skills. Positive values favour the original decreasing direction. Lines are shown only where all five skill triplets are measurable; N/A denotes an unavailable five-skill summary. Measurable individual contrasts remain included in Supplementary Fig.~\ref{suppfig:residual-resolution-refit} and Supplementary Table~\ref{supptable:residual-resolution-skills}. Colours and line styles match Supplementary Fig.~\ref{suppfig:residual-resolution-refit}; native baselines are specific to each refitting/postprocessing condition.}
\label{suppfig:residual-resolution-smoothing}\end{figure}\FloatBarrier

\paragraph{Broad contrasts retained under downsampling.}
Some broad proficiency differences remained detectable with fewer spatial observations. Subsampling the full-resolution residual fields gave decreasing-contrast counts of (10/10, 7/10, 8/10, 10/10) at 54 nodes, compared with (9/10, 9/10, 8/10, 10/10) at 864 nodes. The effects of sparse sampling on local regional boundaries and temporal relationships are evaluated directly in Supplementary Notes~11 and 13, respectively; the residual comparisons provide complementary evidence. Processing details are given in Supplementary Methods~S13.

\section*{Supplementary Note 13: Temporal sampling of regional coordination}
\label{suppnote:temporal-fig4cd}

To test how sampling rate affects the regional coordination workflow, we recomputed conditional transfer entropy (TE) relationships, regional ordering and their three summaries at 100, 90, 80, 70, 60, 50, 40, 30, 25 and 20~Hz. The summaries were the two geometric ordering metrics in main Fig.~4c and inferred regional sequence duration, evaluated with the selected reference partitions. Duration was summarized as the mean span of ordering trees reconstructed across participant-bootstrap resamples (Supplementary Methods~S14). Every run tested all directed pairs on the same proficiency-specific ROI maps and retained the same strokes. Uniform polyphase resampling used anti-alias filtering, and the lag search included every positive integer delay within 50~ms. Rates below 20~Hz cannot represent such a delay and were excluded (Supplementary Methods~S14).

\paragraph{Coordination across sampling rates.}
The main coordination patterns remained visible at 50~Hz, with some change in their consistency across skills. Sequence-duration peaks occurred in 5/5 skills at 100~Hz and 4/5 at 50~Hz. Geometric ordering comparisons used 204 participant--skill pairs with both metrics available at every rate. Skills were averaged within participants before estimating High-minus-Low contrasts and bootstrap intervals, with Advanced and Expert pooled as High. All three comparisons used a recomputed 100-Hz baseline with matched participants and processing.

The endpoints differed in their sensitivity to lower sampling rates; all frequency-specific peak counts and effect sizes are retained in Supplementary Table~\ref{supptable:temporal-fig4cd-d}. Paired bootstrap intervals for changes from 100~Hz are supplied in the analysis records.

\begin{figure}[!htbp]\centering
\includegraphics[width=\textwidth]{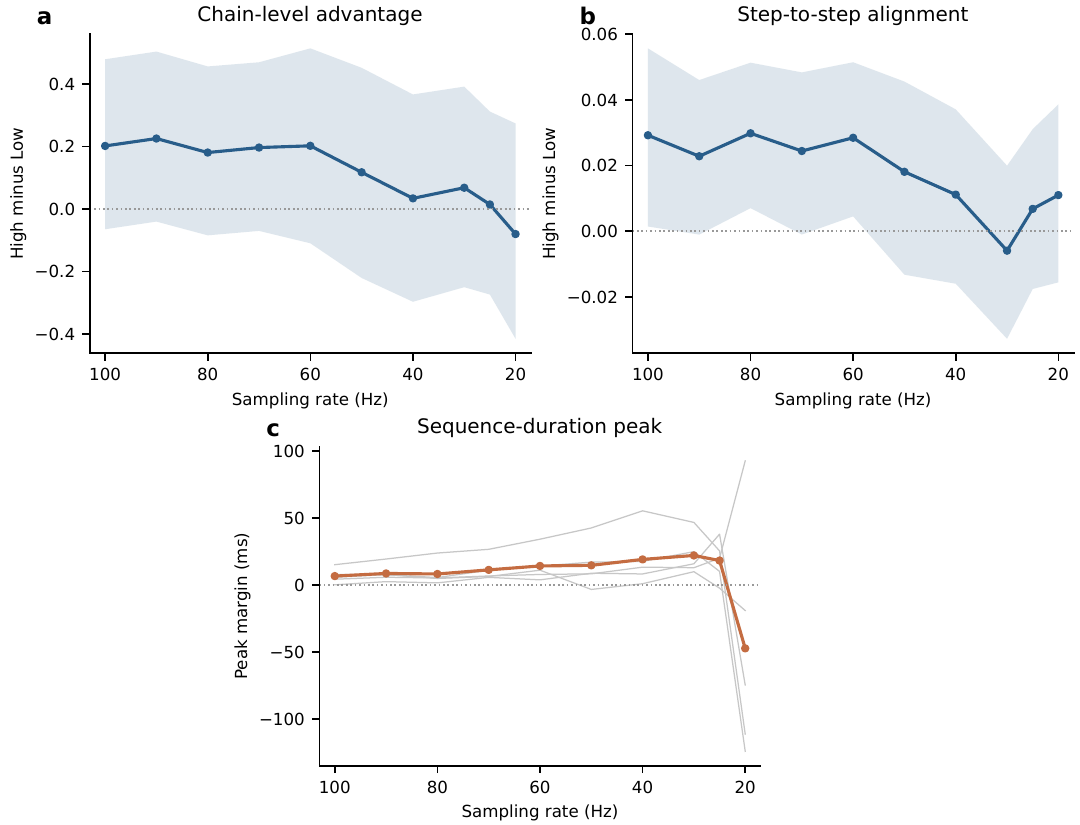}
\caption{\textbf{Sampling-rate effects on regional coordination.} Top, participant-average High-minus-Low geometric ordering contrasts with percentile 95\% participant-bootstrap intervals. Bottom, Advanced minus the larger Low or Expert value for sequence duration. Thin lines show skills and thick lines their mean. All tested rates are shown, including 90--60~Hz. Maps, candidate links and participants are fixed within comparisons; sampling changes the delay grid and effective sample count.}
\label{suppfig:temporal-fig4cd}\end{figure}\FloatBarrier

\begin{table}[!htbp]\centering\small
\caption{\textbf{Descriptive temporal peak contrasts across sampling rates.} Sequence duration uses the five-skill Advanced-minus-maximum-endpoint contrast. Counts give skills with an Advanced peak.}
\label{supptable:temporal-fig4cd-d}
\begin{tabular}{rrr}\toprule Rate (Hz) & Duration margin (ms) & Skills with Advanced peak\\\midrule
100 & 6.68 & 5/5\\
90 & 8.65 & 5/5\\
80 & 8.25 & 5/5\\
70 & 11.32 & 5/5\\
60 & 14.24 & 5/5\\
50 & 14.70 & 4/5\\
40 & 19.12 & 5/5\\
30 & 22.08 & 5/5\\
25 & 18.26 & 4/5\\
20 & -47.29 & 1/5\\
\bottomrule\end{tabular}\end{table}
\FloatBarrier

\paragraph{Temporal links retained after downsampling.}
Temporal downsampling progressively reduced agreement with the native link structure. Mean retention across skills at 50~Hz was 95.2\%, 93.2\% and 93.8\% for Low, Advanced and Expert, respectively; at 30~Hz it fell to 54.6\%, 55.2\% and 64.3\%, and at 20~Hz to 9.7\%, 11.1\% and 19.2\%. Retention is the fraction of a group's 100-Hz reference links detected again on the same map and participant set. Jaccard similarity measures agreement between the complete lower-rate and 100-Hz stable directed edge sets, accounting for both lost and newly detected links. Moderate temporal downsampling could preserve the overall acquire--refine pattern, even as the underlying regional timing relationships were progressively lost. At 20~Hz, the pattern became largely unrecoverable.

\begin{figure}[!htbp]\centering
\includegraphics[width=\textwidth]{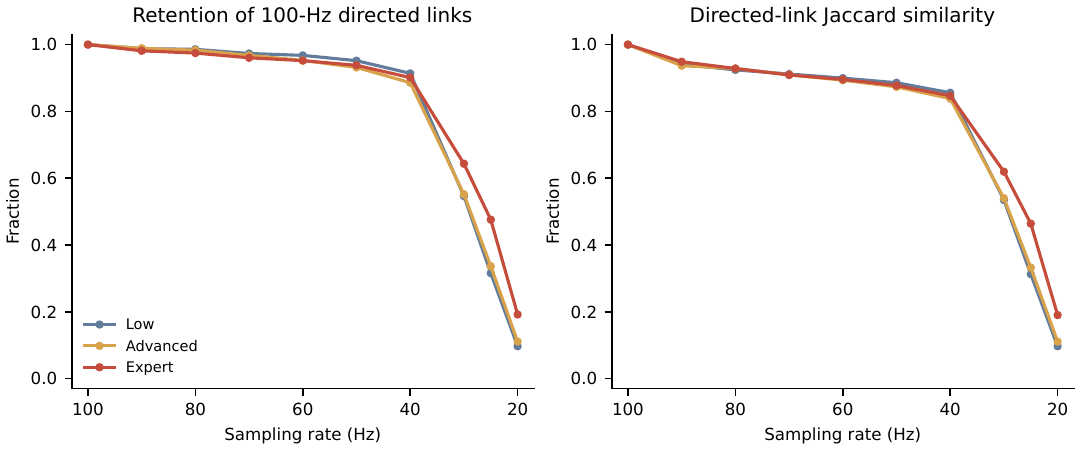}
\caption{\textbf{Retention and similarity of temporal relationships.} Left, fraction of each group's 100-Hz stable directed links retained at lower rates. Right, Jaccard similarity between the lower-rate and matched 100-Hz stable directed edge sets. Lines average five skill-specific values per group. Both measures describe agreement with the native graph.}
\label{suppfig:temporal-edge-retention}\end{figure}\FloatBarrier

\begin{table}[!htbp]\centering\small
\caption{\textbf{Retention of native stable directed links.} Entries are mean fractions across five skills. The same participant membership is used at every rate; eligibility requires at least 120 effective observations at 20~Hz.}
\label{supptable:temporal-retention}
\begin{tabular}{rrrr}\toprule Rate (Hz) & Low & Advanced & Expert\\\midrule
100 & 1.000 & 1.000 & 1.000\\
90 & 0.988 & 0.989 & 0.981\\
80 & 0.986 & 0.983 & 0.975\\
70 & 0.973 & 0.968 & 0.960\\
60 & 0.968 & 0.952 & 0.952\\
50 & 0.952 & 0.932 & 0.938\\
40 & 0.914 & 0.886 & 0.901\\
30 & 0.546 & 0.552 & 0.643\\
25 & 0.315 & 0.336 & 0.476\\
20 & 0.097 & 0.111 & 0.192\\
\bottomrule\end{tabular}\end{table}
\FloatBarrier

\paragraph{Scope of the comparison.}
The rate comparison held region maps and candidate links fixed, so it traced changes in temporal information through the same spatial representation. Previous-sample conditioning and a fixed physical lag ceiling made the available delays and statistical information vary with rate. Every tested rate is shown; fixed-count and pooled-map controls are reported in Supplementary Note~8.

\section*{Supplementary Note 14: Personalized force--motion trajectory optimization}

Motion guidance sought trajectories that better coordinated hand movement with a supplied contact pattern. A skill-conditioned reward model scored tactile torque-proxy features together with candidate inertial trajectories, keeping tactile input fixed during selection. The model dataset contained 7,736 strokes from 50 participants, allocated to training and validation using the participant--skill split in Supplementary Methods~S6.

The reward model favoured higher-proficiency and appropriately timed force--motion combinations. Pairwise preference accuracy was 0.828 for pooled higher- versus lower-proficiency pairs and 0.718 for ordered proficiency-level pairs. Temporal-shift ranking accuracy was 0.726 and 0.720, respectively, and mismatch ranking accuracy was 0.737. A conditional motion model fitted on the same training split generated eight candidates per supplied tactile pattern for the examples below. These tests assessed trajectory ranking; Supplementary Note~15 reports learners' performance.

\begin{figure}[!htbp]
\centering
\includegraphics[width=0.98\textwidth]{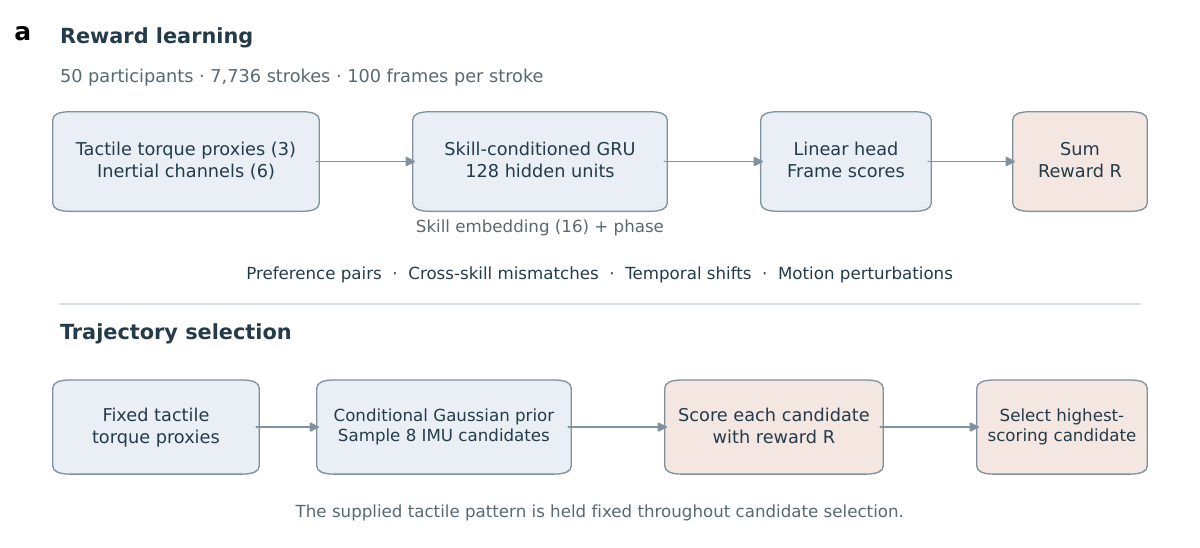}\par\vspace{8pt}
\figpanelthree{b}{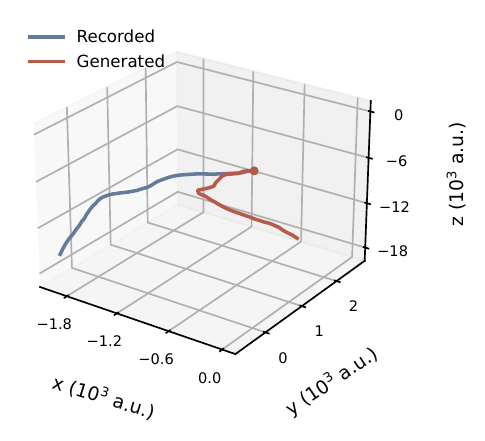}\hspace{0.04\textwidth}
\figpanelthree{c}{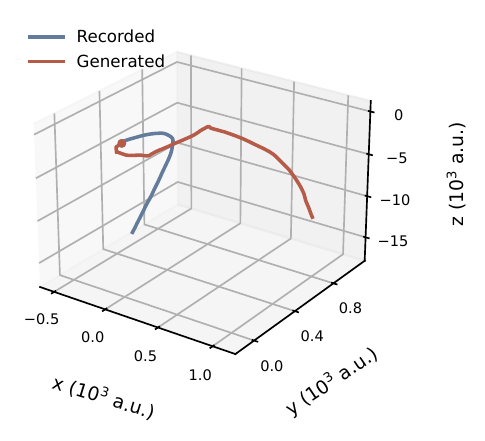}
\caption{\textbf{Scoring force--motion coordination and selecting motion guidance.} \textbf{a,} Reward model and candidate IMU trajectories. \textbf{b,c,} Recorded and generated trajectories for two predetermined validation participant--skill combinations: backhand rear-court transition and backhand inward spinning. Recorded trajectories average strokes within each combination. IMU-derived displacement is in arbitrary units. Tactile input is held fixed while the model ranks eight candidate inertial sequences. Validation is reported in Supplementary Note~14.}
\label{suppfig:trajectory-optimization}
\end{figure}
\FloatBarrier

\section*{Supplementary Note 15: Skill-specific learning outcomes and rating robustness}

\begin{figure}[!htbp]
\centering
\includegraphics[width=0.98\textwidth]{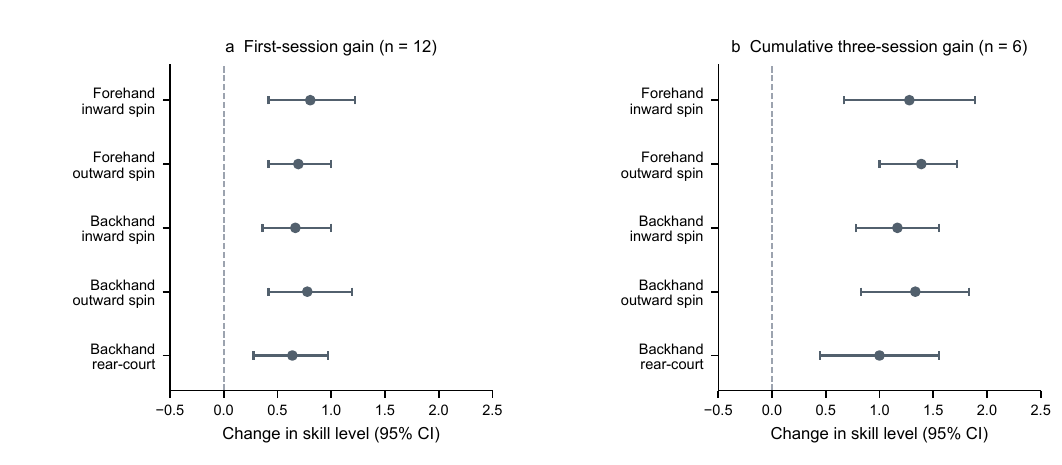}
\caption{\textbf{Mechanism-feedback learning gains across five racket-control skills.} \textbf{a,} First-session pre-to-post changes in 12 participants. \textbf{b,} Cumulative changes from session-1 pre to session-3 post in six participants. Circles show expert-rated mean gains; bars show 95\% percentile intervals from 20,000 participant bootstrap samples. Each contrast retains the paired assessment states and averages the three raters before estimating participant-level uncertainty.}
\label{suppfig:skill-gains}
\end{figure}

\begin{figure}[!htbp]
\centering
\includegraphics[width=0.91\textwidth,height=0.68\textheight,keepaspectratio]{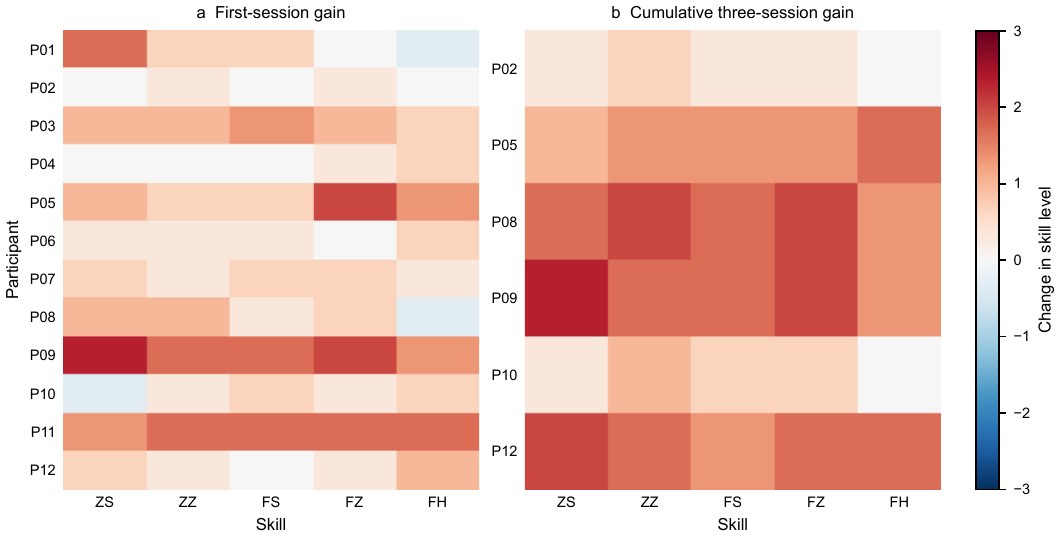}
\caption{\textbf{Variation in learning gains across participants and skills.} \textbf{a,} First-session expert-rated changes in 12 learners across five skills. \textbf{b,} Cumulative changes in the six-person three-session subset. Each cell is a paired endpoint difference averaged across three experts. Both panels use the same colour scale and participant identifiers. ZS, forehand inward spinning; ZZ, forehand outward spinning; FS, backhand inward spinning; FZ, backhand outward spinning; FH, backhand rear-court transition.}
\label{suppfig:gain-heterogeneity}
\end{figure}

Mechanism-guided learners improved across all five skills. Mean expert-rated gains ranged from 0.64 to 0.81 levels after the first session and from 1.00 to 1.39 levels across three sessions (Supplementary Fig.~\ref{suppfig:skill-gains}). Participant--skill maps show how gains varied among learners (Supplementary Fig.~\ref{suppfig:gain-heterogeneity}).

To assess skill-specific gains and rating consistency, the guidance cohort contributed 48 assessment states, each scored for five skills by three experts: 240 video--skill targets and 720 ratings. The first session included 12 participants, four at each of the beginner, intermediate and advanced levels. Six continued for three sessions, two from each level, with both session intervals fixed at 1~day. The separate 12-participant level-only comparison is reported in Supplementary Note~16.

\begin{figure}[!htbp]
\centering
\includegraphics[width=0.94\textwidth,height=0.66\textheight,keepaspectratio]{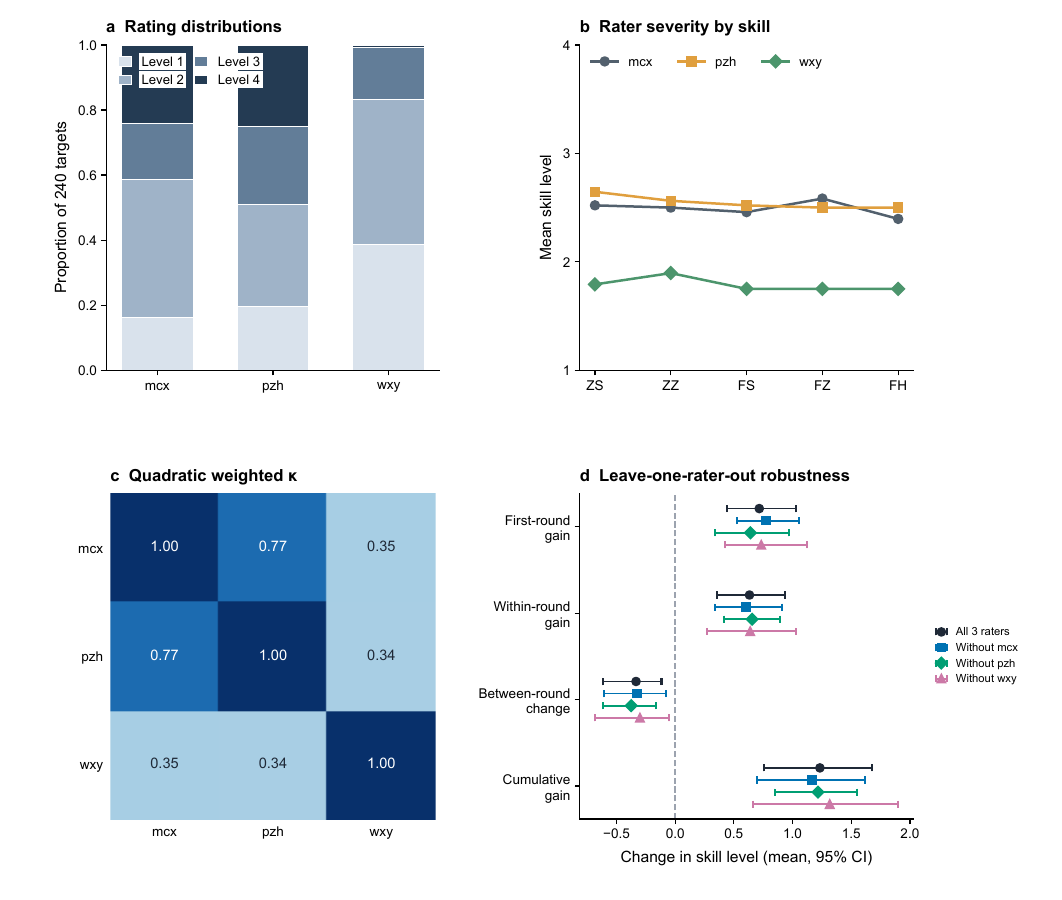}
\caption{\textbf{Expert-rating reliability and sensitivity to rater inclusion.} \textbf{a,} Rating distributions for three experts over 240 unique video--skill targets. \textbf{b,} Mean skill level by skill. \textbf{c,} Pairwise quadratic-weighted $\kappa$. \textbf{d,} Participant-average first-session gain, within-session gain, between-session change and cumulative gain using all three experts or leaving one out. Error bars are participant-bootstrap 95\% intervals from 20,000 resamples; the first-session contrast uses 12 participants and the three-session contrasts six. Rater abbreviations identify the same raters across panels.}
\label{suppfig:rater-robustness}
\end{figure}

The learning gains were retained when any one expert's ratings were removed. Across 240 targets, absolute-agreement reliability was ICC(A,1)$=0.489$ for a single expert and ICC(A,3)$=0.741$ for the three-expert mean. Pairwise quadratic-weighted $\kappa$ ranged from 0.340 to 0.771, indicating variation in rating severity. Omitting one expert at a time left mean first-session gains of 0.64--0.78 levels and cumulative gains of 1.17--1.32 levels; all corresponding participant-bootstrap 95\% intervals excluded zero (Supplementary Fig.~\ref{suppfig:rater-robustness}).

\FloatBarrier

\FloatBarrier

\FloatBarrier

\FloatBarrier

\section*{Supplementary Note 16: Mechanism-based guidance versus level-only feedback}
\label{suppnote:feedback-control}

To test whether mechanism guidance improved learning beyond proficiency-level feedback, we compared two groups with equal practice time. After selection, a computer program randomly assigned 24 learners, 12 per group, with four beginners, four intermediate players and four advanced players in each. One group received proficiency-level feedback plus force/contact and motion guidance; the other received proficiency-level feedback alone. Both practised for 10~min across five skills, freely allocating time among them. The system delivered feedback.

To assess improvement, the same three experts scored shuffled videos without labels for proficiency or feedback group. They were instructed to rate each clip independently without inferring assessment time. Ratings were averaged across experts within participant, assessment and skill, then equally across skills. The group comparison covered the first session. Six guided learners continued for three sessions, with both intervals fixed at 1~day.

Mechanism guidance produced larger skill-level gains than level-only feedback. Mean gains were $+0.72$ and $+0.05$ levels (Supplementary Table~\ref{supptable:feedback-control-overall}), a difference of $+0.67$ levels (participant-bootstrap 95\% CI $+0.39$ to $+0.98$; two-sided Welch $P=0.0014$). All 12 guided learners improved in their five-skill mean rating; among level-only learners, five improved, six were unchanged and one declined. Adjusting for baseline skill level gave a similar group coefficient of $+0.66$ levels (HC3-robust 95\% CI $+0.29$ to $+1.03$; Supplementary Methods~S15).

\begin{table}[!htbp]
\centering\small
\caption{\textbf{First-session skill-level comparison.} Skill levels are on the 1--4 scale and are averaged across three experts and five skills for each participant. Brackets report 95\% confidence intervals. Group gains and the difference in gains use participant percentile bootstraps; the baseline-adjusted coefficient uses an HC3-robust $t$ interval. Each group contains 12 different participants: four beginners, four intermediate players and four advanced players.}
\label{supptable:feedback-control-overall}
\begin{tabular}{L{0.44\textwidth}ccL{0.24\textwidth}}
\toprule
Feedback condition or contrast & Pre & Post & Change or difference [95\% CI] \\
\midrule
Mechanism guidance ($n=12$) & 1.85 & 2.57 & 0.72 [0.44, 1.02] \\
Level-only feedback ($n=12$) & 1.91 & 1.96 & 0.05 [$-0.01$, 0.11] \\
Difference in gains & --- & --- & 0.67 [0.39, 0.98] \\
Baseline-adjusted group coefficient & --- & --- & 0.66 [0.29, 1.03] \\
\bottomrule
\end{tabular}
\end{table}
\FloatBarrier

\begin{figure}[!htbp]
\centering
\includegraphics[width=\textwidth]{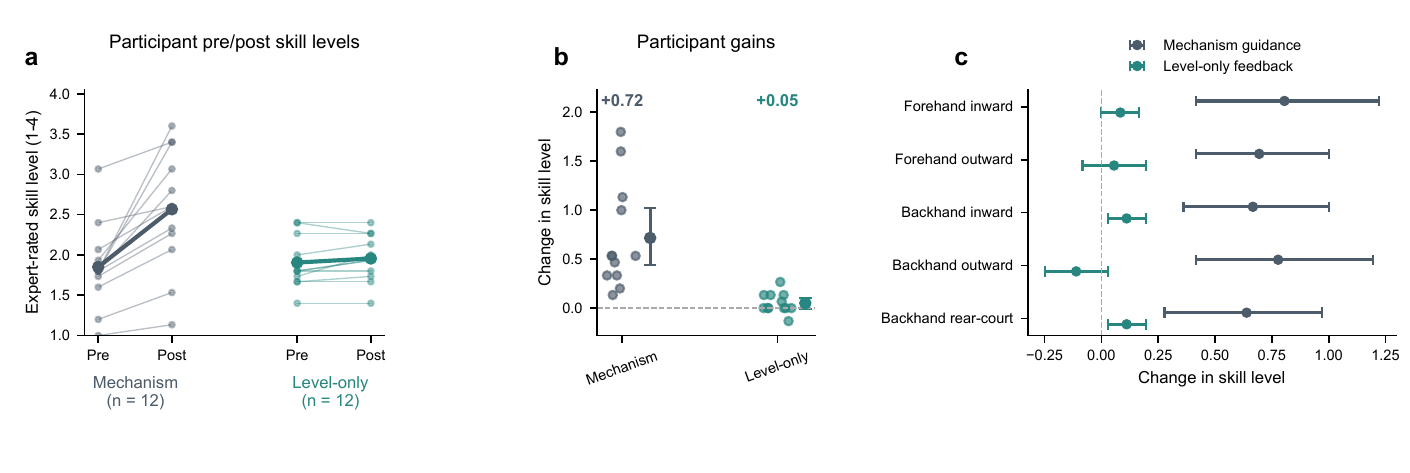}
\caption{\textbf{Expert-rated gains with mechanism guidance and level-only feedback.} \textbf{a,} Participant-average pre/post ratings, with thin lines joining paired assessments and thick lines showing group means. \textbf{b,} Participant changes and group means with bootstrap 95\% intervals. \textbf{c,} Mean changes per skill with bootstrap 95\% intervals. Each condition contains 12 different participants; resampling retains each person's skills and raters together. Grey denotes guidance and teal level-only feedback. Practice duration and experts were matched. Skill-specific intervals are descriptive and unadjusted for multiple comparisons.}
\label{suppfig:feedback-control}
\end{figure}
\FloatBarrier

\begin{table}[!htbp]
\centering\small
\caption{\textbf{Expert-rated changes by skill in the two feedback conditions.} Values are mean post-minus-pre changes [95\% participant-bootstrap intervals], after averaging the three experts. The last column is mechanism minus level-only mean change; bootstrap draws are independent across groups. There are 12 participants per group. Skill-specific comparisons are descriptive.}
\label{supptable:feedback-control-skills}
\begin{tabular}{L{0.27\textwidth}L{0.21\textwidth}L{0.21\textwidth}L{0.20\textwidth}}
\toprule
Skill & Mechanism guidance & Level-only feedback & Difference in gains \\
\midrule
Forehand inward spinning & 0.81 [0.42, 1.22] & 0.08 [0.00, 0.17] & 0.72 [0.31, 1.14] \\
Forehand outward spinning & 0.69 [0.42, 1.00] & 0.06 [$-0.08$, 0.19] & 0.64 [0.33, 0.97] \\
Backhand inward spinning & 0.67 [0.36, 1.00] & 0.11 [0.03, 0.19] & 0.56 [0.22, 0.92] \\
Backhand outward spinning & 0.78 [0.42, 1.19] & $-0.11$ [$-0.25$, 0.03] & 0.89 [0.50, 1.33] \\
Backhand rear-court transition & 0.64 [0.28, 0.97] & 0.11 [0.03, 0.19] & 0.53 [0.17, 0.89] \\
\bottomrule
\end{tabular}
\end{table}
\FloatBarrier

\paragraph{Control-related ratings.}
Skill-level gains accompanied improvements in both force release and force--motion coordination. Guided learners improved by 0.64 levels in force-release control (participant-bootstrap 95\% CI $+0.43$ to $+0.87$) and 0.62 levels in hand motion--force coordination (95\% CI $+0.41$ to $+0.85$). Mean overall gains were positive for every skill and remained positive when any one expert was omitted (Supplementary Note~15).

\paragraph{Exploratory tactile changes during practice.}
To examine whether practice changed the control organization targeted by guidance, we compared early and late trial windows within each participant, skill and session. The discovery encoder and channel normalization remained fixed. Drive-peak time and the cross-skill range of drive--grip recruitment angles therefore used the same representation as mechanism discovery, with endpoint selection specified in the main Methods.

Guided learners showed trends towards earlier drive recruitment and greater skill specificity. Their mean drive peak advanced from 33.57 to 18.78~ms after impact, a change of 14.79~ms; the level-only mean changed from 48.60 to 50.09~ms. Regression adjusting for early timing estimated an additional advance of 28.55~ms with guidance (HC3-robust 95\% CI $-23.51$ to $+80.60$~ms). Recruitment-angle range increased from $42.85^\circ$ to $45.98^\circ$ (change $+3.13^\circ$, participant-bootstrap 95\% CI $-6.03^\circ$ to $+12.28^\circ$), compared with $53.72^\circ$ to $51.07^\circ$ under level-only feedback. The directions matched higher-proficiency patterns, although intervals included zero.

Repeated guided practice produced cumulative performance gains with partial retention between sessions. Across S1-pre to S3-post in six learners, expert-rated skill level increased by $1.23$ levels (95\% CI $+0.76$ to $+1.68$). Mean within-session gains were $+0.63$ levels and between-session changes were $-0.33$ levels. Over the same period, recruitment-angle range increased from $43.50^\circ$ to $55.45^\circ$ (change $+11.95^\circ$, participant-bootstrap 95\% CI $-3.05^\circ$ to $+28.16^\circ$). This exploratory tactile trend accompanied the accumulated expert-rated improvement.

The combined contact and motion guidance yielded larger mean gains in all five skills (Supplementary Fig.~\ref{suppfig:feedback-control}; Supplementary Table~\ref{supptable:feedback-control-skills}). Separate interventions could identify each component's contribution, while longer follow-up could assess sustained retention and transfer.

\clearpage
\section*{Supplementary Methods S1: Tactile--inertial glove hardware}

To measure contact and hand motion together, the glove combines flexible arrays on the palm side with a six-axis inertial measurement unit on the dorsal board. The inertial stream contains three-axis acceleration and angular velocity. Eight tactile arrays provide 864 nodes: four 96-node arrays for the non-thumb fingers, one 96-node thumb array and three 128-node palm arrays. The acquisition board reads eight tactile channels in parallel, addresses nodes sequentially within each array, and samples inertial signals in the same acquisition cycle. Both streams are synchronized at 100~Hz.

To establish consistent force readings before stroke collection, we checked array continuity, baseline stability and responses to known normal loads. Sensor-specific calibration curves converted raw readings into relative normal-force values. Subsequent analyses used these readings as distributed normal contact measurements; tangential forces were not measured.

\paragraph{Main-figure sensor characterization.}
To relate tactile response to applied force, a motor-driven linear stage delivered repeated normal loads (FuYu FM60102 FD04-B, HM86D driver and FMC4030 controller). A reference sensor (ZNLBM-VI-5KG, CHINO SENSOR; rated capacity 5~kg) measured force while the tactile specimen was sampled synchronously. Loading and unloading provided paired force--response measurements (Supplementary Fig.~\ref{suppfig:sensor-performance}).

To assess linearity during repeated pressing and release, we collected automated loading--unloading measurements over 0--3~N. The 2,000 mechanical cycles yielded 1,968 complete synchronized records (98.4\% data yield). Within each branch of each cycle, valid 0--3~N samples were grouped into 0.1-N bins containing at least two samples. Ordinary least squares fitted the bin-median force and response values with an intercept and equal bin weights. Fits used measured points without imputing endpoints. All loading cycles and 1,784 unloading cycles contained both endpoint bins; 184 unloading cycles lacked the 3.0-N bin and reached 2.922--2.942~N. Median cycle-wise $R^2$ was 0.968 for loading and 0.962 for unloading. Normalized root-mean-square error (NRMSE), divided by each branch's observed response span, had corresponding medians of 5.42\% and 5.88\%.

To separate short-term repeatability from slow drift, we removed trends in force-adjusted residuals using robust locally weighted smoothing (LOWESS). A centred 101-cycle window displayed the median and 5th--95th percentiles. The force-conditioned coefficient of variation was 0.592\% after detrending (segmented moving-block-bootstrap 95\% confidence interval, 0.567--0.668\%), compared with an undetrended within-session coefficient of variation of 5.67\%. These values describe local repeatability and overall session variation, respectively. The 95th-percentile recovery error was 0.749\% of the loaded response.

To compare uniformity across locations and repeated presses at the same location, we tested ten sites with 30 planned presses each, retaining 29--30 after quality control. Loading slopes over 0.3--2.8~N were normalized to the ten-site mean; within-site response variation was measured from the loading-branch 3.0-N bin (2.95--3.05~N), requiring at least two valid samples per cycle. This yielded 27--30 valid responses per site. The across-site slope coefficient of variation was 7.77\%, and the median within-site response coefficient of variation was 0.95\%.

\section*{Supplementary Methods S2: Participants and racket-control tasks}

All experimental procedures were reviewed and approved by the Ethics Committee of the Institute of Automation, Chinese Academy of Sciences (protocol no.~IA21-2302-140202). All participants provided informed consent to participate in the study and to the public sharing of de-identified tactile data. The consent did not permit public release of the underlying visual-recording dataset. Consent for publication was obtained from all individuals whose identifiable faces appear in this article or the accompanying videos.

To compare control across proficiency, participants were classified by training background, playing experience and competition record. The 50-person discovery cohort contained 8 beginners, 23 intermediate players, 11 advanced players and 8 expert or national-team-level players. Each performed repeated trials of up to five skills: forehand inward spinning, forehand outward spinning, backhand inward spinning, backhand outward spinning and backhand rear-court transition. Beginners could skip skills they could not execute reliably. Cross-sectional analyses pooled Beginner and Intermediate as Low; pooled High comparisons combined Advanced and Expert.

To obtain comparable single-contact strokes, data collection used repeated multi-shuttle feeding. A valid trial required one racket-string contact with the shuttle. Racket-frame contact, shaft contact and repeated racket--shuttle contacts were excluded.

The skill-learning cohort comprised a separate 24 participants (Supplementary Methods~S15). Outside these two primary cohorts, eight additional participants provided an auxiliary tactile--pose dataset. They held the racket in varied grip postures to supply paired tactile recordings and pose labels, expanding posture coverage for supervised pose estimation (Supplementary Methods~S7).

\section*{Supplementary Methods S3: Data preprocessing}

To compare the same phase across strokes, each valid trial was aligned to its manually annotated video impact frame. We extracted a 100-frame segment containing timestamps, six inertial channels and 864 tactile readings. Abnormal-duration segments were removed, including those longer than 1.1 s, which corresponded to an effective rate below 90 Hz.

To remove baseline offsets, tactile signals were corrected using calibration frames collected before and during each skill block. The discovery dataset contained 7,736 synchronized tactile--inertial stroke sequences from 50 participants and supplied the dynamic data for mechanism analysis and model development. The separate learning cohort contributed 6,989 strokes, giving 14,725 strokes overall.

\section*{Supplementary Methods S4: Metric definitions and statistics}

\paragraph{Definitions of main reported metrics.}
Supplementary Table~\ref{supptable:main-reported-metrics} links the main Fig.~3 labels to their computational endpoints. Frames are numbered from 1; impact is frame 50 and the sample interval is 10~ms. The model supplies $d_t=\|Z_{\mathrm{drive}}(t)\|_2$, $g_t=\|Z_{\mathrm{grip}}(t)\|_2$, $q_t=\|\boldsymbol\tau_{\mathrm{drive}}(t)\|_2$ and $a_t=\tfrac13\sum_{j=1}^3 k_{\mathrm{grip},j}(t)$. These are functional model coordinates; $q_t$ is a torque proxy in model units and $a_t$ is dimensionless.

\begin{table}[htbp]
\centering
\caption{\textbf{Definitions of main reported metrics.} Computational details below specify preprocessing and boundary rules. Group labels describe discovery-cohort proficiency; observations are aggregated only at the stated unit.}
\label{supptable:main-reported-metrics}
\small
\setlength{\tabcolsep}{4pt}
\renewcommand{\arraystretch}{1.15}
\begin{tabular}{L{0.22\textwidth}L{0.45\textwidth}L{0.27\textwidth}}
\toprule
Metric & Signal-to-scalar definition & Unit and displayed summary \\
\midrule
Angular spread (Fig.~3a) & For participant $i$ and skill $s$, $\theta_{is}=(\theta^{\rm pre}_{is}+\theta^{\rm post}_{is})/2$. Spread is $\max_s\theta_{is}-\min_s\theta_{is}$. Each phase angle is the direction of the normalized $(\Delta g,\Delta d)$ trajectory. & Degrees per participant with at least two skills; group mean. \\
Spatial force-basis distance (Fig.~3b) & Centre each spatial basis vector over sensors and normalize its length. Match $m=\min(K_s,K_{s'})$ vectors one-to-one to maximize cosine similarity $c$; distance is $1-m^{-1}\sum_{(u,v)\in\mathcal M}c_{uv}$. & Dimensionless distance per skill pair within group; median of ten pairs. \\
Temporal force-transfer variance (Fig.~3c) & Let $z_{is}$ be the person-centred, regional-standardized recruitment-time vector. Compute $100\sum_s n_s\|\bar z_s-\bar z\|_2^2/\sum_{i,s}\|z_{is}-\bar z\|_2^2$. & Percentage per proficiency group; full-data $R^2$. Dots are participant-delete-one estimates. \\
Drive-peak time (Fig.~3d, left) & For each stroke, compute $10(\arg\max_{10\leq t\leq90}d_t-50)$; take the median across strokes for each participant and skill. & Milliseconds relative to impact; pooled median of participant--skill values. \\
Stabilization allocation (Fig.~3d, middle) & Compute the largest prominence of $a_t$ separately in frames 1--50 and 51--100. Take stroke medians $A_{\rm pre},A_{\rm post}$, then $A_{\rm pre}/(A_{\rm pre}+A_{\rm post}+10^{-6})$. & Dimensionless fraction per participant--skill; pooled median. \\
Post-impact peak prominence (Fig.~3d, right) & Largest prominence of $q_t$ in frames 51--100 for each stroke, then median across strokes within participant and skill. & Model units per participant--skill; pooled median. \\
Expected lag / temporal entropy (Fig.~3f) & $L=\sum_{k=3}^8kw_k$; $H_{\mathrm{temp}}=-\sum_{k=3}^8w_k\log(w_k)/\log6$, with $\sum_kw_k=1$. & Frames / dimensionless, one pair per participant; mean $\pm$ SEM. Lag in ms is $10L$. \\
\bottomrule
\end{tabular}
\end{table}

\paragraph{Recruitment-angle construction.}
For Fig.~3a, retain valid latent samples over frames 10--90. Split each stroke at impact: frames 10--50 use the mean of frames 10--14 as baseline, and frames 51--90 use the mean of frames 86--90. Subtract the respective baseline from $g_t$ and $d_t$, retaining frames 14--50 and 51--86 for the directional fits. Divide the grip and drive differences by fixed scales $s_g=9.0861022949$ and $s_d=8.2009086609$, respectively: the 95th percentiles of non-zero absolute phase differences pooled across the discovery data. For each participant, skill and phase, pool the resulting $u_t=(\Delta g_t/s_g,\Delta d_t/s_d)$ from all strokes, omit non-finite or effectively zero vectors ($\|u_t\|_2\leq10^{-8}$), and take the leading eigenvector $v$ of $\sum_tu_tu_t^{\mathsf T}/n$. No further centring is applied. The phase angle is $(180/\pi)\operatorname{atan2}(|v_d|,|v_g|)$, between $0^\circ$ and $90^\circ$; both phases must be available. Averaging the two phase angles before taking the cross-skill range gives one value per person.

\paragraph{Spatial-basis matching and regional timing variance.}
For Fig.~3b, the skill--proficiency NMF bases from Supplementary Methods~S5 are compared on the same 864 sensors. In the order forehand inward spinning, forehand outward spinning, backhand inward spinning, backhand outward spinning and backhand rear-court transition, the basis counts were $(6,5,4,5,4)$ for Low, $(3,12,3,3,10)$ for Advanced and $(7,8,11,9,11)$ for Expert. Subtract each vector's sensor mean, divide by its Euclidean norm, and use the Hungarian assignment to maximize the sum of pairwise dot products. When ranks differ, only the smaller number of matched vectors enters the mean; unmatched vectors carry no additional penalty. All ten pairs among five skills enter the group median with equal weight.

For Fig.~3c, each participant--skill profile contains regional cumulative delays from the TE ordering tree in Supplementary Methods~S10. Subtract each person's regional mean across their available skills, requiring at least two skills. Within each proficiency group, replace missing person-centred entries by that regional coordinate's observed mean, then subtract the coordinate mean and divide by its population standard deviation. Coordinates with zero variance contribute zero. In the table's formula, $n_s$ is the number of participant--skill profiles for skill $s$, $\bar z_s$ its centroid and $\bar z$ the centroid of all profiles in the group. This is the fraction of standardized timing variation associated with skill identity, not the variance of TE edge strengths.

\paragraph{Peak prominence and aggregation.}
For the gate and torque-proxy metrics, use all valid decoder samples within frames 1--50 (pre-impact, including impact) or 51--100 (post-impact); unavailable history-dependent samples are omitted. Within each contiguous valid segment, identify local maxima, also allowing a segment endpoint if it exceeds its adjacent sample. From each maximum, search left and right until encountering a higher sample or the segment boundary. Subtract the higher of the two side minima from the peak; at an endpoint use the available side only. Keep the largest non-negative prominence in the phase; a flat phase has prominence zero. First compute stroke-wise prominences, then their medians within each participant--skill cell. The stabilization ratio is formed from these two medians, rather than by averaging stroke-wise ratios. Drive-peak time is likewise reduced to one within-cell median; a tied maximum uses its first frame. Main Fig.~3d displays medians of these 217 cells from 50 participants, with participant-clustered tests as specified below.

To match inference to the data structure, each analysis used its stated sampling unit. Impact-aligned measures included peak timing, peak prominence, full width at half maximum and pre/post-impact summaries. Sensor-wise residual tests used Benjamini--Hochberg correction. Main Fig.~3f used one temporal-weight estimate per participant, with Spearman tests for ordered proficiency trends and unadjusted two-sided Mann--Whitney pairwise tests.

To distinguish variation across tasks from variation across people, main Fig.~4b summarized five skill-specific residual estimates per group with 95\% $t$ intervals across skills. Its scaled participant--skill scatter shows variation around the corresponding aggregate estimate. Main Fig.~4c used participant summaries, bootstrap intervals and two-sided Welch tests comparing pooled Advanced/Expert with Low, with Benjamini--Hochberg correction across displayed metrics.

For main Fig.~4d, acquire required Advanced $>$ Low and refine required Advanced $>$ Expert; both inequalities were strict. We counted initialization-wise recovery of this pattern for regional counts and five-skill-averaged sequence duration. Supplementary Note~8 also reports all 50 initialization--skill duration comparisons. These frequencies describe algorithmic repeatability on the same participants; no significance test treated initializations as independent participants.

To preserve each person's repeated observations, learning intervals used 20,000 bootstrap samples of participants, keeping skills and assessment states together. Exact paired sign-flip tests enumerated participant-level sign assignments. Temporal-rate comparisons used matched eligible participant--skill records and paired bootstrap draws (Supplementary Methods~S14). Spatial subsampling averaged the parameter grid within each repeat and reported standard deviations across five repeats. Parameter settings and repeated skills were not counted as independent people. Seeds, configurations and numerical checks accompany the code.

\paragraph{Participant-level inference for main Fig.~3c,d.}
To estimate uncertainty in skill-related timing $R^2$ while preserving each person's skill set, Fig.~3c used leave-one-participant recomputation ($n=30$, 11 and 8 for Low, Advanced and Expert). For group $g$, jackknife variance was $V_g=(n_g-1)n_g^{-1}\sum_i(\widehat R^2_{g,-i}-\overline R^2_{g,-\cdot})^2$. The ordered contrast weighted full-data group estimates by $(-0.5,0,0.5)$, combining group variances for a two-sided $t$ test with Welch--Satterthwaite degrees of freedom. Displayed leave-one-participant estimates visualize this sensitivity.

To test proficiency trends while accounting for task and repeated observations, Fig.~3d regressed each metric on ordinal proficiency with skill fixed effects. The 217 participant--skill observations came from 50 participants. CR1 participant-clustered standard errors and a $t$ reference with 49 degrees of freedom gave two-sided $P$ values. The three outcomes formed one Holm-corrected comparison family. For drive-peak time, stabilization allocation and post-impact torque-proxy prominence, respectively, raw two-sided values were 0.0147393, 0.0038838 and 0.0245753; Holm-adjusted values were 0.0294786, 0.0116514 and 0.0294786. Main Fig.~3d displays these adjusted values, rounded to 0.029, 0.012 and 0.029. Plotted medians and percentile 95\% bands used 19,999 within-group participant-bootstrap samples, retaining all skills per sampled person.

\section*{Supplementary Methods S5: Spatial residual analysis}

To identify recurring force distributions and activity beyond a Low reference, we decomposed each skill-by-group dataset using non-negative matrix factorization:
\[
X \approx WH,
\]
where \(X\in\mathbb{R}_{+}^{T\times N}\) contains $T$ frames and $N$ sensor readings, \(W\) contains temporal activations and \(H\) contains spatial force patterns. To select the number of patterns, we compared bases across repeated split halves and fitted an adaptive piecewise curve to mean similarity over ranks 1--15. The 864-node Low reference used ranks 6, 5, 4, 5 and 4 for forehand inward spinning, forehand outward spinning, backhand inward spinning, backhand outward spinning and backhand rear-court transition. Rank scans retained all 50 repeats, each holding out two participants and using 320 fitting epochs, 300 activation-refit steps during the scan and 90 holdout-refit steps. Final bases used seed 9000002. Each skill's rank and basis were fixed for comparison groups.

To compare force patterns against the same reference, target-group data were fitted to the fixed Low spatial basis \(H\):
\[
W^{*}=\arg\min_{W\geq 0}\|X-WH\|_{F}^{2},
\]
and the residual was computed as
\[
R=X-W^{*}H.
\]
The positive component \(R_+=\max(R,0)\) identifies activity beyond the reference reconstruction. Held-out Low data supplied the baseline comparison, with Benjamini--Hochberg correction for sensor-wise tests. For main Fig.~4a,b, reconstruction energy was matched per segment before expressing residuals as $100(X-\widehat X)/\max(|\widehat X|,\epsilon)$. Signed curves were clipped to $[-300,300]$, averaged and smoothed within skill-by-group combinations, then thresholded at 10\%. Values are percentage residual responses. Spatial coverage counted each sensor once if it reached threshold in any of the 100 frames of the skill-by-group mean field.

To assess whether local contributions reinforce or oppose each other, we calculated torque at node $i$ and frame $t$ as $\boldsymbol\tau_i(t)=(\mathbf r_i(t)-\mathbf p(t))\times[S_{ti}\mathbf n_i(t)]$. Here $S_{ti}$ is thresholded residual response, $\mathbf r_i(t)$ reconstructed position, $\mathbf p(t)$ the estimated pivot and $\mathbf n_i(t)$ the unit contact normal. Cumulative geometric cancellation was
\[
D_\tau=\sum_t\left(\sum_i\|\boldsymbol\tau_i(t)\|-\left\|\sum_i\boldsymbol\tau_i(t)\right\|\right).
\]
The difference between summed local magnitudes and net torque measures opposition among contributions, accumulated over the 100-frame window in residual-response--geometry units. Cancellation per unit residual is $q_\tau=D_\tau/F_R$, with $F_R=\sum_{t,i}S_{ti}$ the cumulative retained response. It is unavailable when $F_R=0$. Uniform amplitude scaling cancels from the ratio when geometry and active locations are fixed. These measures provide a torque-based interpretation of contact organization; energy expenditure was not directly measured (Supplementary Methods~S13).

To summarize across tasks, main Fig.~4b averages five skill-specific estimates per group with 95\% $t$ intervals across skills. In the magnitude panel, participant--skill values were scaled within each skill-by-group combination to centre on its aggregate estimate. Cancellation panels show the five skill estimates, their mean and interval. The eight decreasing adjacent contrasts for the normalized metric describe consistency across skills; they are not a participant-level significance test.

\section*{Supplementary Methods S6: Feedback reward and trajectory optimization}

\paragraph{From population patterns to individual contact guidance.}
Contact guidance followed five steps: learner measurement, task-matched reference comparison, discrepancy calculation, priority selection and instruction display. The population findings identified relevant dimensions of control---spatial force distribution and regional timing---and each learner's measured differences determined the specific locations and timing adjustments presented.

\emph{Measurement and reference.} For the selected skill, the system processed the learner's tactile--inertial stroke into contact maps, regional activity traces and drive--grip profiles using the same signal definitions as the discovery analysis. Contact locations were expressed on the common hand--racket layout, and timing was aligned to racket--shuttle impact. The higher-proficiency reference comprised the spatiotemporal force-control patterns obtained from Expert players performing the same skill, so differences reflected how that task was executed. The drive--grip profiles provided functional context for interpreting local contact differences. For spatial feedback, the same Expert-derived basis for the selected skill was used to compute both learner and Expert residual fields, and the Expert residual field served as the feedback target. This guidance basis differs from the Low-derived basis used for the discovery analysis in Supplementary Methods~S5.

\emph{Discrepancy and selection.} The system compared spatial contact activity at corresponding locations and the timing of activity in corresponding hand regions. Spatial guidance used percentage residuals $r=100(X-\widehat X)/\max(|\widehat X|,\epsilon)$, where $X$ is the measured tactile signal, $\widehat X$ its reconstruction using that Expert-derived basis and $\epsilon=1$ prevents division by a near-zero reconstructed value in the processed signal units; residuals were clipped to $[-300,300]$. The learner--Expert difference was computed between their mean residual fields and averaged over each retained sensor--time component. Its magnitude supplied the displayed percentage, and its sign determined whether contact should be reduced (excess residual) or increased (deficient residual). Spatial differences retained the direction of excess or insufficient contact relative to the reference; timing differences retained whether the learner's activity occurred earlier or later. The two sets of differences were ranked separately by absolute magnitude. The three largest spatial discrepancies selected the contact locations for force-distribution guidance, and the three largest timing discrepancies selected the regions for temporal guidance. Spatial and temporal quantities were not combined into a single score.

\emph{Instruction display.} The selected spatial differences were translated into named hand locations with instructions to increase or reduce contact. Timing differences were translated into instructions to advance or delay the selected regional activity, expressed in milliseconds relative to the reference. A learner who activated a selected region later than the reference received an earlier-activation cue; the opposite difference produced a later-activation cue. These instructions appeared with the learner's tactile maps, regional timing visualizations and synchronized stroke playback, alongside skill-specific verbal cues. Thus the selected targets depended on the individual's discrepancies, while the task-matched reference supplied the desired direction of adjustment. The complementary motion-candidate search described below addressed hand movement with the measured tactile pattern held fixed.

To identify movements compatible with a supplied contact pattern, a skill-conditioned reward model \(\mathcal{R}(X,U)\) scored tactile sequence $X$ together with inertial trajectory $U$. The 864 tactile channels were converted into three signed torque proxies using fixed hand--racket geometry. We subtracted each sensor's 10th-percentile baseline, divided positive responses by their 99th percentile, clipped them to [0,1], and combined them with geometric moment vectors. A 128-unit gated recurrent unit (GRU) received these three proxies, six inertial channels, impact-relative phase and a 16-dimensional skill representation. Frame scores were summed over the 100-frame stroke. A learned conditional motion model generated candidate inertial sequences \(\mathcal{C}\), which were scored with the tactile pattern \(X_{\mathrm{fixed}}\) held constant:
\[
U^{*}=\arg\max_{U_s\in\mathcal{C}}\left[\mathcal{R}(X_{\mathrm{fixed}},U_s)-\frac{\lambda_d}{6T}\|U_s-U_{\mathrm{orig}}\|_F^2\right].
\]
Here $U_{\mathrm{orig}}$ is the recorded trajectory, $T$ the number of frames, and $\lambda_d$ the weight on mean-squared deviation in normalized IMU coordinates. The highest-scoring candidate became a hand-motion suggestion, evaluated alongside contact guidance in the learning experiment.

To test whether the score reflected skill and force--motion compatibility, validation used three ranking measures. Preference accuracy counted pairs where the higher-proficiency sequence scored higher: pooled comparisons combined Advanced/Expert and Beginner/Intermediate, while all-level comparisons retained ordered proficiency pairs. Mismatch accuracy compared recorded motion with motion substituted from another skill or altered in amplitude/noise. Temporal-shift accuracy compared reference alignment with shifted inertial sequences. Expert ratings separately assessed learning outcomes.

To keep strokes from the same participant--skill combination together, training and validation used fixed cell assignments. Validation contained 19 cells from 15 people; a participant could contribute another skill to training. Feature means and standard deviations were fitted on training data only. An initial higher-proficiency reward model supplied temporal alignment centres for lower-proficiency training strokes.

To learn the ranking, the full reward model used Adam for 500 epochs at learning rate $3\times10^{-4}$, batches of 64 and up to 6,000 preference pairs per skill. Preference loss had unit weight; cross-skill mismatch, temporal-shift, perturbed-motion and absolute-score terms had weights 1.1, 3.2, 0.5 and 0.3. Shifted examples used offsets of 6, 8, 10, 12, 14 or 16 frames (60--160~ms), with one offset sampled per training anchor. Epoch-wise validation used one shift draw per anchor and final evaluation used three. Each anchor also supplied six cross-skill and six perturbed-motion examples. Checkpoint selection maximized $0.5a_{\rm preference}+0.3a_{\rm mismatch}+0.2a_{\rm shift}$ on validation data, where each $a$ is the corresponding ranking accuracy.

To generate candidate movements, the conditional Gaussian motion model used a 128-unit GRU and 16-dimensional skill representation with tactile proxies and phase as input. It trained for 30 epochs on the same split, with checkpoint selection by validation negative log likelihood. Illustrated searches used eight draws and $\lambda_d=0$, holding tactile input at the participant--skill mean. Examples were predetermined; within each, the candidate with the highest coordination score was selected.

\section*{Supplementary Methods S7: Mechanics-informed tactile representation}

To express tactile activity in mechanical coordinates, we estimated each sensor's position and contact-normal direction relative to the racket. A graph-based temporal model learned hand--racket pose prediction from the auxiliary static-posture dataset provided by eight additional participants. The dataset covered more than 100 grip postures, supplying diverse paired tactile recordings and pose labels for supervised training. Predicted poses initialized the geometry. Physical refinement then adjusted the hand model to bring active nodes towards the handle and align their contact normals.

To connect functional control with local forces, the organization model and spatial residual analysis used this common geometry for contact descriptors and racket-driving torque proxies. Regional-order analysis used the two-dimensional unit-torque construction defined in Supplementary Methods~S10.

\paragraph{Temporal physical reconstruction.}
To stabilize contact geometry over time, reconstruction fitted the current frame together with up to four preceding frames. Hand layout, phalanx lengths and handle dimensions were fixed. Joint and hand--racket pose variables minimized active-node contact distances, contact-normal mismatch, penetration and deviations from pose and joint priors. Regularization also constrained changes from previous joint and projected racket-pose estimates. Robust velocity and acceleration terms for all 864 reconstructed nodes had weights 0.002 and 0.05 after normalization by handle radius and node count. Previously output positions stayed fixed while internal window states could be refitted. Temporal differences used actual timestamps relative to a 10-ms interval, and the solver allowed 70 function evaluations per window. Configuration selection considered contact distance, normal alignment, penetration, two-dimensional continuity, movement amplitude and three-dimensional acceleration; settings and selection rules accompany the code.

To test reconstruction with and without local contact evidence, each of the 50 discovery participants contributed a fixed seven-frame sequence. Data from ten participants were used for configuration selection; parameters were then frozen before evaluation in the remaining 40. Reconstruction development and this assessment used the same existing dynamic cohort. The regional-omission condition removed one fixed sensor-group fold per participant from every physical contact-data term throughout the sequence; complete-input fits were run separately. Both used saved pose predictions for initialization. Each node's baseline was the median response among the lowest-load 10\% of frames in its source recording, with at least ten frames. A node was active when its positive corrected response exceeded both three robust noise standard deviations and one recorded count. Noise was estimated as 1.4826 times the median absolute deviation in the low-load frames. The physical solver used the stated history window; pose initialization and baseline estimation could use the wider recording.

To summarize contact agreement without over-weighting people or frames with larger signals, we first normalized tactile-response weights within each frame, then weighted frames and participants equally. For participant $p$, valid frame $t$ and active node $i$, $s_{pti}$ is positive baseline-corrected response and $d_{pti}$ is nearest-boundary distance to the finite octagonal handle, divided by inradius $R$. We assigned
\[
w_{pti}=\frac{1}{N T_p}\frac{s_{pti}}{\sum_{j\in A_{pt}}s_{ptj}},
\]
where $N$ is participant count, $T_p$ is participant $p$'s number of valid contact frames, and $A_{pt}$ contains active nodes in the evaluated region: all nodes for complete input, or omitted nodes for regional omission. Mean distance is $\sum_{pti}w_{pti}d_{pti}$; the weighted median is the distance where cumulative weight reaches 0.5, with the 90th percentile defined similarly. These quantiles describe the weighted node-distance distribution. We retained all three summaries and used 5,000 participant-bootstrap resamples for confidence intervals on the mean. Frames without detected contact in the evaluated region were missing values. Penetration was assessed over all nodes as the proportion exceeding depth $0.05R$. Mean node acceleration and endpoint displacement over the final three frames measured smoothness and movement amplitude.

\paragraph{Displacement and solver diagnostics.}
In the reconstruction benchmark, complete-input mean endpoint displacement increased from $0.150R$ to $0.199R$, with one sequence reaching $2.110R$. Mean normalized two-dimensional racket-pose acceleration increased from 0.0350 to 0.0411; this pose measure is distinct from reconstructed three-dimensional node acceleration. In each condition, 231 of 240 windows met the solver's termination criterion; windows reaching the iteration limit were retained.

\section*{Supplementary Methods S8: Control-organization model}

To separate motion generation from grip stabilization while preserving tactile structure, a graph-informed encoder mapped each frame to 192 features. It was pretrained for 30 epochs to reconstruct tactile input, then frozen. A multilayer projection applied separately to each frame produced non-negative 8-dimensional representations \(Z_{\mathrm{drive}}\) and \(Z_{\mathrm{grip}}\). The decoder combined their history over \(k=3,\ldots,8\), corresponding to 30--80~ms at 100~Hz. Each participant shared one temporal-weight profile across branches and skills:
\[
w_{p,k}=\frac{\exp(a_k+b_{p,k})}{\sum_{j=3}^{8}\exp(a_j+b_{p,j})},
\qquad \widetilde Z_p(t)=\sum_{k=3}^{8}w_{p,k}Z_p(t-k).
\]
Here \(a_k\) gives shared log-weights and \(b_{p,k}\) gives participant-specific offsets; the exponential normalization makes weights sum to one. Proficiency labels were not inputs. Main control training held weights uniform for 30 epochs, followed by eight epochs updating only temporal weights with the encoder, projections and decoder fixed. Expected lag, \(\bar{k}_p=\sum_k w_{p,k}k\), measures the weighted delay in frames; normalized entropy, \(-\sum_k w_{p,k}\log w_{p,k}/\log 6\), measures its spread. Multiplying lag by the frame interval gives the model-estimated perception-to-control interval. Proficiency comparisons used one fitted profile per participant.

To give drive and grip distinct roles in motion prediction, the decoder combined their weighted histories with current estimated pose, velocity and IMU state. Drive predicted a three-axis torque proxy, converted to a velocity increment \(\Delta\mathbf v_{\mathrm{drive}}\) by per-axis gains initialized from training statistics and then fixed. Grip produced a sigmoid gate \(\mathbf{k}_{\mathrm{grip}}\in[0,1]^3\), giving
\[
\Delta\mathbf v_{\mathrm{pred}}=(\mathbf{1}-\mathbf{k}_{\mathrm{grip}})\odot\Delta\mathbf v_{\mathrm{drive}}.
\]
Larger gate values reduce the drive-generated increment, expressing stabilization within the model. Predicted increments updated velocity and pose in normalized coordinates. Torque-proxy and motion objectives established these functional roles, supplemented by regularization terms. The gate is dimensionless and represents attenuation of predicted movement.

To train across holding and movement, we used the 7,736 discovery-cohort stroke sequences together with 2,779 auxiliary posture sequences from eight additional participants. Each stroke or posture sequence contained 100 frames. A random 90/10 sequence-level split (seed 42) allocated these stroke and posture sequences to training and validation. Temporal-weight comparisons used 50 dynamic participants (31 Low, 11 Advanced and 8 Expert). Validation responses are shown in Supplementary Fig.~\ref{suppfig:macro-distributions}.

\paragraph{Training parameters and functional priors.}
To preserve local sensor relationships, the encoder used an eight-neighbour graph and 192 hidden features, with no input or head dropout. Tactile channels were divided by training-set standard deviations without mean centring. Learning rates for reconstruction pretraining, control training and attention fitting were $5\times10^{-4}$, $10^{-3}$ and $2\times10^{-4}$, respectively; control-training batch size was eight. Validation reconstruction loss selected encoder epoch 30, and the attention stage selected epoch eight. Reconstruction and drive--grip orthogonality weights during control training were zero.

To anchor the representations in movement and stabilization, the objective combined motion alignment (weight 1), a torque-proxy target (10; tolerance 0.1), drive--grip co-activation regularization ($5\times10^{-4}$), grip smoothing ($10^{-3}$) and attention regularization (0.01). Grip-logit bias started at 2.2 (gate approximately 0.900), with logits bounded to $[-4,5]$. Fixed torque-to-increment gains were $1.116\times10^{-4}$, $1.231\times10^{-4}$ and $10^{-3}$ in normalized coordinates. These objectives supplied functional roles; validation assessed the resulting task responses.

\section*{Supplementary Methods S9:\texorpdfstring{\\}{ }Temporal-weight sensitivity experiments}
\label{suppmethods:lag-robustness}

\paragraph{Data and fixed model components.}
To test whether the proficiency-related timing pattern depended on modelling choices, we held the data fixed and changed selected model or sampling settings. The inputs were the 7,736 discovery-cohort stroke sequences and 2,779 auxiliary posture sequences, each containing 100 frames (Supplementary Methods~S3), with the same torque-proxy targets. Training and validation used the random 90/10 sequence-level split specified in Supplementary Methods~S8. The encoder, split, objective and functional priors from Supplementary Methods~S8 remained fixed. Comparisons concerned the model-estimated interval in the 50-person dynamic cohort. Configurations and implementation checks accompany the code.

\paragraph{Refits and model capacity.}
To assess optimization sensitivity, new fits used Adam without gradient clipping, compared with the original gradient-norm cap of 5. Checkpoints minimized the full validation objective, including regularization. Conditions within each experiment shared an optimization schedule.

To test sensitivity to starting weights and sample weighting, attention-only refits used 64 epochs, batches of 64 and Adam learning rate 0.0016, compared with the original eight epochs, batch size 8 and rate 0.0002. Global log-weights started at zero or followed a linear slope from $+0.25$ to $-0.25$ to favour recent history, or the reverse to favour remote history; personal offsets started at zero. Balanced sampling gave equal weight to participants within each dataset, then to each person's available actions, while retaining the posture/dynamic mixture. Regularization controls used 0 and 0.1. We retained complete trajectories and selected checkpoints by minimum validation loss, without using proficiency contrasts.

To test representation capacity, each branch used latent dimensions 4, 8 or 16 with seeds 41--43. The encoder stayed fixed while projection and control models trained for 30 epochs at learning rate 0.001 with uniform temporal weights, followed by 32 attention epochs. Sampling experiments used the same staged procedure.

\FloatBarrier

\begin{table}[!htbp]
\centering\small
\caption{\textbf{Frozen-backbone attention refits.} Contrasts are Expert minus Low in nominal milliseconds; brackets show 95\% participant-bootstrap intervals. Epoch is selected by the validation objective, without using proficiency contrasts.}
\label{supptable:lag-attention}
\begin{tabular}{L{0.46\textwidth}rL{0.34\textwidth}}
\toprule Setting & Epoch & Lag contrast (ms) \\
\midrule
Balanced; recent & 54 & -2.16 [-4.86, +0.82] \\
Balanced; no penalty & 63 & -2.21 [-5.22, +1.10] \\
Balanced; penalty 0.1 & 54 & -1.86 [-3.83, +0.33] \\
Balanced; remote & 54 & -2.16 [-4.83, +0.79] \\
Balanced; uniform & 54 & -2.16 [-4.85, +0.81] \\
Window sampling; recent & 63 & -3.57 [-5.93, -0.97] \\
Window sampling; remote & 63 & -3.58 [-5.91, -1.00] \\
Window sampling; uniform & 63 & -3.57 [-5.92, -0.99] \\
\bottomrule
\end{tabular}
\end{table}
\FloatBarrier

\paragraph{Sampling controls.}
To compare sampling rates over the same physical history, we applied a fourth-order causal Butterworth low-pass filter to the 100-Hz stream (20-Hz cutoff, initialized at the first sample). The matched 100-Hz control and both 50-Hz decimation phases used this filtered signal. Single-frame features were calculated before selecting sampling times. Lag lists were $(4,6,8)$ at 100~Hz and $(2,3,4)$ at 50~Hz, giving the same delays. Pose, IMU and torque targets used the same prefilter. Current-time samples were restricted to native indices 24--90 and aligned subsets at 50~Hz.

To keep motion targets comparable across rates, velocity was recalculated from filtered pose differences in native 10-ms displacement units. Target velocity changes were divided by the sampling-interval ratio, and squared gate-difference regularization by its square. Fixed gains retained native units. Both 50-Hz phases used the same participants and recordings; these comparisons therefore isolate sampling and discretization under matched history times and loss scaling.

\paragraph{Inference and prediction interventions.}
To estimate uncertainty across people, confidence intervals used 20,000 percentile participant-bootstrap samples (seed 20260919), preserving group membership and each person's seeds, windows and sampling phases. To assess the predictive role of personal weights, we held the decoder fixed and substituted uniform, global-only, six single-lag profiles, or 100 random permutations of dynamic participants' complete profiles. Errors were averaged within participant before paired comparison; shuffle replicates were averaged before resampling. These substitutions test model sensitivity, including responses outside its training distribution. Supplementary Note~9 summarizes all conditions; weights, training trajectories, splits and checks accompany the records.

\section*{Supplementary Methods S10: Regional coordination from spatial partitioning to temporal order}

We inferred predictive lead--lag relationships between regions and organized them into directed sequences to characterize the temporal structure of contact coordination. A region of interest (ROI) is a group of neighbouring sensors whose activity varies together. The 864-node graph combined physical adjacency, tactile co-activation and normal-force similarity. Person-level cross-validation selected a common normal-to-tactile weight and a Leiden resolution for each proficiency group. Participant-bootstrap graphs measured how often sensors were grouped together; spatially constrained consensus converted these co-assignment values into the final ROIs, merging small disconnected islands with neighbours. The five consensus seeds were 43--47. The consensus-seed score was highest for seed 45; seed 45 was used for the displayed Low/High maps in main Fig.~3e and the reference partitions for quantitative ordering. Main Fig.~4d summarizes ten additional consensus initializations (seeds 63--72), keeping spatial settings and participant membership fixed. Supplementary Note~8 reports their regional counts, skill-specific durations and fixed-count representation controls.

To determine how changes in one region preceded changes elsewhere, we aggregated signals within each ROI and took frame-to-frame differences. Linear-Gaussian conditional transfer entropy (TE) measured the improvement in predicting a destination region from a source region's past. For source $x$, destination $y$ and whole-hand force difference $c$, the reduced regression predicted $y_t$ from an intercept, $y_{t-1}$ and $c_{t-1}$; the full regression added $x_{t-\ell}$. TE was $\max[0,\tfrac12\log((\sigma_{\mathrm{red}}^2+10^{-12})/(\sigma_{\mathrm{full}}^2+10^{-12}))]$, in nats, with variances estimated by mean squared residuals. The maximum over $\ell=1,\ldots,5$ frames selected delays of 10--50~ms. Valid comparisons stayed within strokes and required at least 120 effective samples.

To retain reproducible temporal relationships, each participant-level link was tested against 200 within-stroke circular shifts of the source signal, repeating lag selection for each shift. Empirical $P$ was $(1+\#\{\mathrm{TE}_{\mathrm{surr}}\geq\mathrm{TE}_{\mathrm{obs}}\})/201$, followed by Benjamini--Hochberg correction across links at $q<0.05$. Cohort link strength was the median among participants with a significant non-zero link; stable links required recurrence in at least 70\% of participants. Aggregation used these participant tests without rerunning the surrogate comparison. The separate bootstrap support threshold for tree links was 80\%.

To convert cohort relationships into an ordering, a directed tree maximized retained TE weight, excluded cycles and allowed at most one parent per region. An auxiliary root allowed disconnected components to remain separate after it was removed. Adding delays along branches assigned relative times; their range defined completion time. Thus, the region map supplied the spatial units and TE supplied their temporal organization.

For main Fig.~4d, region count was measured once per proficiency-specific partition. Sequence duration used 1,000 participant-bootstrap samples at the original group sizes, rebuilding cohort relationships and ordering trees in each sample. Low pooled Beginner and Intermediate participants on one pooled-Low map. Bootstrap-mean durations were averaged equally across the five skills within each initialization; points and paired lines show the ten initialization results. The same saved participant draws were used across initializations.

For main Fig.~4c, Beginner and Intermediate participants used the same pooled-Low map with 16 regions; Advanced and Expert participants used their respective 20- and 17-region maps. To assess the mechanical coherence of regional ordering, ROI centres on the two-dimensional hand map were expressed relative to their mean. Each received a unit-length torque direction,
\[
\widetilde{\boldsymbol\tau}_i=
\frac{(\Delta y_i,-\Delta x_i,0)}{\sqrt{\Delta x_i^2+\Delta y_i^2}}.
\]
Only valid non-zero vectors contributed; equal magnitudes isolated location and order from force amplitude. Consistency along the TE-derived sequence and alignment of successive steps were compared with 1,000 randomized orders. This hand-map construction measured ordering geometry; the three-dimensional reconstruction in Supplementary Methods~S5 supplied residual torque geometry.

\section*{Supplementary Methods S11: Anatomical-region comparison}
\label{suppmethods:anatomical-roi}

\paragraph{Representations and participant separation.}
The three-way comparison used 7,736 strokes from 50 participants (Supplementary Methods~S3). A fixed proficiency-stratified split assigned 24 participants to map fitting (15 Low, five Advanced and four Expert) and 26 to evaluation (16 Low, six Advanced and four Expert). All strokes from evaluation participants were excluded from map fitting. Reversing these roles provided a sensitivity analysis. Available skills were retained without imputation; proficiency-specific maps were evaluated within their corresponding groups.

Anatomical15 assigned 864 nodes to 14 finger regions and one palm region, following Sundaram et al.\cite{sundaram2019}. Learned maps used the original eight-dimensional tactile and reconstructed-normal representations, physical adjacency, finger-segment grouping constraints, minimum region size and equal participant weighting. Normal-to-tactile weight 0.6, group resolutions 0.5/0.9/0.6, 50 bootstrap consensus graphs and seed~45 remained fixed. The native pipeline produced 18 regions per group in the original split and 17/19/18 in the reversed split. For the count-matched ablation, adjacent native regions were merged using node-weighted bootstrap co-assignment until 15 remained. Thus, learned15 was a nested coarsening of the adaptive map; evaluation data were not used to select counts or merges.

\paragraph{Whole-hand waveform reconstruction.}
Each node's 100-frame force signal was normalized to unit temporal energy, $z_i=x_i/(\lVert x_i\rVert_2+\epsilon)$, with $\epsilon=10^{-8}$. Reconstruction replaced it with the mean normalized waveform of its region, $\bar z_{R(i)}$. The trial-level error was
\[
E=\frac{\sum_{i,t}[z_i(t)-\bar z_{R(i)}(t)]^2}{\sum_{i,t}z_i(t)^2+\epsilon}.
\]

\paragraph{Within-region palm activity timing.}
For each stroke, active nodes had nonzero force range at least 1\% of the largest whole-hand node range. After subtracting each node's temporal minimum, its force-weighted time centroid was $c_i=\sum_t t f_i^{+}(t)/\sum_t f_i^{+}(t)$, using 10-ms samples. Within-region timing dispersion was the equal-node root-mean-square deviation of active palm-node centroids from their region means,
\[
D=\sqrt{\frac{1}{N_A}\sum_R\sum_{i\in R\cap A}(c_i-\bar c_{R\cap A})^2},
\]
where $A$ denotes active palm nodes and $N_A=|A|$. Strokes required at least two active palm nodes. Lower $D$ denotes more temporally compact regional activity.

\paragraph{Within-region palm force direction.}
Let $f_i(t)$ be scalar contact force and $\hat{\boldsymbol n}_i(t)$ the unit contact normal from the existing reconstruction. Direction coherence was
\[
C=\frac{\sum_t\sum_R\left\lVert\sum_{i\in R\cap\mathcal P}f_i(t)\hat{\boldsymbol n}_i(t)\right\rVert_2}{\sum_t\sum_{i\in\mathcal P}f_i(t)},
\]
where $\mathcal P$ contains palm nodes. All palm forces contributed without thresholding; no force was excluded for invalid normals. Higher $C$ indicates more consistent directions within regions. Because reconstructed normals also enter map learning, this is a mechanical representation measure. Finer nested partitions favour all three metrics by construction; the comparison does not determine an optimal region count.

\paragraph{Participant-level statistics and analysis scope.}
Scores were averaged over strokes within participant--skill cells, then equally over available skills per participant. Two-sided paired tests used 100,000 random sign flips. Ten thousand proficiency-stratified participant resamples supplied marginal percentile 95\% confidence intervals conditional on the fitted maps and upstream reconstruction. This was a follow-up exploratory analysis. Joint Holm adjustment retained all 24 contrasts from three metrics, four map contrasts and two split directions, together with the ten preceding exploratory comparisons. The fourth contrast used 100 local boundary randomizations preserving region counts, region sizes, palm-node counts and connectivity, with boundary length and compactness within 5\%. These controls and all contrasts remain in the analysis records.

\paragraph{Palm waveform illustration.}
Supplementary Fig.~\ref{suppfig:anatomical-palm} retains an earlier within-participant split: successive trial pairs supplied one training and one evaluation stroke; unpaired final trials entered evaluation. Only training trials fitted the eight-dimensional tactile and normal representations and bootstrap co-assignment maps. The original 50-bootstrap procedure used normal weight 0.6 and group resolutions 0.5/0.9/0.6; adjacent regions were merged by co-assignment to 15. Deterministic trial-assignment seeds combined base seed 20260926 with participant--skill identifiers. For backhand inward spinning, adjacent pairs containing at least 16 palm nodes per region were ranked by training-waveform contrast within each group. Highest-ranked pairs were plotted using evaluation trials only, with own-peak normalization within participant and participant means and SEM. The whole-palm reference was normalized separately to its own peak within participant before averaging, as in the regional curves.

\section*{Supplementary Methods S12: Resolution-retention definitions}

To quantify spatial information preserved after subsampling, lower-density labels were projected onto the 864-node reference by nearest sensor coordinates. Co-activation modularity measured grouping on the same held-out graph. Boundary recall was $1-M/B$, where $B$ counts adjacent sensor pairs crossing reference boundaries and $M$ counts those assigned the same projected label. Fine-scale preservation used the smallest quartile of reference regions, rounded up. A region was retained if its best projected match had Dice overlap of at least 0.5; preservation was the fraction of these small regions retained.

To track changes in temporal relationships, retention measured the fraction of stable 100-Hz links also stable at a lower rate. Jaccard similarity measured the intersection of the lower-rate and 100-Hz stable directed edge sets divided by their union. Maps, eligible participants and the full directed candidate set were fixed, with a common 50-ms lag ceiling. Regional ordering and completion time were then recomputed from each rate's TE matrix, maintaining the link from spatial regions to temporal coordination (Supplementary Methods~S14).

\section*{Supplementary Methods S13: Paired residual-resolution experiment}
\label{suppmethods:residual-resolution}

\paragraph{Sampling and fixed comparisons.}
To reduce sensor density without changing stroke timing, we recursively divided the 864 two-dimensional sensor coordinates along their largest span into balanced spatial groups. The sensor nearest each group centre was retained, giving 432, 216, 108, 54 and 27 nodes alongside the native 864. Signals came from those sensors, without block averaging, and used their corresponding three-dimensional geometry. Trials, proficiency labels and 100-frame windows were fixed.

To hold model capacity constant, every resolution retained the native Low training and held-out participants and skill-specific ranks: 6, 5, 4, 5 and 4 for forehand inward spinning, forehand outward spinning, backhand inward spinning, backhand outward spinning and backhand rear-court transition. Each skill and resolution used seeds 42, 43 and 44 for 320 epochs, giving 90 fits. The non-negative model, training-set scaling, smooth-$L_1$ loss and regularizers were unchanged. Activation fitting retained the original convergence settings, followed by per-segment reconstruction-energy matching and percentage-residual calculation. Evaluation used 3,820 segments: 260 Low, 2,090 Advanced and 1,470 Expert. These formed 102 participant--skill combinations (10, 54 and 38) from 25 people (6, 11 and 8). Participants contributed their available skills.

\paragraph{Residual processing, geometry and metrics.}
To separate sampling effects from spatial smoothing, we compared two processing conditions. Both averaged signed curves within skill-by-group combinations as in main Fig.~4b, applied the original temporal filter and then used a 10\% positive-residual threshold. One also retained the original 12-neighbour spatial smoother, with mixing coefficient 0.8 and six passes; the other omitted spatial smoothing. With neighbour count fixed, the smoother covered a larger physical area at lower sensor density.

To evaluate the mechanics of each residual field, sensor positions, normals and racket axes were averaged over the same segments as the residual curves. Mean normals were normalized before computing torque about the estimated pivot. Let $S_{ti}$ be thresholded percentage residual response and $\boldsymbol\tau_i(t)$ its local torque contribution (Supplementary Methods~S5). We used
\[
 F_R=\sum_{t,i}S_{ti},\qquad
 N_R=\sum_i\mathbf{1}(\max_t S_{ti}>0),
\]
\[
 D_\tau=\sum_t\left(\sum_i\|\boldsymbol\tau_i(t)\|-\left\|\sum_i\boldsymbol\tau_i(t)\right\|\right),
 \qquad q_\tau=\frac{D_\tau}{F_R}.
\]
These four measures describe how much residual activity remains, where it occurs and how its torque contributions combine. $F_R$ is cumulative residual magnitude; $N_R$ counts each active sensor once across the 100-frame window; $D_\tau$ sums within-frame geometric cancellation; and $q_\tau$ gives cancellation per unit residual. They use residual-response and reconstructed-geometry units, rather than total force or energy. With no active residual, $F_R$, $N_R$ and $D_\tau$ were zero and $q_\tau$ was unavailable. Skill-by-group mean $q_\tau$ required a non-zero denominator for all three seeds.

To compare totals across node counts, $F_R$, $N_R$ and $D_\tau$ were multiplied by $864/N$ (16 at 54 nodes). The ratio $q_\tau$ was unchanged because the factor cancels. This correction accounts for the number of sampled spatial groups without reconstructing missing signals; both raw and corrected values were retained.

\paragraph{Directional and amplitude summaries.}
To summarize the consistency and size of proficiency differences, metrics were calculated per fit and averaged across three seeds. Directional consistency was $C=(n_{L>A}+n_{A>E})/10$, counting two adjacent decreases in each of five skills. Ties were not decreases; unavailable contrasts remained in the planned denominator and were counted separately. Bands show the seed-to-seed range. The amplitude summary averaged the two signed reductions within each skill, divided by that skill's Low value at 864 nodes, then averaged across skills. It required all five triplets and native denominators to be valid. These shared-observation contrasts were described without trend tests.

\paragraph{Subsampling a full-resolution residual field.}
To isolate reduced spatial readout after inference, we first used all 864 tactile nodes to infer the positive residual fields, then sampled those fields at the same sensor subsets and applied the same geometry and metrics. This condition therefore tests subsampling of a full-resolution result, rather than inference from sparse tactile input. Refitted models used their own native 864-node baseline. All five skills and three seeds were retained. Sampling maps, per-fit values, residual and geometry caches, and reconstruction checks accompany the records.

\paragraph{Scope of the resolution controls.}
The sparse-input refits measure how the residual analysis behaves with fewer measured tactile signals. At 54 nodes, the original 12-neighbour smoother covered a larger physical area and all five Expert fields fell below the positive-residual threshold. Omitting spatial smoothing changed the directional pattern, while subsampling full-resolution residual fields retained more of the original decreasing contrasts. These controls separate reduced readout after full-resolution inference from the combined effects of sparse inputs and residual processing. Each refitting and processing condition used its own native-resolution baseline.

\section*{Supplementary Methods S14: Temporal-resolution experiment}
\label{suppmethods:temporal-fig4cd}

\paragraph{Inputs, uniform sampling and eligibility.}
To trace sampling effects through the regional coordination workflow, we kept ROI maps and strokes fixed while recomputing temporal relationships at 100, 90, 80, 70, 60, 50, 40, 30, 25 and 20~Hz. Saved 100-frame regional and whole-hand signals were resampled with a polyphase rational filter using the default Kaiser-window FIR settings (shape parameter 5) and linear boundary extension. This produced uniform time points at every rate, including rates not evenly dividing the native grid. The 50-ms source-history ceiling allowed five positive lags at 100~Hz, four at 90 and 80~Hz, three at 70 and 60~Hz, two at 50 and 40~Hz, and one at 30, 25 and 20~Hz. Lower rates could not represent a positive lag within this window and were excluded.

To compare the same recordings at every rate, eligibility required at least 120 effective observations at the lowest rate, equivalent to seven 100-frame strokes per participant--skill combination. This rule was applied before rate comparisons. Ordering summaries also required both metrics at all ten rates; missing or isolated-node orders were unavailable. Participant lists, exclusions and metric availability are supplied in the analysis records.

\paragraph{Conditional dependence and full candidate coverage.}
To estimate whether one region's changes predicted another's, regional and whole-hand signals were first-differenced within stroke boundaries. The reduced model predicted destination change from an intercept and the previous destination and whole-hand changes; the full model added lagged source change. Both residual variances used the original $10^{-12}$ numerical offset. Maximum conditional Gaussian TE was evaluated across eligible lags for every directed pair except self-links, without restricting candidates by reference links or spatial location.

To test these relationships against time-shifted controls, each source supplied 200 surrogates shifted independently within each stroke. The same source shifts served all destination tests, with lag selection repeated for every surrogate. Link $P$ values were $(1+b)/201$, where $b$ counts surrogate values at least as large as observed TE. Benjamini--Hochberg adjustment covered all directed links within each participant--skill graph at 0.05 (seed 20260922). Valid comparisons never crossed stroke boundaries. The previous-sample conditioning rule was fixed, so its physical delay changed with sampling rate while the maximum source delay remained 50~ms.

\paragraph{Ordering endpoints.}
To assess how the inferred timing organized contact mechanically, significant participant graphs were converted into regional ordering trees under the original spatial constraints. The ROI geometry was fixed across rates. Chain-level advantage measured the area under cumulative unit-vector torque consistency, standardized against 1,000 shuffled orders of the same regions. Step alignment measured the mean cosine between each new contribution and the preceding cumulative vector. Each participant--skill record retained its original random-seed rule.

To compare proficiency groups, skills were averaged within each participant. Contrasts compared pooled Advanced/Expert with Low participants. Percentile 95\% intervals used 10,000 participant-bootstrap draws within these groups. Matching records and draws across rates supplied paired intervals for changes from 100~Hz, conditional on the fitted graphs.

\paragraph{Cohort relationships and sequence duration.}
To summarize relationships shared across people, cohort link strengths were the median non-zero participant TE, with stable links requiring at least 70\% recurrence. Cohort relationships excluded self-links. Retention was the intersection of lower-rate and 100-Hz stable links divided by the native set size. Directed-link Jaccard similarity was the intersection of the lower-rate and 100-Hz stable directed edge sets divided by their union. It was calculated separately for each skill and proficiency group, then averaged equally across the five skills.

To measure the temporal span of regional coordination, each cohort graph was reduced to a directed tree maximizing retained TE weight, and delays were summed along its branches. For each skill and proficiency group with at least two eligible participants, 1,000 participant-bootstrap samples were reaggregated before deriving the tree and its lag span in milliseconds. Mean bootstrap span supplied the skill--group estimate. Low bootstrap samples pooled eligible Beginner and Intermediate participants and reconstructed ordering trees on the pooled-Low map. In the fixed-count and anatomical representation controls of Supplementary Note~8, Low participants were aggregated on one map within each condition before deriving sequence duration. Samples without directed links were unavailable.

\paragraph{Peak summaries and reproducibility.}
To assess the expansion and refinement pattern at each sampling rate, sequence duration compared Advanced with the larger Low or Expert value within each of five skills. We reported the mean duration margin and the number of skills with an Advanced peak as descriptive summaries. Every rate comparison included all tested rates and skills and used its own matched 100-Hz baseline. Matrices, sample counts, bootstrap spans and full endpoint tables are supplied in the analysis records.

\section*{Supplementary Methods S15: Training feedback evaluation}

To compare mechanism guidance with proficiency-level feedback alone, a computer program randomly assigned 24 learners after selection. Twelve received proficiency-level feedback plus mechanism-derived force/contact and motion guidance, and 12 received proficiency-level feedback alone. Each group contained four beginners, four intermediate players and four advanced players, with no overlap. Both practised for 10~min across five skills, freely allocating time among them. The system delivered feedback; experts did not deliver instruction.

To assess learning, the same three experts scored pre/post videos with proficiency and feedback-group identifiers removed. Clips were shuffled, and experts were instructed to rate each independently without inferring assessment time. The primary endpoint was the expert-rated skill level (1--4), reflecting the execution of each recorded skill. The same four proficiency labels were used in both analyses. Learning assessments were assigned from observed execution of each recorded task, whereas discovery-cohort proficiency additionally incorporated training and competition background. Changes are expressed in skill levels. Ratings were averaged across experts within participant, assessment and skill, then equally across the five skills to produce one pre/post change per participant. Groups were analysed as independent samples.

To assess retention and accumulated learning, six guided learners---two beginners, two intermediate and two advanced---completed three sessions. Both S1--S2 and S2--S3 intervals were 1~day for all six. S1-pre was baseline, S2-pre and S3-pre measured between-session maintenance, and S3-post was the cumulative endpoint. Repeated sessions were evaluated in the guidance group only.

To assess rating reliability, we linked 48 guidance-group videos to participant and assessment state, giving 240 video--skill targets and 720 expert ratings. The 120 first-session state--skill pairs overlap with the 180 pairs in the repeated-session subset and were retained as repeated observations. Two-way absolute-agreement ICC measured reliability for single ratings and the three-rater mean; quadratic-weighted $\kappa$ measured pairwise agreement on the ordinal scale. Omitting one rater at a time tested whether conclusions depended on that expert.

\paragraph{Interpreting the four-level skill framework.}
The original scoring sheets labelled levels 1--4 as beginner, intermediate, advanced and expert. Learning ratings use the same four-level proficiency framework as the discovery cohort, but are assigned from observed task execution. Discovery-cohort proficiency additionally incorporated training and competition background. To express the skill levels in observable terms, Supplementary Table~\ref{supptable:task-performance-anchors} provides task-focused interpretive descriptors informed by a published four-category badminton game-performance assessment rubric\cite{casebolt2020badminton}. The descriptors clarify the meaning of the scale; the original expert ratings and their numerical coding were retained.

\begin{table}[htbp]
\centering
\caption{\textbf{Observable interpretation of skill levels.} The descriptors apply the shared four-level proficiency framework to execution of the recorded task. They draw on the progression from incomplete execution to reliable, controlled performance in badminton assessment\cite{casebolt2020badminton}; the study's five-task rating is not a direct administration of that publication's full game-play rubric.}
\label{supptable:task-performance-anchors}
\small
\setlength{\tabcolsep}{5pt}
\renewcommand{\arraystretch}{1.12}
\begin{tabular}{L{0.07\textwidth}L{0.24\textwidth}L{0.61\textwidth}}
\toprule
Level & Task-performance descriptor & Observable interpretation \\
\midrule
1 & Incomplete execution & Critical stroke elements are absent or repeatedly break down; the intended task action is not reliably produced. \\
2 & Developing execution & The intended stroke is recognizable, but frequent errors or inconsistency remain in movement and shuttle control. \\
3 & Consistent execution & The required stroke elements are generally coordinated and produce the intended task outcome reliably. \\
4 & Refined execution & Execution is consistently controlled and fluent, with precise adjustment of the stroke to the task demands. \\
\bottomrule
\end{tabular}
\end{table}

The overall skill level summarized execution of each skill. Secondary items assessed the visible outcome of force release (spin for the four net-spinning skills and release speed for the rear-court transition), hand motion--force coordination and whole-body coordination. For each participant $i$, skill $s$ and assessment $a$, the three expert ratings $y_{isar}\in\{1,2,3,4\}$ were averaged; the participant-average skill level was $Y_{ia}=\tfrac15\sum_{s=1}^{5}(\tfrac13\sum_{r=1}^{3}y_{isar})$. Improvement was $Y_{i,\mathrm{post}}-Y_{i,\mathrm{pre}}$ in skill levels on the 1--4 scale. A gain of 0.72 levels therefore describes an average change in ratings of the five tasks.

\paragraph{Participant-level outcomes.}
To estimate learner-level improvement, first-session ratings were averaged over five skills within each of 12 guided participants. Mean gain was $+0.72$ levels (participant-bootstrap 95\% CI, $+0.44$ to $+1.02$). In six repeated-session learners, mean within-session, between-session and cumulative changes were $+0.63$, $-0.33$ and $+1.23$ levels (95\% CIs, $+0.36$ to $+0.94$, $-0.62$ to $-0.12$, and $+0.76$ to $+1.68$). Bootstrap resampling kept each participant's skills and assessment states together.

\paragraph{Between-group comparison.}
To estimate the additional gain from guidance, we compared each group's mean pre/post change. Level-only feedback supplied 60 paired participant--skill records, each with three integer expert ratings before and after practice (360 ratings). Across both groups, the first session contained 24 participants, 240 participant--state--skill targets and 720 ratings, excluding later sessions. Percentile 95\% intervals used 20,000 bootstrap draws, resampling 12 participants independently within each group with all their skills and states retained. A two-sided Welch test compared the 12 changes per group. The gain difference was $+0.6667$ levels (bootstrap 95\% CI $+0.3889$ to $+0.9778$; Welch $t=4.1607$, degrees of freedom $=11.7834$, $P=0.001373$). The corresponding Welch 95\% interval was $+0.3168$ to $+1.0165$ levels.

To account for measured baseline differences, we fitted $Y_{\mathrm{post}}=\beta_0+\beta_1 G+\beta_2Y_{\mathrm{pre}}+\epsilon$, with $G=1$ for guidance and $G=0$ for level-only feedback. Pre-training means were 1.8500 and 1.9056, respectively. Ordinary least squares with HC3 standard errors, allowing unequal residual variance, gave $\beta_1=0.6643$ levels, standard error $=0.1779$, and a 95\% $t$ interval of $0.2944$ to $1.0342$ (21 residual degrees of freedom; two-sided $P=0.001223$). Skill-specific contrasts were descriptive and were not separately tested for significance.

\paragraph{Accompanying analysis records.}
Throughout this supplement, analysis records denote the supporting tables, configurations and checks indexed in the repository directory \path{analysis_records/}, with links to the corresponding result files. Complete exploratory endpoints, sensitivity analyses and implementation checks are supplied there. Material-formulation screening is documented separately in these records. The records retain the original comparison families and endpoint definitions.

\section*{Supplementary participant characteristics}

To describe the proficiency range sampled for mechanism discovery, Supplementary Table~\ref{supptable:participant-characteristics} lists the 50 dynamic-recording participants: eight beginners, 23 intermediate players, 11 advanced players and eight experts. Table-specific anonymous identifiers replace names, and achievements omit named institutions and locations. Experience, professional training and competition history follow the participant roster. ``Not reported'' marks a blank entry; ``None reported'' marks an explicit absence of an achievement; a plus sign preserves an open-ended duration. Proficiency follows the study assignment using the combined background information. Individual analyses applied their stated eligibility rules to this cohort.

\begingroup
\small
\setlength{\tabcolsep}{4pt}
\renewcommand{\arraystretch}{1.12}
\begin{longtable}{@{}L{0.09\textwidth}L{0.16\textwidth}L{0.14\textwidth}L{0.15\textwidth}L{0.35\textwidth}@{}}
\caption{\textbf{Anonymized characteristics of the 50-participant dynamic-recording cohort.}}\label{supptable:participant-characteristics}\\
\toprule
ID & Proficiency & Playing experience (years) & Professional training (years) & Highest reported achievement / competitive background \\
\midrule
\endfirsthead
\multicolumn{5}{l}{\tablename~\thetable\ (continued)}\\
\toprule
ID & Proficiency & Playing experience (years) & Professional training (years) & Highest reported achievement / competitive background \\
\midrule
\endhead
\midrule
\multicolumn{5}{r}{Continued on next page}\\
\endfoot
\bottomrule
\endlastfoot
P01 & Beginner & 0 & 0 & None reported \\
P02 & Beginner & 0 & 0 & None reported \\
P03 & Beginner & Not reported & 0 & None reported \\
P04 & Beginner & 0 & 0 & None reported \\
P05 & Beginner & 0 & 0 & None reported \\
P06 & Beginner & 0 & 0 & None reported \\
P07 & Beginner & 2 & 0 & None reported \\
P08 & Beginner & 4 & 0 & None reported \\
P09 & Intermediate & 12 & 0 & None reported \\
P10 & Intermediate & 6 & 0.5 & Institute-level third place \\
P11 & Intermediate & 10 & 0 & Institute-level runner-up \\
P12 & Intermediate & 6 & 0 & Institute-level runner-up \\
P13 & Intermediate & 3 & 0 & Institute-level champion \\
P14 & Intermediate & 3 & 0 & Institute-level champion \\
P15 & Intermediate & 8 & 0 & Institute-level champion \\
P16 & Intermediate & 8 & 0 & None reported \\
P17 & Intermediate & 3 & 0 & Institute-level runner-up \\
P18 & Intermediate & 4 & 0 & None reported \\
P19 & Intermediate & 3.5 & 3.5 & None reported \\
P20 & Intermediate & 6 & 0 & None reported \\
P21 & Intermediate & 5 & 0 & None reported \\
P22 & Intermediate & 2 & 0 & School/college-level champion \\
P23 & Intermediate & 14 & 1 & School/university-level semifinalist \\
P24 & Intermediate & 6 & 0 & None reported \\
P25 & Intermediate & 7 & 0 & None reported \\
P26 & Intermediate & 8 & 0 & None reported \\
P27 & Intermediate & 4 & 1 & None reported \\
P28 & Intermediate & 4 & 3 & None reported \\
P29 & Intermediate & 5 & 0 & None reported \\
P30 & Intermediate & 10 & 3 & None reported \\
P31 & Intermediate & 4 & 0 & None reported \\
\pagebreak
P32 & Advanced & 9 & 0 & School/university-level champion \\
P33 & Advanced & 10 & 1 & Institute-level champion \\
P34 & Advanced & 10+ & 6 & School/university-level champion \\
P35 & Advanced & 8 & 0 & City-level fifth place \\
P36 & Advanced & 11 & 1 & Provincial-level fifth place \\
P37 & Advanced & 7 & 2 & City-level third place \\
P38 & Advanced & 6 & 1 & School/university-level champion \\
P39 & Advanced & 10+ & 1 & School/university-level champion \\
P40 & Advanced & 6 & 1 & School/university-level champion \\
P41 & Advanced & 9 & 3 & School/university-level champion \\
P42 & Advanced & 6 & 3 & School/university-level champion \\
P43 & Expert & 11 & 5 & Provincial-level runner-up \\
P44 & Expert & 15 & 10+ & Provincial-level champion \\
P45 & Expert & 20+ & 10+ & Former national-team player; coach \\
P46 & Expert & 14 & 5 & Provincial-level champion \\
P47 & Expert & 7 & 5 & Provincial-level runner-up \\
P48 & Expert & 4 & 2.5 & Provincial-level champion \\
P49 & Expert & 8 & 1 & Provincial-level runner-up; coach \\
P50 & Expert & 13 & 0 & Provincial-level runner-up; coach \\
\end{longtable}
\endgroup

\bibliographystyle{sn-nature}
\bibliography{refs}